\documentclass{article}

\usepackage{microtype}
\usepackage{graphicx}
\usepackage{subfigure}
\usepackage{booktabs} 

\usepackage{hyperref}

\usepackage[accepted]{icml2025}

\usepackage{multirow}
\usepackage{makecell}
\usepackage{hhline}
\usepackage{booktabs}
\usepackage{bm}
\usepackage{amssymb}
\usepackage{amsmath}
\usepackage{url}
\usepackage{graphicx}
\usepackage{float}
\usepackage{subfigure}
\usepackage{paralist}
\usepackage[ruled, lined, linesnumbered, commentsnumbered, longend]{algorithm2e}
\usepackage{xcolor}
\usepackage{colortbl}
\usepackage{pifont}
\usepackage{tabularx}
\usepackage[T2A,LAE,T1]{fontenc}
\usepackage{CJKutf8}
\usepackage{enumitem}
\usepackage{longtable}
\setlist[itemize]{noitemsep, topsep=0pt}
\usepackage[russian,english]{babel}
\usepackage[most]{tcolorbox}

\definecolor{LightCyan}{rgb}{0.88,1,1}
\newcommand{\boldtitle}[1]{\noindent\textbf{#1}\xspace}
\newcommand{\inlinetitle}[1]{\noindent\textbf{#1}\quad}

\newcommand{\fasize}{\fontsize{8pt}{9pt}\selectfont}
\newcommand{\cmark}{\ding{51}}
\newcommand{\pfix}{\text{\ding{72}}}
\newtcbox{\hlprimarytab}{on line, rounded corners, box align=base, colback=white!10,colframe=white,size=fbox,arc=3pt, before upper=\strut, top=-2pt, bottom=-4pt, left=-2pt, right=-2pt, boxrule=0pt}

\newcommand{\uashifted}{\raisebox{0.5\depth}{\tiny$\uparrow$}}

\newcommand{\uagray}[1]{{\small\hlprimarytab{\uashifted{#1}}}}

\newcommand{\bLozenge}{\mathbin{\blacklozenge}}
 
\newcommand{\ouradapter}{Franken-Adapter\xspace}
\newcommand{\palmtwo}{\texttt{PaLM2}\xspace}
\newcommand{\gemmatwo}{\texttt{Gemma2}\xspace}
\newcommand{\aya}{\texttt{Aya23}\xspace}

\newcommand{\pt}{\textsc{Pt}\xspace}
\newcommand{\la}{Language Adaptation\xspace}
\newcommand{\es}{\textsc{La}\xspace}
\newcommand{\flan}{\textsc{Fl}\xspace}
\newcommand{\franken}{\textsc{Fa}\xspace}
\newcommand{\fala}{\textsc{+Lr}\xspace}

\newcommand{\sea}{\textsc{Sea}\xspace}
\newcommand{\afr}{\textsc{Afr}\xspace}
\newcommand{\ind}{\textsc{Ind}\xspace}

\newcommand{\flores}{\textsc{Flores-200}\xspace}
\newcommand{\belebele}{\textsc{BeleBele}\xspace}
\newcommand{\xorqain}{\textsc{XorQA-In}\xspace}
\newcommand{\xsumin}{\textsc{XSum-In}\xspace}
\newcommand{\sib}{\textsc{Sib-200}\xspace}

\usepackage{ntheorem}

\theoremstyle{nonumberplain}

\def\eg{{e.g.,}\xspace}
\def\ie{{i.e.,}\xspace}
\def\versus{{\em v.s.}\xspace}

\def\etc{{\em etc.}\xspace}
\definecolor{lightblue}{HTML}{bdd6fb}
\definecolor{boxgray}{gray}{0.9}
\definecolor{bgyellow}{HTML}{fcebde}
\definecolor{bgred}{HTML}{d77470}
\definecolor{bggrey}{HTML}{dcc0e5}

\usepackage{amsmath}
\usepackage{amssymb}
\usepackage{mathtools}

\usepackage[capitalize,noabbrev]{cleveref}

\usepackage[textsize=tiny]{todonotes}

\icmltitlerunning{\ouradapter: Cross-Lingual Adaptation of LLMs by Embedding Surgery}

\begin{document}

\twocolumn[
\icmltitle{\ouradapter: Cross-Lingual Adaptation of LLMs by Embedding Surgery}



\icmlsetsymbol{equal}{*}

\begin{icmlauthorlist}
\icmlauthor{Fan Jiang}{equal,unimelb}
\icmlauthor{Honglin Yu}{gra}
\icmlauthor{Grace Chung}{gra}
\icmlauthor{Trevor Cohn}{gra}
\end{icmlauthorlist}

\icmlaffiliation{unimelb}{The University of Melbourne}
\icmlaffiliation{gra}{Google}

\icmlcorrespondingauthor{Fan Jiang}{fan.jiang1@student.unimelb.edu.au}

\icmlkeywords{Machine Learning, ICML}

\vskip 0.3in
]



\printAffiliationsAndNotice{\icmlEqualContribution} 

\begin{abstract}
The capabilities of Large Language Models (LLMs) in low-resource languages lag far behind those in English, making their universal accessibility a significant challenge.
To alleviate this, we present \emph{\ouradapter}, a modular language adaptation approach for decoder-only LLMs with embedding surgery. Our method begins by creating customized vocabularies for target languages and performing language adaptation through embedding tuning on multilingual data. These pre-trained embeddings are subsequently integrated with LLMs that have been instruction-tuned on English alignment data to enable zero-shot cross-lingual transfer. 
Our experiments on \gemmatwo models with up to 27B parameters demonstrate improvements of up to 20\% across 96 languages, spanning both discriminative and generative tasks, with minimal regressions ($<$1\%) in English. Further in-depth analysis reveals the critical role of customizing tokenizers in enhancing language adaptation, while boosting inference efficiency.
Additionally, we show the versatility of our method by achieving a 14\% improvement over a math-optimized LLM across 20 languages, offering a modular solution to transfer reasoning abilities across languages post hoc.
\end{abstract}

\section{Introduction}


Large Language Models (LLMs) have transformed the field of natural language processing through pre-training on extensive web-scale corpora~\citep{gpt3, geminiteam2024geminifamilyhighlycapable}. Despite these advancements, their success has been primarily centered on English, leaving the multilingual ability less explored. While the multilingual potential of LLMs has been demonstrated across multiple languages~\citep{shi2023language}, their practical applications remain largely confined to a limited set of high-resource languages. This limitation reduces their utility for users speaking under-represented languages~\citep{ahia-etal-2023-languages}.

\begin{figure}[t]
    \centering
    \includegraphics[width=\linewidth]{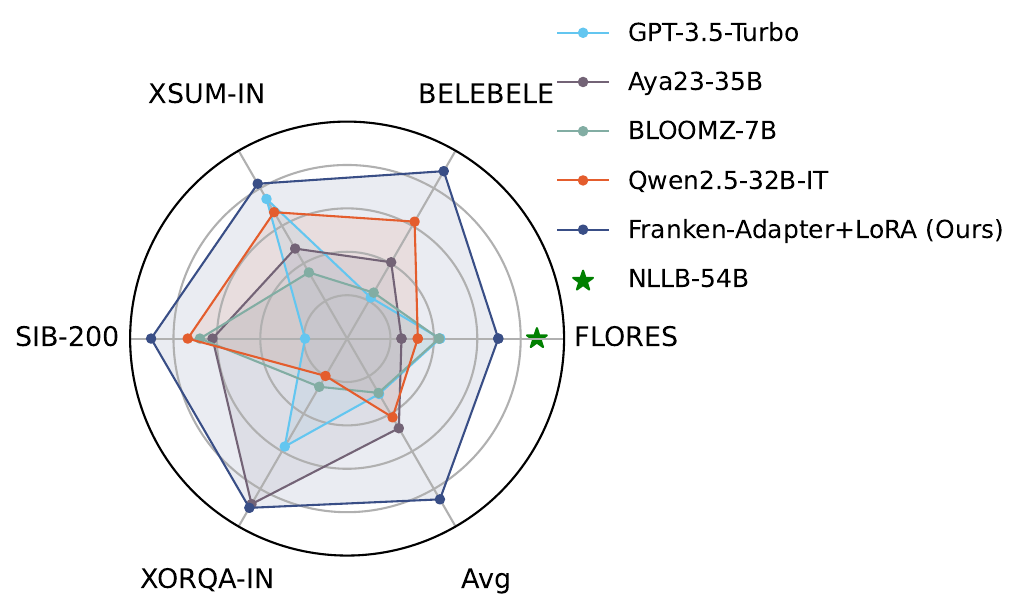}
    \vspace{-9mm}
    \caption{Zero-shot performance comparison between our best model (\gemmatwo-\texttt{27B}-\texttt{\ouradapter-LoRA}) and state-of-the-art LLMs on five benchmarks.}
    \vspace{-4mm}
    \label{fig:sota_comparison}
\end{figure}
\begin{figure}[t]
    \centering
    \includegraphics[width=0.95\linewidth]{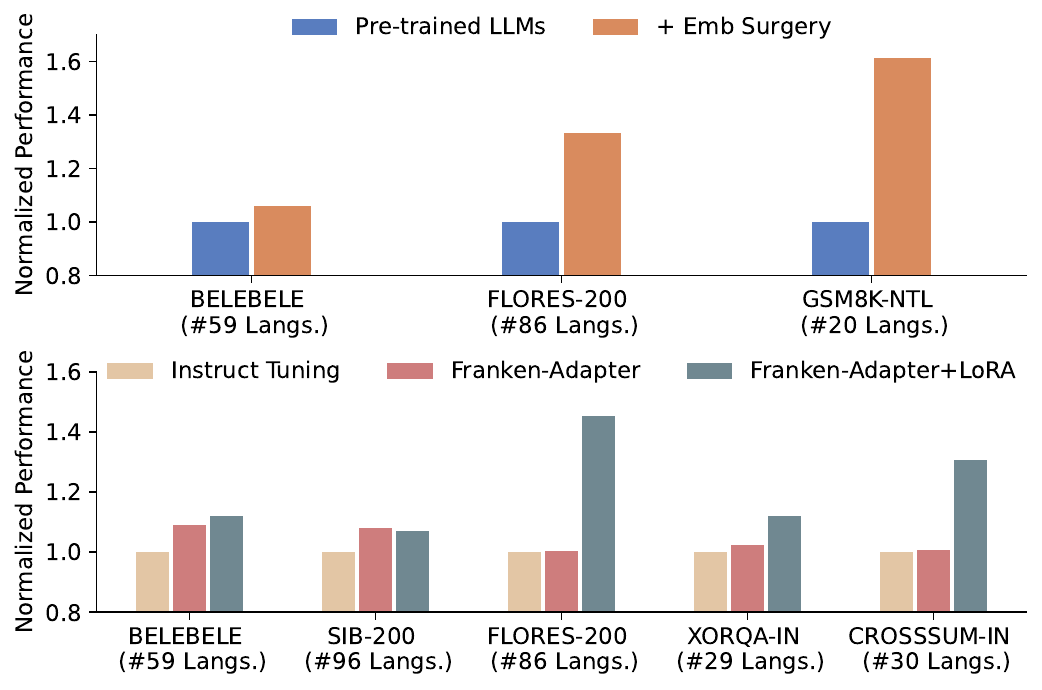}
    \vspace{-5mm}
    \caption{Result summary across diverse benchmarks. Scores are normalized \versus Pre-trained (top) and instruction-tuned (bottom) LLMs. All scores are macro-averaged across all sizes of \gemmatwo.}
    \vspace{-6mm}
    \label{fig:result_summarization}
\end{figure}

A widely adopted approach to multilingual adaptation involves continued pre-training on additional data in target languages by using pre-trained LLMs as initialization~\citep{fujii2024continual, zheng-etal-2024-breaking}. This paradigm typically requires full-parameter tuning on vast data, making the adaptation of a new LLM to accommodate every language prohibitively costly. Moreover, such adaptation poses a significant risk of catastrophic forgetting, whereby the LLM loses previously acquired knowledge from the initial pre-training phase~\citep{luo2024empiricalstudycatastrophicforgetting, shi2024continuallearninglargelanguage}. Although alternative methods such as adapters~\citep{pfeiffer-etal-2021-unks} or LoRA~\citep{hu2022lora} offer more efficient solutions for language adaptation, their capacity to acquire new knowledge remains limited~\citep{biderman2024lora}. 
Model composition \citep{bansal2024llm} can achieve cross-lingual skill transfer by combining a general LLM with a smaller specialist model, to realize synergies over both models' capabilities. However, it requires decoding on the two composed models, which introduces extra inference costs.
These challenges underscore the importance of finding new methods to adapt LLMs to new languages efficiently and effectively.

\citet{artetxe-etal-2020-cross} propose an efficient zero-shot cross-lingual transfer method that adapts English pre-trained language models (PLMs) to unseen languages by learning new embeddings while keeping the transformer layers fixed, hypothesizing that monolingual models learn semantic linguistic abstractions
that generalize across languages. Despite showing promising results competitive to full pre-training, this \emph{embedding surgery} paradigm still has several shortcomings: its reliance on `outdated` encoder-only PLMs limits the applicability, and its monolingual transfer strategy (\ie one embedding per language) reduces its efficiency. These limitations prompt important questions: 1) Can this approach be extended to modern decoder-only LLMs? 2) Can we achieve efficient cross-lingual transfer to many languages and avoid the creation of monolingual embeddings?

\begin{figure*}
    \setlength{\abovecaptionskip}{-0.0001cm}
    \setlength{\belowcaptionskip}{-0.35cm}
    \centering
    \includegraphics[width=0.7\linewidth]{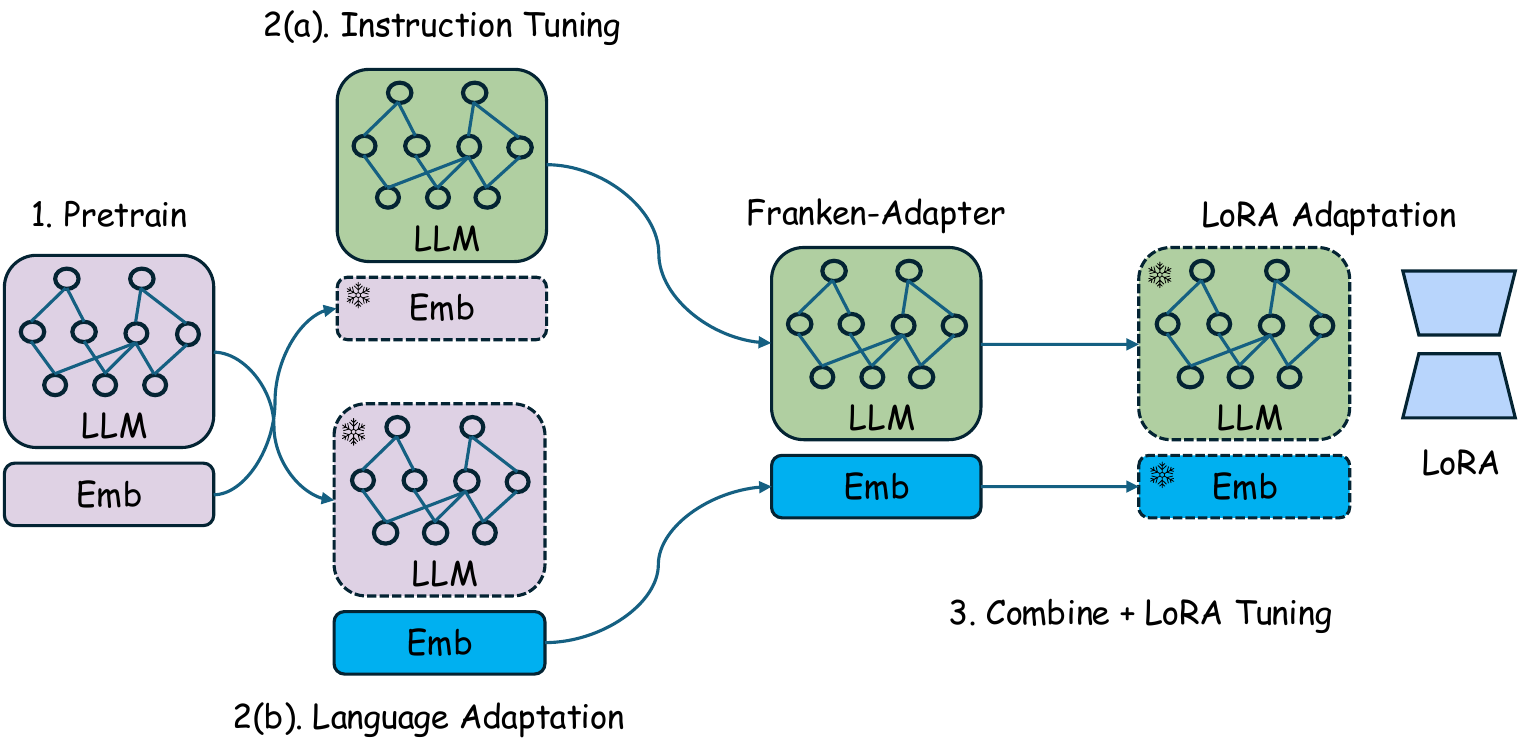}
    \vspace{-4mm}
    \caption{Overview of our \ouradapter pipeline: 1) pre-train a LLM on English-dominant data; 2a) freeze the original embeddings of LLMs and instruction-tune the transformer body using English alignment data; 2b) learn new multilingual embeddings by freezing the transformer body for target language adaptation of LLMs; 3) combine new embeddings with instruction-tuned transformer body as the \emph{\ouradapter} and further perform LoRA tuning to connect the combined components for enhanced cross-lingual transfer.}
    \vspace{-5mm}
    \label{fig:franken_adapter}
\end{figure*}

In this work, we explore \emph{multilingual embedding surgery} on decoder-only LLMs and demonstrate its effectiveness in improving cross-lingual transfer.
As shown in Figure~\ref{fig:franken_adapter}, our method starts with two parallel training from an existing LLM primarily pre-trained on English data: 1) We adapt it to a target language group\footnote{We focus on three language groups: South East Asia (\sea), Indic (\ind), and African (\afr).} by learning new multilingual embeddings while freezing the transformer body; 2) We employ English alignment data to instruction-tune the transformer body, keeping the original embeddings fixed. After this, the newly trained multilingual embeddings are combined with the instruction-tuned transformer body for efficient zero-shot cross-lingual transfer, which we designate as \emph{\ouradapter}. Optionally, we can further enhance the compatibility of the composed components within \emph{\ouradapter} via a cost-effective LoRA-based adaptation. 

We show that a single \emph{language adaptation} step, employing the customized embeddings with pre-trained LLM, can significantly enhance the multilingual performance across diverse tasks (Figure~\ref{fig:result_summarization} top). Furthermore, the \emph{\ouradapter} framework, which integrates new embeddings into instruction-tuned LLMs, enables zero-shot cross-lingual transfer straight away, and can further benefit from cost-effective LoRA adaptation (Figure~\ref{fig:result_summarization} bottom). 
In summary, our contributions are:
\begin{compactenum}
    \item We demonstrate that embedding tuning is effective for language adaptation of LLMs, and systematically evaluate the critical role of tokenizers in this process. Our results show large performance improvement on low-resource languages when using customized tokenizers.
    \item Our \ouradapter approach provides a modular framework for efficient zero-shot cross-lingual transfer of LLMs via embedding surgery. Notably, we show that our best model outperforms benchmark LLMs at comparable sizes across diverse tasks (Figure~\ref{fig:sota_comparison}).
\end{compactenum}

\section{Methodology}
\begin{figure}
    \setlength{\abovecaptionskip}{-0.0001cm}
    \setlength{\belowcaptionskip}{-0.35cm}
    \centering
    \includegraphics[width=0.95\linewidth]{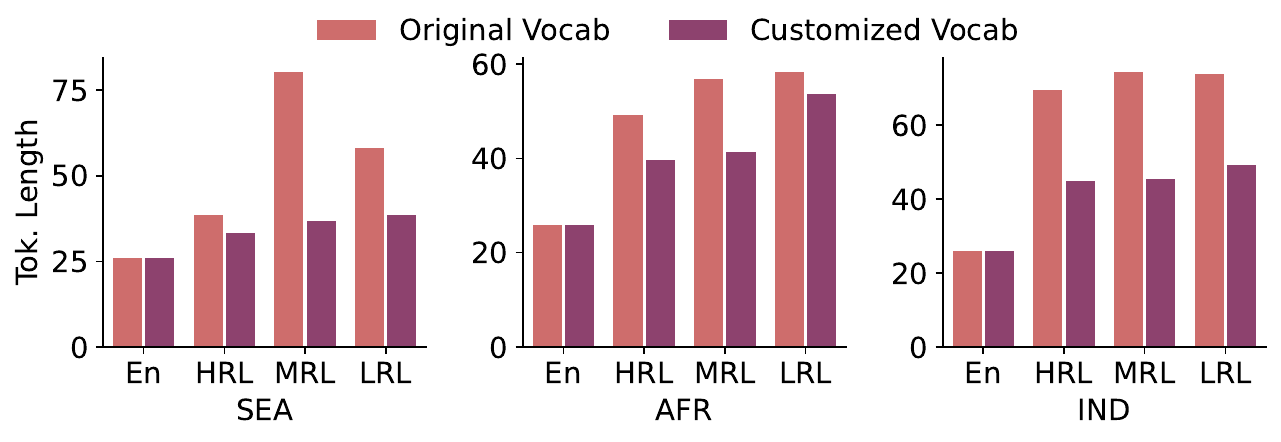}
    \vspace{-6mm}
    \caption{The tokenization comparison between using the vanilla and customized multilingual tokenizers on \gemmatwo. Tok. Length refers to the average number of tokens required to represent the same amount of texts.} 
    \vspace{-6mm}
    \label{fig:fertility_comparison}
\end{figure}

Modern LLMs exhibits limited multilingual ability, primarily due to the dominance of English training data and the relatively small multilingual proportion~\citep{dubey2024llama3herdmodels}. Additionally, the tokenizers employed in these LLMs, which are constructed from sub-sampled pre-training corpora, are biased towards English and several high-resource languages. This results in fragmenting texts from long-tail languages into too many tokens (Figure~\ref{fig:fertility_comparison}), thereby degrading the performance and efficiency of processing such languages~\citep{ahuja-etal-2023-mega}. In this paper, we demonstrate that having tokenizers that provide equitable representation for low-resource languages is critical to the effectiveness of our proposed \emph{\ouradapter} approach. 

\subsection{Customized Vocabulary Construction}\label{sec:vocab_construct}
Our strategy involves constructing distinct tokenizers for each language group (\S\ref{sec:languages}).\footnote{Our preliminary findings suggest that employing distinct tokenizers for each language group enhances cross-lingual transfer. Please refer to Appendix Figure~\ref{fig:joint_vs_distinct_tokenizer} for a comparison with the variant of using a single tokenizer for all language groups.} Tailoring tokenizers to specific language groups enhances cross-lingual transfer among geographically related long-tail languages compared to using monolingual tokenizers. Moreover, this approach avoids the shortcomings of a universal tokenizer that treats all low-resource languages uniformly poorly. Based on this, we propose a \emph{Prune-with-Extension} approach for developing tokenizers optimized for language adaptation while maintaining English ability. 
First, we prune the tokenizer of target LLMs by removing non-English tokens. Then the pruned tokenizer is extended through adding new tokens, which are obtained by training tokenizers for low-resource languages using BPE~\citep{bpe,sennrich-etal-2016-neural}.

\inlinetitle{Pruning the Tokenizer}
To preserve the pre-trained knowledge embedded in the language model, current approaches often expand the vocabulary by adding new tokens~\citep{fujii2024continual, cui2024efficienteffectivetextencoding}. This, however, can substantially increase pre-training time due to the extra computational cost of the softmax layer~\citep{liang-etal-2023-xlm}.\footnote{Embedding and $\operatorname{softmax}$ layers are typically tied in modern LLMs. LLMs evaluated in this work all follow this setting.} 
To avoid this, we first prune the existing tokenizers by retaining only English tokens before adding those from low-resource languages.
Given the predominant English training data for LLMs, we hypothesize that a significant portion of their knowledge is associated with English tokens, and reusing English tokens can effectively retain this knowledge~\citep{garcia-etal-2021-towards}. 
In our implementation, for a given LLM, we identify English tokens by tokenizing a set of 20 million English sentences using its tokenizer, with further filtering through removing non-Latin script tokens.\footnote{40\% of the tokens are discarded: these are non-Latin scripts tokens from high-resource languages, domain-specific tokens (\eg code), and very rare English tokens.}

\inlinetitle{Training Multilingual Tokenizers}
To get the data for building a multilingual vocabulary for long-tail languages, we sample from the Next Thousand Languages (NTL) corpus~\citep{caswell-etal-2020-language, bapna2022building}.
Our empirical analysis reveals that sampling data for each language up to a maximum of 500K lines from NTL effectively addresses the imbalance between high- and low-resource languages, outperforming temperature-sampling techniques. 
Subsequently, we train a BPE tokenizer using the sampled data to generate a vocabulary whose size aligns with that of the target LLMs.

\inlinetitle{Extending the Pruned Tokenizer}
We sequentially add the identified English tokens followed by tokens from the newly built multilingual tokenizer. Both types of tokens are added in the same preference order as in their respective tokenizers. The final vocabulary maintains the same size as the original tokenizer, with over 60\% token overlap, resulting in negligible variations in English tokenizations (see  Table~\ref{tab:tokenization_qualitative_examples} in Appendix).
Figure~\ref{fig:fertility_comparison} shows our final vocabulary achieves significant compression rate improvements by consistently producing shorter sequences across languages of a spectrum of resource levels while barely affecting English.

\subsection{Model Training}\label{sec:model_training}

\inlinetitle{Embedding Initialization}
To maximally inherit the pre-trained knowledge embedded in the target LLM's embedding layer, we adopt a strategy inspired by \citet{gee-etal-2022-fast}. For tokens that overlap between the target LLM's vocabulary and our multilingual vocabulary, we directly copy the corresponding embeddings. For new tokens, we employ the LLM's original tokenizer to decompose them into subtokens and initialize their embeddings using the average of their subtoken embeddings.

\inlinetitle{Language Adaptation}
To adapt the LLMs to new languages, we follow~\citet{artetxe-etal-2020-cross} to only fine-tune the customized embeddings on curated multilingual data $\mathcal{D}_{la}$ while keeping the transformer body frozen (Figure~\ref{fig:franken_adapter} step 2(b)), with the same training objective used in the initial LLM pre-training phase. This is based on the assumption that the pre-trained transformer body encapsulates universal cross-lingual knowledge~\citep{zhao2024how,wendler-etal-2024-llamas,tang-etal-2024-language}, while the embedding layer encodes language-specific information, which suggests embedding tuning should be effective for language adaptation. 

\inlinetitle{\ouradapter}
To facilitate zero-shot cross-lingual transfer, \citet{artetxe-etal-2020-cross} fine-tunes the transformer body of PLMs on a specific English downstream task, then incorporates monolingual embeddings tailored for a second language. This approach necessitates task-specific fine-tuning of the transformer and language-specific embedding creation. In contrast, our \ouradapter method instruction-tunes the transformer body of LLMs on a diverse range of English tasks $\mathcal{D}_{it}$~\citep{wei2022finetuned} (Figure~\ref{fig:franken_adapter} step 2(a)). Notably, we employ the LLM's \emph{original} embeddings and keep them frozen in this step.
We then integrate the customized embeddings obtained from the language adaptation stage into the instruction-tuned transformer body. This results in the \ouradapter (Figure~\ref{fig:franken_adapter} step 3), a modular framework designed for zero-shot cross-lingual transfer.\footnote{Freezing embeddings makes the instruction tuning and language adaptation processes symmetry. This enhances modularity and improves the parameter compatibility of \ouradapter.}

\inlinetitle{LoRA-Adaptation}
Since the transformer body and customized embeddings are independently trained, the \ouradapter approach may suffer from incompatible parameters. Our empirical findings indicate that \ouradapter is effective for discriminative tasks but sometimes underperforms an instruction-tuned baseline on generative tasks. To mitigate this and ensure the assembled model's effectiveness across various tasks, we insert LoRA weights into the self-attention layer of the tuned transformer body (Figure~\ref{fig:franken_adapter} step 3). These weights are then fine-tuned on a sub-sampled joint corpus $\mathcal{D}_{mix}=\mathcal{D}_{la}\cup\mathcal{D}_{it}$, while keeping both the transformer body and embeddings frozen.

\subsection{Discussion}
Compared to the typical continued pre-training approach with full-parameter tuning on target languages, \ouradapter offers several advantages: 1) Customized tokenizers ensure fairer representations for target languages, which not only facilitates cross-lingual transfer but also improves training and inference efficiency. 2) Embedding tuning further enhances language adaptation efficiency by keeping the transformer body intact. This mitigates catastrophic forgetting that is prevalent in traditional methods. 3) The modular nature of \ouradapter enables the reuse of existing models with pre-established capabilities through simple and cost-effective embedding swapping in a fully \emph{post-hoc} manner. This property distinguishes \ouradapter from typical methods that requires costly and separate adaptations to acquire different skills, making it more desirable for effective and efficient cross-lingual transfer in many settings. (See Appendix~\ref{appendix:cpt} for more details with empirical evidence.)

\section{Experiment setup}
\subsection{Pre-training Data}
The data $\mathcal{D}_{la}$ for embedding training is a mixed corpus with 65\% sentence-level and 35\% document-level texts. The sentence-level data is exclusively from the Next Thousand Languages (NTL) corpora~\citep{caswell-etal-2020-language, bapna2022building}, which provides web-crawled monolingual sentences and translation pairs for over 1000 languages. For document-level texts, we sample data from multilingual Wikipedia and mC4~\citep{xue-etal-2021-mt5}. We use UniMax sampling~\citep{chung2023unimax} with $N=5$ to up-sample low-resource languages. Additionally, we consider English as a high-resource language and always include it in the training mixture to prevent catastrophic forgetting.

We take FLAN~\citep{wei2022finetuned} as the instruction tuning data $\mathcal{D}_{it}$. 
The data $\mathcal{D}_{mix}$ used in LoRA-Adaptation consists of a 10\% sample of $\mathcal{D}_{it}$, combined with an equal number of instances from $\mathcal{D}_{la}$.\footnote{This ensures the instruction-following ability isn't forgotten. Please refer to Appendix Figure~\ref{fig:data_ablation_on_lora_adaptation} for detailed analysis.}

\subsection{Languages}\label{sec:languages}
We select languages from three families based on geographic relations: South East Asian (\sea), African (\afr), and Indic (\ind). 
This results in 212 languages from \sea, 392 from \afr, and 170 from \ind. Each regional dataset is processed separately, with a tailored tokenizer, language-adapted embeddings and LoRA update parameters.


\subsection{Models}
Our evaluation is focused on \gemmatwo (2B, 9B, 27B)~\citep{team2024gemma}. We also test the generalization ability of our method in two LLMs with varying degrees of multilinguality: \aya (8B, 35B)~\citep{aryabumi2024aya} and \palmtwo (XXS, S)~\citep{anil2023palm}.

As shown in Figure~\ref{fig:franken_adapter}, we end up having four types of models: (i) \texttt{$*$-FLAN} (step-2a): models that undergo instruction-tuning. (ii) \texttt{Lang-Adapt} (step-2b): the LLMs after language adaptation with embedding tuning. (iii) \ouradapter (step-3 Combine): model denoted as \texttt{$*$-FA} is constructed by combining the transformer body of \texttt{$*$-FLAN} with the embeddings from \texttt{Lang-Adapt}. (iv): \texttt{LoRA-Adapt} (step-3 LoRA Tuning): \ouradapter models with the LoRA-Adaptation process. Detailed training procedures for each model type are in Appendix~\ref{appendix:training_details}.

  

\subsection{Evaluation Tasks}
For LLMs after language adaptation, we adopt the \emph{five-shot} prompting strategy. In contrast, \ouradapter is evaluated in a \emph{zero-shot} setting, given it has been instruction tuned. We also evaluate \ouradapter using a compiled English benchmark (Appendix~\ref{appendix:english_tasks}) to examine potential regressions in general English ability.

\textbf{\belebele~\citep{bandarkar-etal-2024-belebele}}
is a multiple-choice reading comprehension (MRC) dataset 
with 122 languages.
The dataset contains 900 instances created from short passages sampled from \flores. 
We evaluation on all the 15 \sea, 25 \afr, and 19 \ind languages. 
We follow the original paper to use the \texttt{Accuracy} metric and sample five-shot prompts from the English training dataset. 

\textbf{\sib~\citep{adelani-etal-2024-sib}}
is a topic classification task based on texts sampled from \flores. 
We select all the 25 \sea, 45 \afr and 26 \ind languages for evaluation with zero-shot prompting and use the \texttt{Accuracy} as the metric for performance comparison.

\textbf{\flores~\citep{goyal-etal-2022-flores}}
is a machine translation dataset with 200 languages. 
We evaluate on all the 23 \sea, 42 \afr, and 21 \ind languages using \texttt{ChrF++} and sample five-shot prompts from the dev set for few-shot evaluation.

\textbf{GSM8K-NTL~\citep{shi2023language, bansal2024llm}}
is a silver benchmark created by automatically translating the English GSM8K dataset into 25 low-resource languages. For evaluation, we focus on 5 \sea, 5 \afr, and 10 \ind languages and follow~\citet{bansal2024llm} to use the \texttt{Accuracy} metric and fixed five-shot prompts.

\textbf{\textsc{IndicGenBench}~\citep{singh-etal-2024-indicgenbench}}
is a human-curation benchmark across 29 Indic languages. We evaluate on the cross-lingual question-answering dataset~\textsc{XorQA-In} and the cross-lingual summarization dataset~\textsc{XSum-In} with zero-shot prompting. The token-level \texttt{F1} and \texttt{ChrF} scores are reported in~\textsc{XorQA-In} and~\textsc{XSum-In}, respectively.

\begin{figure}
    \setlength{\abovecaptionskip}{-0.0001cm}
    \setlength{\belowcaptionskip}{-0.5cm}
    \centering
    \includegraphics[width=\linewidth]{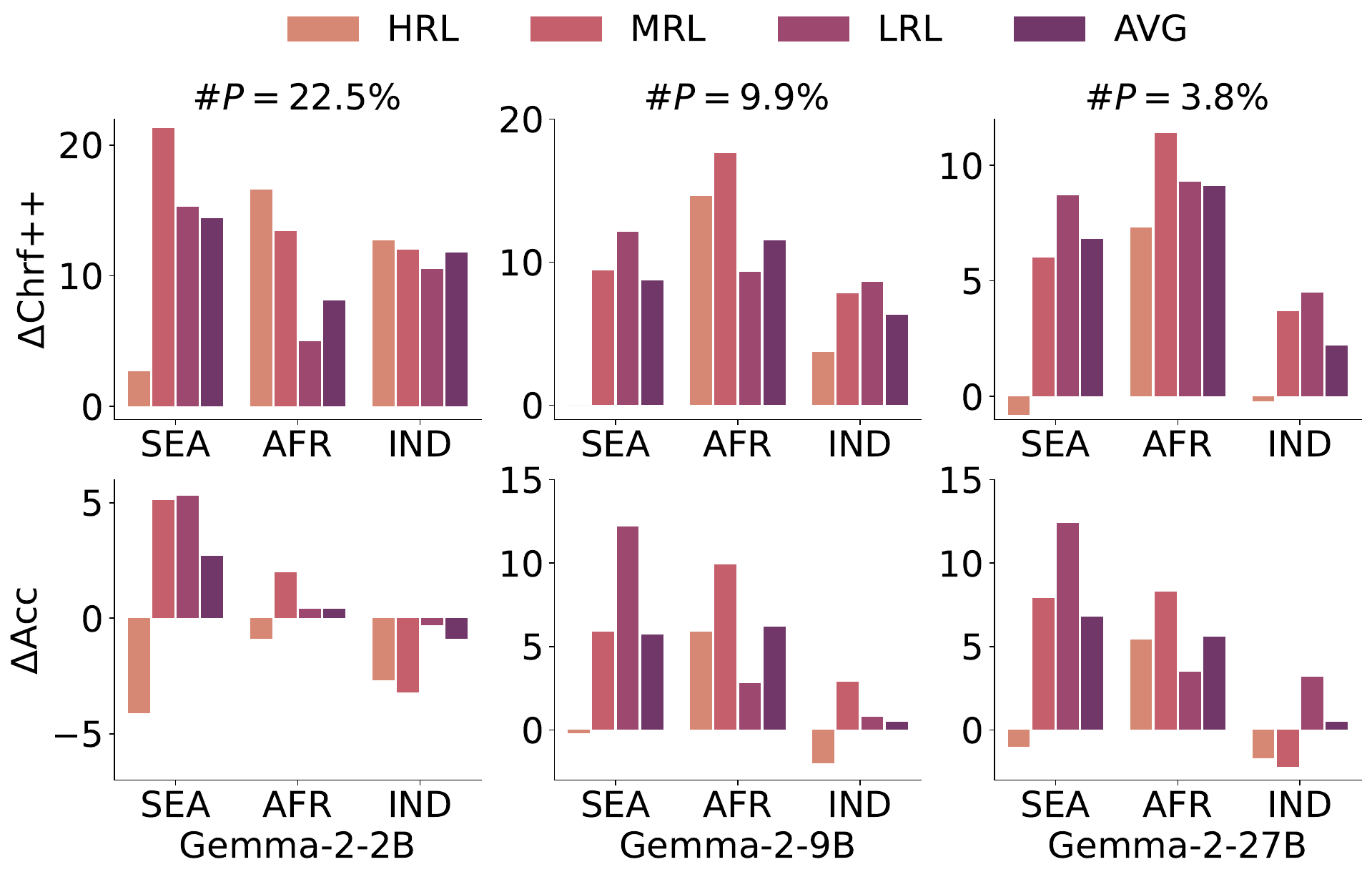}
    \vspace{-6mm}
    \caption{\flores \textsc{En-Xx} and \belebele \la results across all sizes of \gemmatwo models with 5-shot prompting. Absolute gains over the pre-trained models are reported. $\#P$: fraction of tuned parameters (\ie embedding parameters).}
    \vspace{-6mm}
    \label{fig:gemma_results}
\end{figure}

\subsection{Main Results}
We first present the results of \la through embedding tuning on its own (\S\ref{sec:model_training} -- Language Adaptation) in \S\ref{sec:embed_surgery_results}. Then we evaluate the application of \ouradapter alongside LoRA adaptation (\S\ref{sec:model_training} -- \ouradapter \& LoRA-Adaptation) in \S\ref{sec:franken_adapter_results}.

\begin{table}[t]
\setlength{\tabcolsep}{6pt}
\footnotesize
\centering
\vspace{-3mm}
\caption{Language adaptation results on GSM8K-NTL. Best results are in \textbf{bold}. $\star$ denotes results reported by~\citet{bansal2024llm}.}
\resizebox{\linewidth}{!}{
\begin{tabular}{l|ccc|c}
    \toprule
    \bf Model & \bf\sea & \bf\afr & \bf\ind & \bf Avg.\\ 
    \midrule
    CALM (\palmtwo-\texttt{XXS-NTL+S})$^\star$ & 25.3 & 17.8 & 17.9 & 19.8 \\
    \palmtwo-\texttt{S-NTL} & 25.2 & 17.4 & 15.1 & 18.2 \\
    \palmtwo-\texttt{S} & 22.0 & 15.3 & 14.2 & 16.4 \\
    {\pfix}  \texttt{Lang-Adapt} & \bf 25.6 & \bf 18.8 & \bf 22.3 & \bf 22.3 \\
    \midrule
    \gemmatwo-\texttt{9B} & 19.9 & 18.3 & 13.6 & 16.4 \\
    {\pfix}  \texttt{Lang-Adapt} & \bf 36.4 & \bf 25.8 & \bf 29.6 & \bf 30.4 \\
    \gemmatwo-\texttt{27B} & 30.6 & 22.2 & 18.4 & 22.4 \\
    {\pfix}  \texttt{Lang-Adapt} & \bf 41.9 & \bf 31.8 & \bf 27.5 & \bf 32.2 \\
    
    \bottomrule
    \multicolumn{5}{c}{} \vspace{-6mm}
    \end{tabular}
    }
\label{tab:gsm_8k_ntl_results}
\end{table}

\begin{table}[t]
\setlength{\tabcolsep}{8.5pt}
\footnotesize
\centering
\vspace{-3mm}
\caption{\ouradapter results on GSM8K-NTL. The best results are marked in \textbf{bold}. $\dagger$ indicates that the LLMs are instruction-tuned without embedding freezing.}
\resizebox{\linewidth}{!}{
\begin{tabular}{l|ccc|c}
    \toprule
    \bf Model & \bf\sea & \bf\afr & \bf\ind & \bf Avg.\\ 
    \midrule
    \gemmatwo-\texttt{9B-FLAN} & 31.7 & 22.1 & 21.4 & 24.2 \\
    {\pfix}  \texttt{\ouradapter} & \bf 40.4 & \bf 27.8 & \bf 28.9 & \bf 31.5 \\
    \gemmatwo-\texttt{27B-FLAN} & 33.5 & 26.3 & 27.4 & 28.7 \\
    {\pfix}  \texttt{\ouradapter} & \bf 44.9 & \bf 31.0 & \bf 30.9 & \bf 34.4 \\
    \midrule
    \gemmatwo-\texttt{9B-IT}$^\dagger$ & 40.4 & 28.6 & 33.0 & 33.8 \\
    {\pfix}  \texttt{\ouradapter} & \bf 53.0 & \bf 35.6 & \bf 39.4 & \bf 41.9 \\
    \gemmatwo-\texttt{27B-IT}$^\dagger$ & 44.4 & \bf 34.1 & 37.6 & 38.4 \\
    {\pfix}  \texttt{\ouradapter} & \bf 49.6 & 26.5 & \bf 42.7 & \bf 40.4 \\
    
    \midrule
    \gemmatwo-\texttt{9B-Math} & 48.0 & 37.3 & \bf 35.2 & 38.9 \\
    {\pfix}  \texttt{\ouradapter} & \bf 62.1 & \bf 45.0 & 34.8 & \bf 44.2 \\
    \gemmatwo-\texttt{27B-Math} & 49.8 & 35.4 & \bf 32.3 & 37.5 \\
    {\pfix}  \texttt{\ouradapter} & \bf 59.0 & \bf 39.7 & 30.9 & \bf 40.1 \\
    
    \bottomrule
    \multicolumn{5}{c}{} \vspace{-10mm}
    \end{tabular}
    }
\label{tab:gsm_8k_ntl_fa_results}
\end{table}

\subsubsection{\la Results}\label{sec:embed_surgery_results}

\boldtitle{\la improves cross-lingual transfer.}
Figure~\ref{fig:gemma_results} shows the absolute gains of language adaptation on \gemmatwo models. When evaluating across three language groups, we observe that language adaptation consistently outperforms vanilla \gemmatwo models, demonstrating a significant performance advantage. The performance gains are particularly more pronounced on medium and low-resource languages, and this trend becomes increasingly apparent as model size scales. Similar findings are observed for \aya and \palmtwo, as presented in Appendix~\ref{appendix:extra_la_results}.

\begin{table*}[t]
\setlength{\belowcaptionskip}{-0.3cm}
\setlength{\tabcolsep}{6pt}
\footnotesize
\centering
\vspace{-2.5mm}
\caption{\ouradapter performance with zero-shot prompting. The best and second-best results are marked in \textbf{bold} and \underline{underlined}. \textcolor{red!80}{Red} values indicate instances where \ouradapter hurts the performance. English results are excluded when computing the average.}
\resizebox{\linewidth}{!}{
\begin{tabular}{l|c|ccc|ccc|ccc|cc|cc|c}
    \toprule
    \bf Task Type & \multirow{4}{*}[-1ex]{\rotatebox[origin=c]{90}{\bf \textsc{English}}} & \multicolumn{6}{c|}{\bf \textsc{Classification}} & \multicolumn{7}{c|}{\bf \textsc{Generation}} \\
    \cmidrule(lr){3-8} \cmidrule(lr){9-15}
    \multirow{2}{*}{\bf Eval. Metric} & & \multicolumn{3}{c|}{\bf \belebele} & \multicolumn{3}{c|}{\bf \sib} & \multicolumn{3}{c|}{\bf \flores} & \multicolumn{2}{c|}{\bf \textsc{XorQA-In}} & \multicolumn{2}{c|}{\bf \textsc{XSum-In}} & \multirow{4}{*}[2ex]{\bf Avg.} \\
    & & \multicolumn{3}{c|}{\bf \texttt{Accuracy}} & \multicolumn{3}{c|}{\bf \texttt{Accuracy}} & \multicolumn{3}{c|}{\bf \texttt{ChrF++}} & \multicolumn{2}{c|}{\bf \texttt{Token-F1}} & \multicolumn{2}{c|}{\bf \texttt{ChrF}} \\
    \cmidrule(lr){3-5} \cmidrule(lr){6-8} \cmidrule(lr){9-11} \cmidrule(lr){12-13} \cmidrule(lr){14-15}
    \bf Model & & \sea & \afr & \ind & \sea & \afr & \ind & \sea & \afr & \ind & \ind & \textsc{En} & \ind & \textsc{En} \\
    \midrule
    \gemmatwo-\texttt{2B-FLAN} & \bf 73.0 & 52.0 & 36.0 & 50.2 & 65.9 & 47.3 & 67.8 & 27.8 & 9.6 & 20.7 & 7.6 & 47.8 & 3.7 & \bf 36.6 & 31.7 \\ 
    \gemmatwo-\texttt{2B-FA} & \underline{69.8} & \underline{62.0} & \underline{40.2} & \underline{54.4} & \underline{72.9} & \bf 57.9 & \underline{70.5} & \textcolor{red!80}{27.0} & \underline{11.0} & \textcolor{red!80}{17.2} & \underline{8.6} & \underline{54.1} & \underline{6.6} & \underline{\textcolor{red!80}{34.1}} & \underline{35.3} \\
    {\pfix}  \texttt{LoRA-Adapt} & 69.7 & \bf 63.3 & \bf 43.7 & \bf 55.8 & \bf 73.5 & \underline{47.8} & \bf 70.7 & \bf 37.1 & \bf 18.3 & \bf 32.5 & \bf 9.8 & \bf 60.5 & \bf 9.9 & \textcolor{red!80}{31.9} & \bf 38.7 \\
    \midrule
    \gemmatwo-\texttt{9B-FLAN} & \bf 83.5 & 70.6 & 49.9 & 68.9 & 74.4 & 61.0 & 79.1 & 32.0 & 12.6 & 27.8 & 9.8 & 60.3 & 15.1 & \bf 37.5 & 42.0 \\ 
    \gemmatwo-\texttt{9B-FA} & 82.3 & \underline{78.1} & \underline{57.5} & \underline{71.2} & \bf 78.6 & \underline{68.8} & \underline{79.8} & \underline{35.9} & \underline{16.5} & \underline{31.4} & \bf 12.5 & \bf 65.2 & \underline{15.2} & \underline{\textcolor{red!80}{37.1}} & \underline{45.5} \\
    {\pfix}  \texttt{LoRA-Adapt} & \underline{82.5} & \bf 78.4 & \bf 60.2 & \bf 73.1 & \underline{78.5} & \bf 69.2 & \bf 80.1 & \bf 40.0 & \bf 25.6 & \bf 40.5 & \underline{12.1} & \underline{64.1} & \bf 17.4 & \textcolor{red!80}{37.0} & \bf 47.3 \\
    \midrule
    \gemmatwo-\texttt{27B-FLAN} & \bf 84.3 & 71.9 & 52.6 & 72.9 & 72.6 & 60.2 & 76.4 & 33.2 & \underline{15.2} & \underline{29.6} & \bf 20.4 & \underline{61.7} & \underline{15.0} & \bf 37.6 & \underline{43.7} \\ 
    \gemmatwo-\texttt{27B-FA} & \underline{83.6} & \underline{78.8} & \underline{56.3} & \underline{73.1} & \underline{78.1} & \underline{66.1} & \underline{78.5} & \underline{33.7} & \textcolor{red!80}{13.2} & \textcolor{red!80}{23.9} & \textcolor{red!80}{15.2} & \textcolor{red!80}{59.3} & \textcolor{red!80}{12.4} & \textcolor{red!80}{36.8} & \textcolor{red!80}{43.5} \\
    {\pfix}  \texttt{LoRA-Adapt} & 83.4 & \bf 79.5 & \bf 60.4 & \bf 74.3 & \bf 79.5 & \bf 68.3 & \bf 80.0 & \bf 42.5 & \bf 25.7 & \bf 40.6 & \underline{\textcolor{red!80}{16.4}} & \bf 70.1 & \bf 18.0 & \underline{\textcolor{red!80}{36.9}} & \bf 49.1 \\
    \bottomrule
    \multicolumn{13}{c}{}\vspace{-9mm}
    \end{tabular}
    }
\label{tab:franken_adapter_results}
\end{table*}




\boldtitle{\la helps preserve LLMs' general knowledge.}
We evaluate whether language adaptation improves the transfer of English reasoning in LLMs to other languages. Table~\ref{tab:gsm_8k_ntl_results} shows LLMs with language adaptation consistently improves the mathematical reasoning ability across various low-resource languages. Compared to CALM~\citep{bansal2024llm}, a form of adapter enabling model composition, our method achieves more substantial improvements over \palmtwo-\texttt{S} with only embedding tuning and incurring no extra inference costs.\footnote{Our approach, however, requires more training as embedding tuning must be repeated for different linguistic regions.} Moreover, our approach shows superior performance (+4\%) compared to \palmtwo-\texttt{S-NTL} that was full-parameter tuned on NTL.\footnote{Both methods use the same training dataset. However, NTL training did not create region-specific models by splitting training data, potentially diminishing its effectiveness due to the curse of multilinguality~\citep{conneau-etal-2020-unsupervised}, which arises when a single model is trained on too many languages.} Overall, our results suggest that language adaptation offers an effective alternative to CALM and full-parameter tuning. We observe performance improvements with more modern models, with particularly pronounced gains in \gemmatwo.

\subsubsection{\ouradapter Results}\label{sec:franken_adapter_results}

\boldtitle{\ouradapter enhances mathematical reasoning performance in target languages on the fly.}
In Table~\ref{tab:gsm_8k_ntl_fa_results}, we show that \ouradapter consistently outperforms the instruction-tuned baseline for zero-shot evaluation on mathematical reasoning tasks, with an average improvement of up to 8\% across 20 low-resource languages. Moreover, \ouradapter further advances performance by incorporating LLMs with enhanced mathematical reasoning ability (\eg the $\texttt{IT}$ models aligned via reinforcement learning).\footnote{We observe LLMs generate Chain-of-Thought steps more often in English than in the question language, and we hypothesize that the customized embeddings in \ouradapter enable the model to better comprehend the questions.}

\ouradapter can also enhance the cross-lingual performance of LLMs instruction-tuned for reasoning. During the instruction-tuning stage (Figure~\ref{fig:franken_adapter} step 2(a)), we employ a mathematical instruction dataset, \texttt{WebInstruct}~\citep{yue2024mammoth}, for model training, and obtain the composed model by reusing the embeddings from the language adaptation stage. Table~\ref{tab:gsm_8k_ntl_fa_results} shows \ouradapter consistently improves performance in target languages. These results underscore its versatility by facilitating cross-lingual transfer in models post-trained for a specific domain.

\boldtitle{\ouradapter is more effective for classification tasks.}
As shown in Table~\ref{tab:franken_adapter_results}, \ouradapter consistently improves the performance on classification tasks across all model sizes with up to 10\% absolute gains. By contrast, for the generation tasks, the method's behavior is inconsistent, often leading to performance degradation. We attribute this phenomenon to intrinsic difficulty: classification tasks are generally easier as the solution space is typically small compared to generation tasks, making them more robust to embedding changes. However, the embedding layer is used for both text encoding and decoding in generation, and the auto-regressive generation paradigm makes the model sensitive to embedding changes due to error propagation accumulating over time steps.

\boldtitle{LoRA-Adaptation connects both worlds.}
In Table~\ref{tab:franken_adapter_results}, we show that by applying cost-effective LoRA-Adaptation, the two components within the \ouradapter can cooperate better, leading to significant gains on generation tasks. This results in an average improvement of 5.4\% on the largest 27B variant. The results demonstrate the practical use of our \ouradapter method for efficient zero-shot cross-lingual transfer of instruction-tuned LLMs. See Appendix~\ref{appendix:extra_fa_results} and Table~\ref{tab:franken_adapter_results_other_models} for additional results on \aya and \palmtwo. Furthermore, these improvements come at minimal cost on English proficiency (See Appendix~\ref{appendix:english_tasks}) and the same findings are observed for \aya \ \etc (See Table~\ref{tab:franken_adapter_results_other_models})

\subsection{Ablation Analysis}

\begin{figure}[t]
    \setlength{\abovecaptionskip}{-0.0001cm}
    \setlength{\belowcaptionskip}{-0.35cm}
    \centering
    \includegraphics[width=\linewidth]{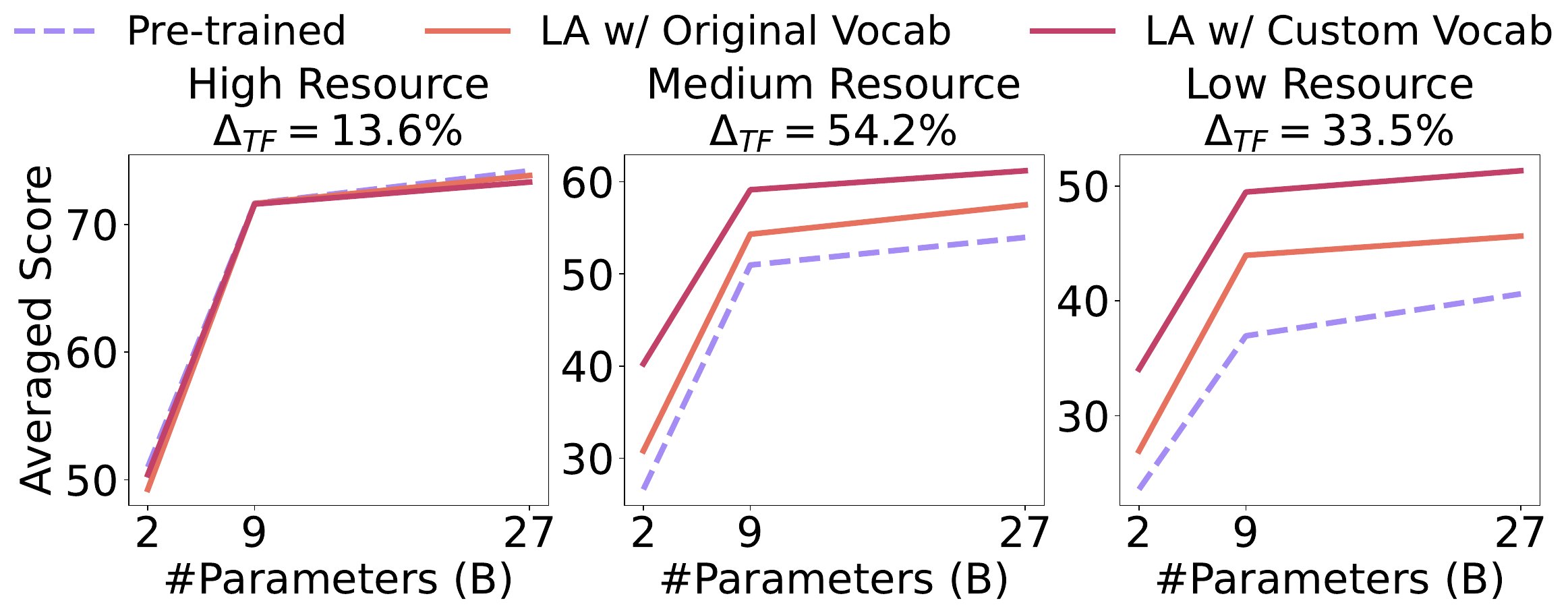}
    \vspace{-8mm}
    \caption{Ablations on tokenizers for language adaptation. Macro-averaged scores on \sea subset of \flores and \belebele are reported. $\Delta_{TF}$: \% tokenizer fertility reduction.}
    \vspace{-6mm}
    \label{fig:effects_customized_vocabulary}
\end{figure}

\begin{figure}[t]
    \setlength{\abovecaptionskip}{-0.0001cm}
    \setlength{\belowcaptionskip}{-0.35cm}
    \centering
    \includegraphics[width=\linewidth]{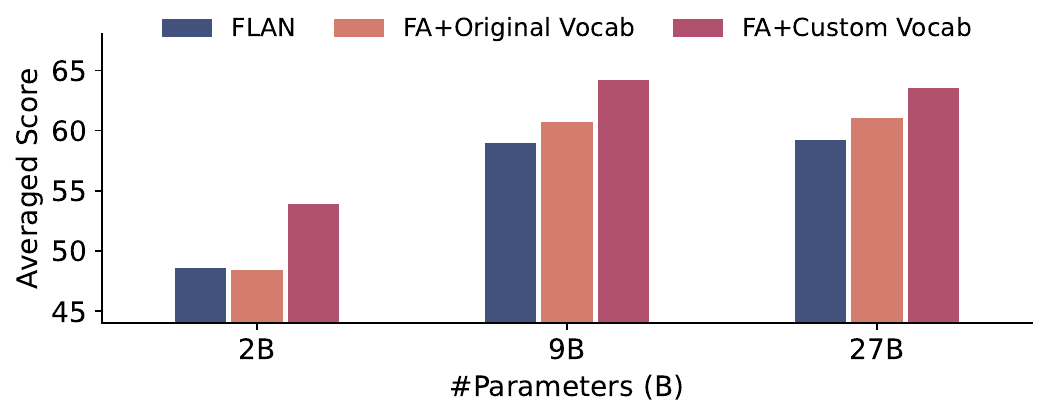}
    \vspace{-9.5mm}
    \caption{\ouradapter result comparison when using original and customized tokenizers. The macro-averaged scores on \sea subset of \belebele, \sib and \flores are reported. }
    \vspace{-4mm}
    \label{fig:effects_customized_vocabulary_franken_adapter}
\end{figure}

\boldtitle{Customized vocabulary amplifies the benefits of training on multilingual data.}
We decouple the effects of multilingual data in language adaptation from the change in vocabularies.
Figure~\ref{fig:effects_customized_vocabulary} shows that simple continued training on additional multilingual data with embedding tuning can significantly improve the performance in medium and low-resource languages, indicating LLMs are under-fitting to these languages. Based on this, employing a customized vocabulary that is fairer to these languages (Figure~\ref{fig:fertility_comparison}) can amplify the benefits of embedding tuning on multilingual data. This enhanced learning process facilitates better knowledge acquisition~\citep{zhang-etal-2022-robust,hofmann-etal-2022-embarrassingly}. Moreover, the improvements are more pronounced in smaller models, highlighting the importance of effective tokenization for these models to adapt well to low-resource languages. 
In addition, Figure~\ref{fig:effects_customized_vocabulary_franken_adapter} indicates that the importance of customized vocabulary is also apparent in the \ouradapter setting.

\begin{table}[t]
\setlength{\belowcaptionskip}{-0.2cm}
\setlength{\tabcolsep}{7pt}
\centering
\vspace{-2mm}
\caption{Comparing latency for using original and customized vocabularies in \ouradapter. The number of instances processed per second (\ie prefilling) by \gemmatwo are reported. We use the passages in \belebele for all \sea, \afr, and \ind languages.}
\resizebox{\columnwidth}{!}{
\begin{tabular}{l|l|lll}
    \toprule
    \bf Size & \bf Vocab & \bf\textsc{Hrl} & \bf\textsc{Mrl} & \bf\textsc{Lrl} \\ 
    \midrule
    \multirow{2}{*}{2B} & Original & 59.1 & 55.6 & 57.1 \\
    & Custom & 61.1 {\tiny\uagray{3.38\%}} & 61.2 {\tiny\uagray{10.07\%}} & 60.3 {\tiny\uagray{5.60\%}} \\
    \midrule

    \multirow{2}{*}{9B} & Original & 30.2 & 27.2 & 28.6 \\
    & Custom & 33.8 {\tiny\uagray{11.92\%}} & 33.3 {\tiny\uagray{22.43\%}} & 32.1 {\tiny\uagray{12.24\%}} \\
    \midrule

    \multirow{2}{*}{27B} & Original & 17.1 & 15.0 & 16.0 \\
    & Custom & 19.6 {\tiny\uagray{14.62\%}} & 19.2 {\tiny\uagray{28.0\%}} & 18.4 {\tiny\uagray{15.0\%}} \\
    \bottomrule
    \multicolumn{5}{c}{}\vspace{-9mm}
    \end{tabular}}
    
\label{tab:latency_compare}
\end{table}
\boldtitle{Customized vocabulary improves latency.}
We evaluate latency by measuring the number of texts processed per second by LLMs. We use the passages from all \sea, \afr, and \ind languages in \belebele as test texts. For comparison, we test \ouradapter models integrated with embeddings trained using either the original \gemmatwo tokenizer or our customized one. Table~\ref{tab:latency_compare} shows that employing customized tokenizer consistently improves latency, particularly in MRLs and LRLs. This trend becomes increasingly pronounced as model size scales, highlighting the importance of customized tokenizers in achieving low-latency processing for long-tail languages in larger LLMs.

\begin{figure}[t]
    \setlength{\abovecaptionskip}{-0.0001cm}
    \setlength{\belowcaptionskip}{-0.35cm}
    \centering
    \includegraphics[width=\linewidth]{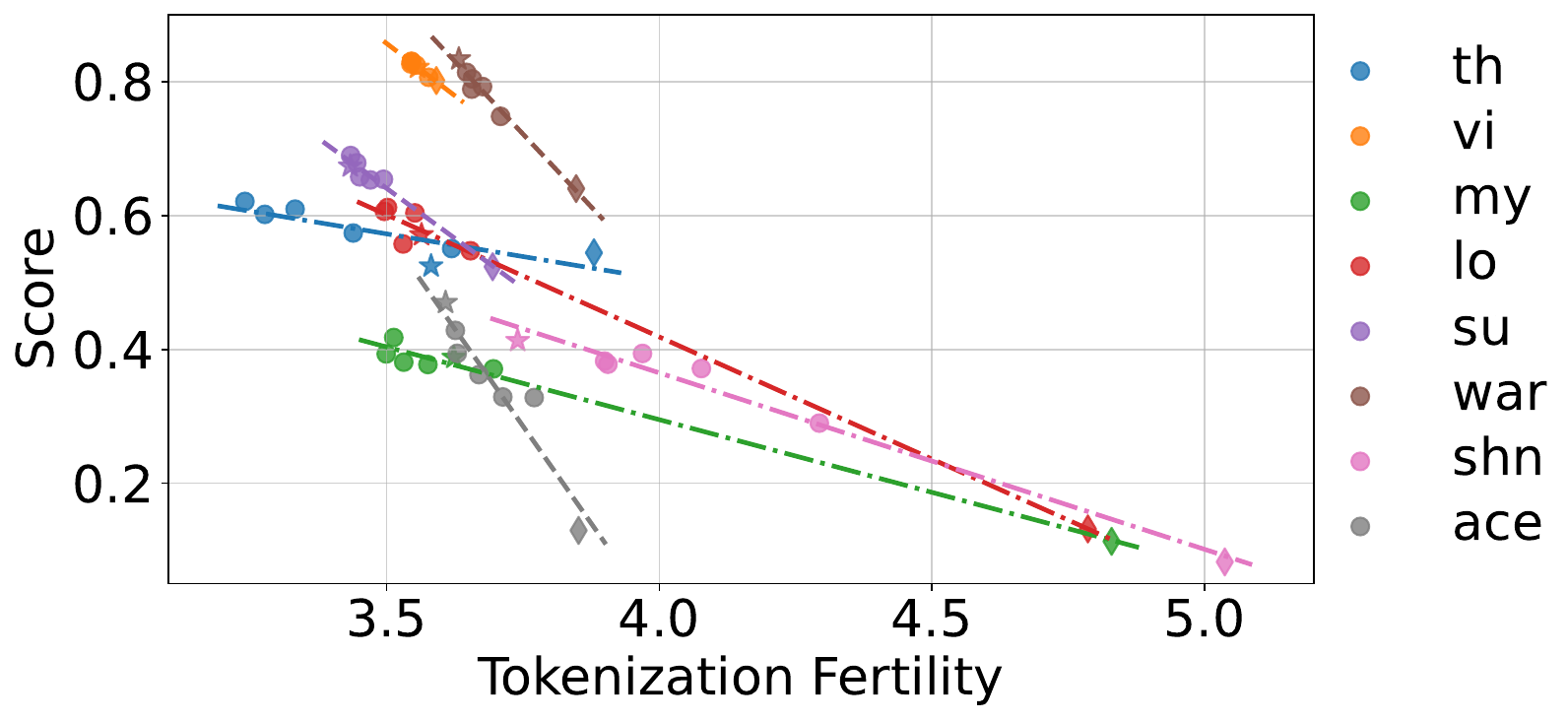}
    \vspace{-8.5mm}
    \caption{Correlation between the results of language adaptation on \gemmatwo-\texttt{2B} with tokenizer fertility. Normalized \texttt{ChrF++} on \textsc{Flores}-\sea is reported. {$\bLozenge$} and {\pfix}  indicate the original and our customized tokenizers used in the other settings, respectively.}
    \vspace{-5.5mm}
    \label{fig:fertility_vs_performance}
\end{figure}

\boldtitle{Tokenization fertility is inversely correlated to downstream performance.}
We study how tokenization fertility (the average number of tokens produced per word) affects the LLM's performance across languages. 
To ablate this effect, we generate several re-sampled replicates of our tokenizer training datasets with different levels of priority given to high \versus lower resource languages. Specifically, we use temperature sampling to manipulate the training sentences of each language for building different tokenizer training data, and train tokenizers with varying fertilities. We then do language adaptation for each of these tokenizers and analyze the downstream performance. As shown in Figure~\ref{fig:fertility_vs_performance}, we find that the performance is inversely correlated to tokenization fertility, but the correlation is not uniform across languages. Notably, slight reductions in fertility lead to significant performance improvements in low-resource languages (\eg \textsc{Ace}) while high-resource languages are less sensitive to fertility changes.
Furthermore, Latin-script languages generally benefit more from fertility reductions compared to those in non-Latin scripts.
Please refer to Appendix Figure~\ref{fig:fertility_vs_performance_palm2} which reports similar findings for \palmtwo-\texttt{XXS}.

\begin{figure}[t]
    \setlength{\abovecaptionskip}{-0.0001cm}
    \setlength{\belowcaptionskip}{-0.35cm}
    \centering
    \includegraphics[width=\linewidth]{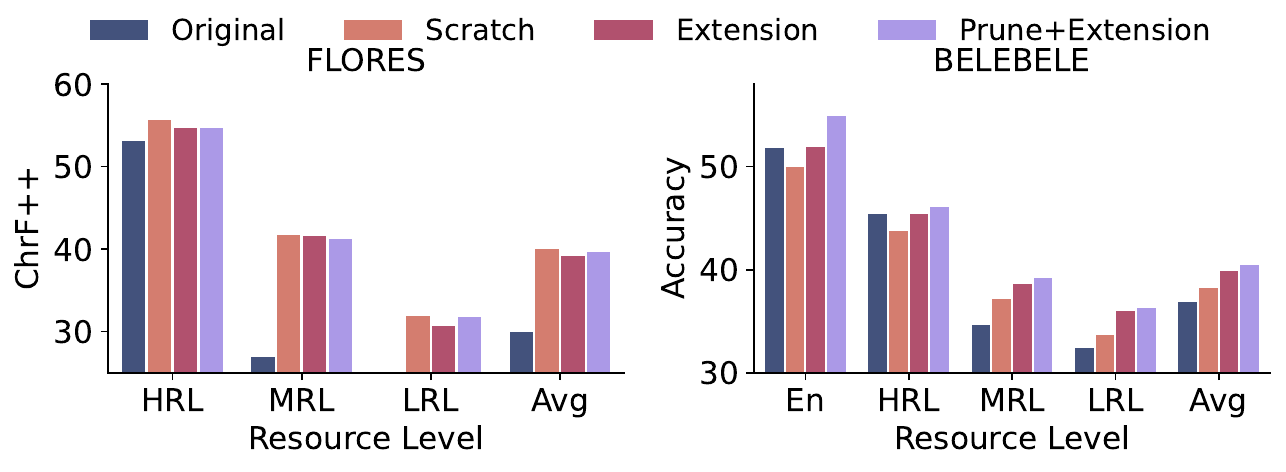}
    \vspace{-9mm}
    \caption{Ablations on tokenizer building methods. We report \sea language adaptation on \gemmatwo-\texttt{2B}.}
    \vspace{-4.5mm}
    \label{fig:ablation_on_tokenizer_building}
\end{figure}

\boldtitle{Pruning with extension outperforms other tokenizer construction methods.}
We study the impact of tokenizer construction methods on the performance of language adaptation. We compare our \emph{Prune-with-Extension} method (\S\ref{sec:vocab_construct}) against two variants: 1) \emph{Scratch}, which trains tokenizers from scratch on English and target languages; and 2) \emph{Extension}, which appends target language tokens to existing tokenizers without pruning. This adds an additional of 34\% embedding parameters. Figure~\ref{fig:ablation_on_tokenizer_building} shows our \emph{Prune+Extension} method achieves the best overall results. Notably, the performance differences on \textsc{Flores} are minimal across the methods. We attribute this to the relatively lower intrinsic complexity of \textsc{Flores} compared to \belebele that demands reasoning ability most likely transferred from English. This claim is supported by the substantial improvements in English on \belebele when switching to our approach that preserves original English tokens. Moreover, we observe that tokenizer built from scratch exhibits significantly fewer overlapping English tokens with the original tokenizer compared to the other methods. This signifies the importance of retaining English tokens to preserve pre-training knowledge, which is vital for the success of language adaptation.\footnote{It's an open question of whether other facets of the tokenizer need to be retained to preserve other behaviours, e.g., markup tokens to facilitate code understanding.} Our method also surpasses \emph{Extension}, indicating that removing irrelevant tokens is beneficial while avoid introducing extra parameters.\footnote{We suspect large vocabulary could increase ambiguity in output $\operatorname{softmax}$. Evidence also reveals that large vocabularies are not optimal for smaller LLMs~\citep{tao2024scaling}.}

\begin{figure}[t]
    \setlength{\abovecaptionskip}{-0.0001cm}
    \setlength{\belowcaptionskip}{-0.35cm}
    \centering
    \includegraphics[width=\linewidth]{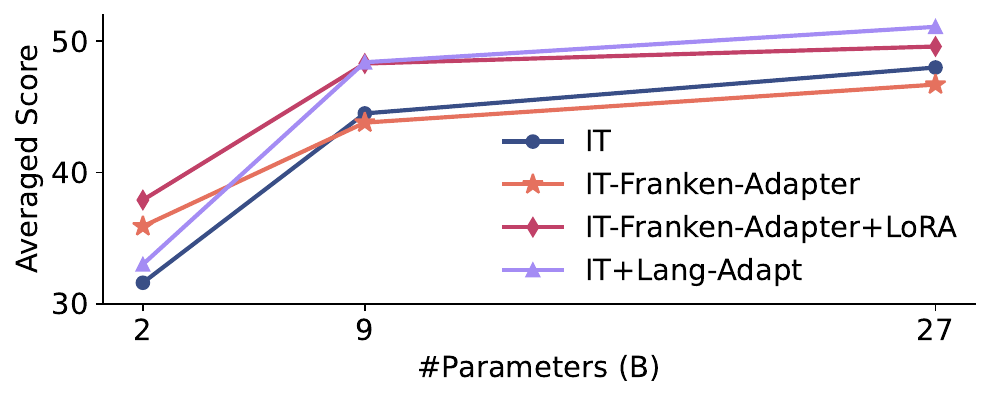}
    \vspace{-9mm}
    \caption{Comparing applying language adaptation and \ouradapter on \gemmatwo-\texttt{IT} models. The averaged scores in five evaluation benchmarks (Table~\ref{tab:franken_adapter_results}) are reported.}
    \vspace{-6mm}
    \label{fig:franken_adapter_vs_emb_surgery_on_it}
\end{figure}

\boldtitle{\la on IT LLMs represents a viable alternative to \ouradapter .}
LLMs are typically released in both pre-trained (\textsc{Pt}) and instruct (\textsc{It}) versions. These versions provide two distinct pathways for achieving zero-shot cross lingual transfer: 1) employing our \ouradapter pipeline, which uses the \textsc{Pt} model for language adaptation and subsequently integrates the new embeddings into the \textsc{It} model through LoRA-Adaptation; 2) employing the \textsc{It}+Lang-Adapt approach, which performs one-step language adaptation directly on \textsc{It} model with the same training costs as on \textsc{Pt} model. Figure~\ref{fig:franken_adapter_vs_emb_surgery_on_it} shows a comparison of these two strategies. Our analysis reveals that for smaller model sizes, \ouradapter outperforms \textsc{It}+Lang-Adapt in downstream tasks. However, as model size increases, \textsc{It}+Lang-Adapt emerges as the more effective solution. 
Overall, the \textsc{It}+Lang-Adapt method provides a highly effective alternative to \ouradapter for zero-shot cross-lingual transfer involving a single model. However, when dealing with multiple \textsc{It} models derived from a common \textsc{Pt} model, \ouradapter offers greater modularity by avoiding costly embedding tuning across individual models.
\section{Related Work}

\inlinetitle{Language Adaptation of LLMs.}
Progress in LLMs have been dominated by English and other high-resource languages.
A growing body of research has shown progress in extending the applicability of LLMs to a broader range of languages. These methods typically involve continued pre-training on additional monolingual data~\citep{cui2024efficienteffectivetextencoding,zhao2024llamaenglishempiricalstudy} or multilingual instruction-tuning on synthetic data ~\citep{chen-etal-2024-monolingual,ustun-etal-2024-aya,aryabumi2024aya}. Notably, such efforts primarily focus on monolingual transfer or the multilingual adaptation of a few high-resource languages. Furthermore, these techniques often rely on computationally expensive full-parameter tuning, a strategy that tends to cause adapted LLMs to lose previously acquired knowledge due to catastrophic forgetting. \citet{bansal2024llm} augments an LLM with a smaller multilingual-specialized LLM by introducing a limited set of trainable parameters, yielding improved language adaptation while retaining general pre-trained knowledge. In contrast, we demonstrate that embedding tuning with additional multilingual data is sufficient for adapting LLMs to massive languages and creating customized vocabularies tailored to these languages can further enhance cross-lingual transfer.

\inlinetitle{Efficient Methods for Cross-Lingual Transfer.}
Diverse approaches have been proposed to efficiently adapt pre-trained LMs to new languages. These methods include using adapters~\citep{pfeiffer-etal-2020-mad,pfeiffer-etal-2022-lifting,yong-etal-2023-bloom,liu2022fewshot} and sparse fine-tuning~\citep{ansell-etal-2022-composable}. While effective, such parameter-efficient methods have been shown to be less effective in acquiring new knowledge~\citep{biderman2024lora}. Embedding surgery, on the other hand, has proven highly effective for adapting PLMs between languages while keeping the transformer body untouched ~\citep{artetxe-etal-2020-cross,de-vries-nissim-2021-good,marchisio-etal-2023-mini,chen2023improving}. However, its application has been limited to encoder-only PLMs and constrained by monolingual transfer, where one embedding is created per target language. Our \ouradapter approach builds upon this idea but differs significantly in that: we demonstrate its effectiveness with more powerful decoder-only LLMs for much wider range of tasks requiring different skill sets, and extend its capability to support massive cross-lingual transfer, enabling adaptation to a broader range of long-tail languages with a single embedding.


\section{Conclusion}
In this work, we introduce \ouradapter, a novel approach that facilitates efficient zero-shot cross-lingual transfer across a wide range of languages with embedding surgery. Our empirical findings reveal that embedding tuning on multilingual data suffices for the effective language adaptation of LLMs. Moreover, this adaptation can be further optimized through the implementation of customized vocabularies tailored to target languages. We also highlight the adaptability of these embeddings, which can be integrated into any instruction-tuned LLMs to enable cross-lingual transfer with minimal training costs.

\section*{Impact Statement}
This work presents novel insights into the language adaptation and cross-lingual transfer of LLMs, which have the potential to change the ways in which LLMs are trained and adapted to new languages. We believe \ouradapter can serve as a versatile modular framework to expand the capabilities of LLMs beyond language, such as domain adaptations for enhanced coding and mathematical skills. There might also be safety concerns arising from the disruption of alignment abilities due to the changes to the embeddings, and how to align the model effectively and efficiently to reject unsafe queries after the \ouradapter pipeline which would need to be carefully addressed.

\section*{Acknowledgment}
The authors would thank Isaac Caswell for the valuable comments and feedback. The authors would also thank Daniel Formoso for helping support the research.

\bibliography{anthology,custom}

\begin{thebibliography}{66}
\providecommand{\natexlab}[1]{#1}
\providecommand{\url}[1]{\texttt{#1}}
\expandafter\ifx\csname urlstyle\endcsname\relax
  \providecommand{\doi}[1]{doi: #1}\else
  \providecommand{\doi}{doi: \begingroup \urlstyle{rm}\Url}\fi

\bibitem[Adelani et~al.(2024)Adelani, Liu, Shen, Vassilyev, Alabi, Mao, Gao, and Lee]{adelani-etal-2024-sib}
Adelani, D., Liu, H., Shen, X., Vassilyev, N., Alabi, J., Mao, Y., Gao, H., and Lee, E.-S.
\newblock {SIB}-200: A simple, inclusive, and big evaluation dataset for topic classification in 200+ languages and dialects.
\newblock In Graham, Y. and Purver, M. (eds.), \emph{Proceedings of the 18th Conference of the European Chapter of the Association for Computational Linguistics (Volume 1: Long Papers)}, pp.\  226--245, St. Julian{'}s, Malta, March 2024. Association for Computational Linguistics.
\newblock URL \url{https://aclanthology.org/2024.eacl-long.14}.

\bibitem[Ahia et~al.(2023)Ahia, Kumar, Gonen, Kasai, Mortensen, Smith, and Tsvetkov]{ahia-etal-2023-languages}
Ahia, O., Kumar, S., Gonen, H., Kasai, J., Mortensen, D., Smith, N., and Tsvetkov, Y.
\newblock Do all languages cost the same? tokenization in the era of commercial language models.
\newblock In Bouamor, H., Pino, J., and Bali, K. (eds.), \emph{Proceedings of the 2023 Conference on Empirical Methods in Natural Language Processing}, pp.\  9904--9923, Singapore, December 2023. Association for Computational Linguistics.
\newblock \doi{10.18653/v1/2023.emnlp-main.614}.
\newblock URL \url{https://aclanthology.org/2023.emnlp-main.614}.

\bibitem[Ahuja et~al.(2023)Ahuja, Diddee, Hada, Ochieng, Ramesh, Jain, Nambi, Ganu, Segal, Ahmed, Bali, and Sitaram]{ahuja-etal-2023-mega}
Ahuja, K., Diddee, H., Hada, R., Ochieng, M., Ramesh, K., Jain, P., Nambi, A., Ganu, T., Segal, S., Ahmed, M., Bali, K., and Sitaram, S.
\newblock {MEGA}: Multilingual evaluation of generative {AI}.
\newblock In Bouamor, H., Pino, J., and Bali, K. (eds.), \emph{Proceedings of the 2023 Conference on Empirical Methods in Natural Language Processing}, pp.\  4232--4267, Singapore, December 2023. Association for Computational Linguistics.
\newblock \doi{10.18653/v1/2023.emnlp-main.258}.
\newblock URL \url{https://aclanthology.org/2023.emnlp-main.258}.

\bibitem[Anil et~al.(2023)]{anil2023palm}
Anil, R. et~al.
\newblock Palm 2 technical report, 2023.

\bibitem[Anil et~al.(2024)]{geminiteam2024geminifamilyhighlycapable}
Anil, R. et~al.
\newblock Gemini: A family of highly capable multimodal models, 2024.
\newblock URL \url{https://arxiv.org/abs/2312.11805}.

\bibitem[Ansell et~al.(2022)Ansell, Ponti, Korhonen, and Vuli{\'c}]{ansell-etal-2022-composable}
Ansell, A., Ponti, E., Korhonen, A., and Vuli{\'c}, I.
\newblock Composable sparse fine-tuning for cross-lingual transfer.
\newblock In \emph{Proceedings of the 60th Annual Meeting of the Association for Computational Linguistics (Volume 1: Long Papers)}, pp.\  1778--1796, Dublin, Ireland, May 2022. Association for Computational Linguistics.
\newblock \doi{10.18653/v1/2022.acl-long.125}.
\newblock URL \url{https://aclanthology.org/2022.acl-long.125}.

\bibitem[Artetxe et~al.(2020)Artetxe, Ruder, and Yogatama]{artetxe-etal-2020-cross}
Artetxe, M., Ruder, S., and Yogatama, D.
\newblock On the cross-lingual transferability of monolingual representations.
\newblock In \emph{Proceedings of the 58th Annual Meeting of the Association for Computational Linguistics}, pp.\  4623--4637, Online, July 2020. Association for Computational Linguistics.
\newblock \doi{10.18653/v1/2020.acl-main.421}.
\newblock URL \url{https://aclanthology.org/2020.acl-main.421}.

\bibitem[Aryabumi et~al.(2024)Aryabumi, Dang, Talupuru, Dash, Cairuz, Lin, Venkitesh, Smith, Campos, Tan, Marchisio, Bartolo, Ruder, Locatelli, Kreutzer, Frosst, Gomez, Blunsom, Fadaee, Üstün, and Hooker]{aryabumi2024aya}
Aryabumi, V., Dang, J., Talupuru, D., Dash, S., Cairuz, D., Lin, H., Venkitesh, B., Smith, M., Campos, J.~A., Tan, Y.~C., Marchisio, K., Bartolo, M., Ruder, S., Locatelli, A., Kreutzer, J., Frosst, N., Gomez, A., Blunsom, P., Fadaee, M., Üstün, A., and Hooker, S.
\newblock Aya 23: Open weight releases to further multilingual progress, 2024.

\bibitem[Bandarkar et~al.(2024)Bandarkar, Liang, Muller, Artetxe, Shukla, Husa, Goyal, Krishnan, Zettlemoyer, and Khabsa]{bandarkar-etal-2024-belebele}
Bandarkar, L., Liang, D., Muller, B., Artetxe, M., Shukla, S.~N., Husa, D., Goyal, N., Krishnan, A., Zettlemoyer, L., and Khabsa, M.
\newblock The belebele benchmark: a parallel reading comprehension dataset in 122 language variants.
\newblock In Ku, L.-W., Martins, A., and Srikumar, V. (eds.), \emph{Proceedings of the 62nd Annual Meeting of the Association for Computational Linguistics (Volume 1: Long Papers)}, pp.\  749--775, Bangkok, Thailand, August 2024. Association for Computational Linguistics.
\newblock \doi{10.18653/v1/2024.acl-long.44}.
\newblock URL \url{https://aclanthology.org/2024.acl-long.44}.

\bibitem[Bansal et~al.(2024)Bansal, Samanta, Dalmia, Gupta, Ganapathy, Bapna, Jain, and Talukdar]{bansal2024llm}
Bansal, R., Samanta, B., Dalmia, S., Gupta, N., Ganapathy, S., Bapna, A., Jain, P., and Talukdar, P.
\newblock {LLM} augmented {LLM}s: Expanding capabilities through composition.
\newblock In \emph{The Twelfth International Conference on Learning Representations}, 2024.
\newblock URL \url{https://openreview.net/forum?id=jjA4O1vJRz}.

\bibitem[Bapna et~al.(2022)Bapna, Caswell, Kreutzer, Firat, van Esch, Siddhant, Niu, Baljekar, Garcia, Macherey, Breiner, Axelrod, Riesa, Cao, Chen, Macherey, Krikun, Wang, Gutkin, Shah, Huang, Chen, Wu, and Hughes]{bapna2022building}
Bapna, A., Caswell, I., Kreutzer, J., Firat, O., van Esch, D., Siddhant, A., Niu, M., Baljekar, P., Garcia, X., Macherey, W., Breiner, T., Axelrod, V., Riesa, J., Cao, Y., Chen, M.~X., Macherey, K., Krikun, M., Wang, P., Gutkin, A., Shah, A., Huang, Y., Chen, Z., Wu, Y., and Hughes, M.
\newblock Building machine translation systems for the next thousand languages, 2022.

\bibitem[Biderman et~al.(2024)Biderman, Portes, Ortiz, Paul, Greengard, Jennings, King, Havens, Chiley, Frankle, Blakeney, and Cunningham]{biderman2024lora}
Biderman, D., Portes, J., Ortiz, J. J.~G., Paul, M., Greengard, P., Jennings, C., King, D., Havens, S., Chiley, V., Frankle, J., Blakeney, C., and Cunningham, J.~P.
\newblock Lo{RA} learns less and forgets less.
\newblock \emph{Transactions on Machine Learning Research}, 2024.
\newblock ISSN 2835-8856.
\newblock URL \url{https://openreview.net/forum?id=aloEru2qCG}.
\newblock Featured Certification.

\bibitem[Bisk et~al.(2019)Bisk, Zellers, Bras, Gao, and Choi]{bisk2019piqa}
Bisk, Y., Zellers, R., Bras, R.~L., Gao, J., and Choi, Y.
\newblock Piqa: Reasoning about physical commonsense in natural language, 2019.

\bibitem[Brown et~al.(2020)Brown, Mann, Ryder, Subbiah, Kaplan, Dhariwal, Neelakantan, Shyam, Sastry, Askell, Agarwal, Herbert-Voss, Krueger, Henighan, Child, Ramesh, Ziegler, Wu, Winter, Hesse, Chen, Sigler, Litwin, Gray, Chess, Clark, Berner, McCandlish, Radford, Sutskever, and Amodei]{gpt3}
Brown, T., Mann, B., Ryder, N., Subbiah, M., Kaplan, J.~D., Dhariwal, P., Neelakantan, A., Shyam, P., Sastry, G., Askell, A., Agarwal, S., Herbert-Voss, A., Krueger, G., Henighan, T., Child, R., Ramesh, A., Ziegler, D., Wu, J., Winter, C., Hesse, C., Chen, M., Sigler, E., Litwin, M., Gray, S., Chess, B., Clark, J., Berner, C., McCandlish, S., Radford, A., Sutskever, I., and Amodei, D.
\newblock Language models are few-shot learners.
\newblock In Larochelle, H., Ranzato, M., Hadsell, R., Balcan, M., and Lin, H. (eds.), \emph{Advances in Neural Information Processing Systems}, volume~33, pp.\  1877--1901. Curran Associates, Inc., 2020.
\newblock URL \url{https://proceedings.neurips.cc/paper_files/paper/2020/file/1457c0d6bfcb4967418bfb8ac142f64a-Paper.pdf}.

\bibitem[Caswell et~al.(2020)Caswell, Breiner, van Esch, and Bapna]{caswell-etal-2020-language}
Caswell, I., Breiner, T., van Esch, D., and Bapna, A.
\newblock Language {ID} in the wild: Unexpected challenges on the path to a thousand-language web text corpus.
\newblock In \emph{Proceedings of the 28th International Conference on Computational Linguistics}, pp.\  6588--6608, Barcelona, Spain (Online), December 2020. International Committee on Computational Linguistics.
\newblock \doi{10.18653/v1/2020.coling-main.579}.
\newblock URL \url{https://aclanthology.org/2020.coling-main.579}.

\bibitem[Chen et~al.(2024)Chen, Ji, Bogoychev, Kutuzov, Haddow, and Heafield]{chen-etal-2024-monolingual}
Chen, P., Ji, S., Bogoychev, N., Kutuzov, A., Haddow, B., and Heafield, K.
\newblock Monolingual or multilingual instruction tuning: Which makes a better alpaca.
\newblock In Graham, Y. and Purver, M. (eds.), \emph{Findings of the Association for Computational Linguistics: EACL 2024}, pp.\  1347--1356, St. Julian{'}s, Malta, March 2024. Association for Computational Linguistics.
\newblock URL \url{https://aclanthology.org/2024.findings-eacl.90}.

\bibitem[Chen et~al.(2023)Chen, Marchisio, Raileanu, Adelani, Stenetorp, Riedel, and Artetxe]{chen2023improving}
Chen, Y., Marchisio, K., Raileanu, R., Adelani, D.~I., Stenetorp, P., Riedel, S., and Artetxe, M.
\newblock Improving language plasticity via pretraining with active forgetting.
\newblock In \emph{Thirty-seventh Conference on Neural Information Processing Systems}, 2023.
\newblock URL \url{https://openreview.net/forum?id=jvEbQBxd8X}.

\bibitem[Chung et~al.(2023)Chung, Garcia, Roberts, Tay, Firat, Narang, and Constant]{chung2023unimax}
Chung, H.~W., Garcia, X., Roberts, A., Tay, Y., Firat, O., Narang, S., and Constant, N.
\newblock Unimax: Fairer and more effective language sampling for large-scale multilingual pretraining.
\newblock In \emph{The Eleventh International Conference on Learning Representations}, 2023.
\newblock URL \url{https://openreview.net/forum?id=kXwdL1cWOAi}.

\bibitem[Clark et~al.(2019)Clark, Lee, Chang, Kwiatkowski, Collins, and Toutanova]{clark-etal-2019-boolq}
Clark, C., Lee, K., Chang, M.-W., Kwiatkowski, T., Collins, M., and Toutanova, K.
\newblock {B}ool{Q}: Exploring the surprising difficulty of natural yes/no questions.
\newblock In \emph{Proceedings of the 2019 Conference of the North {A}merican Chapter of the Association for Computational Linguistics: Human Language Technologies, Volume 1 (Long and Short Papers)}, pp.\  2924--2936, Minneapolis, Minnesota, June 2019. Association for Computational Linguistics.
\newblock \doi{10.18653/v1/N19-1300}.
\newblock URL \url{https://aclanthology.org/N19-1300}.

\bibitem[Clark et~al.(2018)Clark, Cowhey, Etzioni, Khot, Sabharwal, Schoenick, and Tafjord]{clark2018thinksolvedquestionanswering}
Clark, P., Cowhey, I., Etzioni, O., Khot, T., Sabharwal, A., Schoenick, C., and Tafjord, O.
\newblock Think you have solved question answering? try arc, the ai2 reasoning challenge, 2018.
\newblock URL \url{https://arxiv.org/abs/1803.05457}.

\bibitem[Cobbe et~al.(2021)Cobbe, Kosaraju, Bavarian, Chen, Jun, Kaiser, Plappert, Tworek, Hilton, Nakano, Hesse, and Schulman]{cobbe2021trainingverifierssolvemath}
Cobbe, K., Kosaraju, V., Bavarian, M., Chen, M., Jun, H., Kaiser, L., Plappert, M., Tworek, J., Hilton, J., Nakano, R., Hesse, C., and Schulman, J.
\newblock Training verifiers to solve math word problems, 2021.
\newblock URL \url{https://arxiv.org/abs/2110.14168}.

\bibitem[Conneau et~al.(2020)Conneau, Khandelwal, Goyal, Chaudhary, Wenzek, Guzm{\'a}n, Grave, Ott, Zettlemoyer, and Stoyanov]{conneau-etal-2020-unsupervised}
Conneau, A., Khandelwal, K., Goyal, N., Chaudhary, V., Wenzek, G., Guzm{\'a}n, F., Grave, E., Ott, M., Zettlemoyer, L., and Stoyanov, V.
\newblock Unsupervised cross-lingual representation learning at scale.
\newblock In \emph{Proceedings of the 58th Annual Meeting of the Association for Computational Linguistics}, pp.\  8440--8451, Online, July 2020. Association for Computational Linguistics.
\newblock \doi{10.18653/v1/2020.acl-main.747}.
\newblock URL \url{https://aclanthology.org/2020.acl-main.747}.

\bibitem[Costa-jussà et~al.(2022)]{nllbteam2022languageleftbehindscaling}
Costa-jussà, M.~R. et~al.
\newblock No language left behind: Scaling human-centered machine translation, 2022.
\newblock URL \url{https://arxiv.org/abs/2207.04672}.

\bibitem[Cui et~al.(2024)Cui, Yang, and Yao]{cui2024efficienteffectivetextencoding}
Cui, Y., Yang, Z., and Yao, X.
\newblock Efficient and effective text encoding for chinese llama and alpaca, 2024.
\newblock URL \url{https://arxiv.org/abs/2304.08177}.

\bibitem[de~Vries \& Nissim(2021)de~Vries and Nissim]{de-vries-nissim-2021-good}
de~Vries, W. and Nissim, M.
\newblock As good as new. how to successfully recycle {E}nglish {GPT}-2 to make models for other languages.
\newblock In \emph{Findings of the Association for Computational Linguistics: ACL-IJCNLP 2021}, pp.\  836--846, Online, August 2021. Association for Computational Linguistics.
\newblock \doi{10.18653/v1/2021.findings-acl.74}.
\newblock URL \url{https://aclanthology.org/2021.findings-acl.74}.

\bibitem[Dua et~al.(2019)Dua, Wang, Dasigi, Stanovsky, Singh, and Gardner]{dua-etal-2019-drop}
Dua, D., Wang, Y., Dasigi, P., Stanovsky, G., Singh, S., and Gardner, M.
\newblock {DROP}: A reading comprehension benchmark requiring discrete reasoning over paragraphs.
\newblock In \emph{Proceedings of the 2019 Conference of the North {A}merican Chapter of the Association for Computational Linguistics: Human Language Technologies, Volume 1 (Long and Short Papers)}, pp.\  2368--2378, Minneapolis, Minnesota, June 2019. Association for Computational Linguistics.
\newblock \doi{10.18653/v1/N19-1246}.
\newblock URL \url{https://aclanthology.org/N19-1246}.

\bibitem[Dubey et~al.(2024)]{dubey2024llama3herdmodels}
Dubey, A. et~al.
\newblock The llama 3 herd of models, 2024.
\newblock URL \url{https://arxiv.org/abs/2407.21783}.

\bibitem[Fujii et~al.(2024)Fujii, Nakamura, Loem, Iida, Ohi, Hattori, Shota, Mizuki, Yokota, and Okazaki]{fujii2024continual}
Fujii, K., Nakamura, T., Loem, M., Iida, H., Ohi, M., Hattori, K., Shota, H., Mizuki, S., Yokota, R., and Okazaki, N.
\newblock Continual pre-training for cross-lingual {LLM} adaptation: Enhancing japanese language capabilities.
\newblock In \emph{First Conference on Language Modeling}, 2024.
\newblock URL \url{https://openreview.net/forum?id=TQdd1VhWbe}.

\bibitem[Gage(1994)]{bpe}
Gage, P.
\newblock A new algorithm for data compression.
\newblock \emph{C Users J.}, 12\penalty0 (2):\penalty0 23–38, February 1994.
\newblock ISSN 0898-9788.

\bibitem[Garcia et~al.(2021)Garcia, Constant, Parikh, and Firat]{garcia-etal-2021-towards}
Garcia, X., Constant, N., Parikh, A., and Firat, O.
\newblock Towards continual learning for multilingual machine translation via vocabulary substitution.
\newblock In \emph{Proceedings of the 2021 Conference of the North American Chapter of the Association for Computational Linguistics: Human Language Technologies}, pp.\  1184--1192, Online, June 2021. Association for Computational Linguistics.
\newblock \doi{10.18653/v1/2021.naacl-main.93}.
\newblock URL \url{https://aclanthology.org/2021.naacl-main.93}.

\bibitem[Gee et~al.(2022)Gee, Zugarini, Rigutini, and Torroni]{gee-etal-2022-fast}
Gee, L., Zugarini, A., Rigutini, L., and Torroni, P.
\newblock Fast vocabulary transfer for language model compression.
\newblock In \emph{Proceedings of the 2022 Conference on Empirical Methods in Natural Language Processing: Industry Track}, pp.\  409--416, Abu Dhabi, UAE, December 2022. Association for Computational Linguistics.
\newblock \doi{10.18653/v1/2022.emnlp-industry.41}.
\newblock URL \url{https://aclanthology.org/2022.emnlp-industry.41}.

\bibitem[Goyal et~al.(2022)Goyal, Gao, Chaudhary, Chen, Wenzek, Ju, Krishnan, Ranzato, Guzm{\'a}n, and Fan]{goyal-etal-2022-flores}
Goyal, N., Gao, C., Chaudhary, V., Chen, P.-J., Wenzek, G., Ju, D., Krishnan, S., Ranzato, M., Guzm{\'a}n, F., and Fan, A.
\newblock The {F}lores-101 evaluation benchmark for low-resource and multilingual machine translation.
\newblock \emph{Transactions of the Association for Computational Linguistics}, 10:\penalty0 522--538, 2022.
\newblock \doi{10.1162/tacl_a_00474}.
\newblock URL \url{https://aclanthology.org/2022.tacl-1.30}.

\bibitem[Hendrycks et~al.(2021)Hendrycks, Burns, Basart, Zou, Mazeika, Song, and Steinhardt]{hendrycks2021measuring}
Hendrycks, D., Burns, C., Basart, S., Zou, A., Mazeika, M., Song, D., and Steinhardt, J.
\newblock Measuring massive multitask language understanding.
\newblock In \emph{International Conference on Learning Representations}, 2021.
\newblock URL \url{https://openreview.net/forum?id=d7KBjmI3GmQ}.

\bibitem[Hofmann et~al.(2022)Hofmann, Schuetze, and Pierrehumbert]{hofmann-etal-2022-embarrassingly}
Hofmann, V., Schuetze, H., and Pierrehumbert, J.
\newblock An embarrassingly simple method to mitigate undesirable properties of pretrained language model tokenizers.
\newblock In \emph{Proceedings of the 60th Annual Meeting of the Association for Computational Linguistics (Volume 2: Short Papers)}, pp.\  385--393, Dublin, Ireland, May 2022. Association for Computational Linguistics.
\newblock \doi{10.18653/v1/2022.acl-short.43}.
\newblock URL \url{https://aclanthology.org/2022.acl-short.43}.

\bibitem[Hu et~al.(2022)Hu, yelong shen, Wallis, Allen-Zhu, Li, Wang, Wang, and Chen]{hu2022lora}
Hu, E.~J., yelong shen, Wallis, P., Allen-Zhu, Z., Li, Y., Wang, S., Wang, L., and Chen, W.
\newblock Lo{RA}: Low-rank adaptation of large language models.
\newblock In \emph{International Conference on Learning Representations}, 2022.
\newblock URL \url{https://openreview.net/forum?id=nZeVKeeFYf9}.

\bibitem[Kingma \& Ba(2014)Kingma and Ba]{kingma2014adam}
Kingma, D.~P. and Ba, J.
\newblock Adam: A method for stochastic optimization, 2014.

\bibitem[Liang et~al.(2023)Liang, Gonen, Mao, Hou, Goyal, Ghazvininejad, Zettlemoyer, and Khabsa]{liang-etal-2023-xlm}
Liang, D., Gonen, H., Mao, Y., Hou, R., Goyal, N., Ghazvininejad, M., Zettlemoyer, L., and Khabsa, M.
\newblock {XLM}-{V}: Overcoming the vocabulary bottleneck in multilingual masked language models.
\newblock In Bouamor, H., Pino, J., and Bali, K. (eds.), \emph{Proceedings of the 2023 Conference on Empirical Methods in Natural Language Processing}, pp.\  13142--13152, Singapore, December 2023. Association for Computational Linguistics.
\newblock \doi{10.18653/v1/2023.emnlp-main.813}.
\newblock URL \url{https://aclanthology.org/2023.emnlp-main.813}.

\bibitem[Liu et~al.(2022)Liu, Tam, Mohammed, Mohta, Huang, Bansal, and Raffel]{liu2022fewshot}
Liu, H., Tam, D., Mohammed, M., Mohta, J., Huang, T., Bansal, M., and Raffel, C.
\newblock Few-shot parameter-efficient fine-tuning is better and cheaper than in-context learning.
\newblock In Oh, A.~H., Agarwal, A., Belgrave, D., and Cho, K. (eds.), \emph{Advances in Neural Information Processing Systems}, 2022.
\newblock URL \url{https://openreview.net/forum?id=rBCvMG-JsPd}.

\bibitem[Luo et~al.(2024)Luo, Yang, Meng, Li, Zhou, and Zhang]{luo2024empiricalstudycatastrophicforgetting}
Luo, Y., Yang, Z., Meng, F., Li, Y., Zhou, J., and Zhang, Y.
\newblock An empirical study of catastrophic forgetting in large language models during continual fine-tuning, 2024.
\newblock URL \url{https://arxiv.org/abs/2308.08747}.

\bibitem[Marchisio et~al.(2023)Marchisio, Lewis, Chen, and Artetxe]{marchisio-etal-2023-mini}
Marchisio, K., Lewis, P., Chen, Y., and Artetxe, M.
\newblock Mini-model adaptation: Efficiently extending pretrained models to new languages via aligned shallow training.
\newblock In \emph{Findings of the Association for Computational Linguistics: ACL 2023}, pp.\  5474--5490, Toronto, Canada, July 2023. Association for Computational Linguistics.
\newblock \doi{10.18653/v1/2023.findings-acl.338}.
\newblock URL \url{https://aclanthology.org/2023.findings-acl.338}.

\bibitem[Muennighoff et~al.(2023)Muennighoff, Wang, Sutawika, Roberts, Biderman, Le~Scao, Bari, Shen, Yong, Schoelkopf, Tang, Radev, Aji, Almubarak, Albanie, Alyafeai, Webson, Raff, and Raffel]{muennighoff-etal-2023-crosslingual}
Muennighoff, N., Wang, T., Sutawika, L., Roberts, A., Biderman, S., Le~Scao, T., Bari, M.~S., Shen, S., Yong, Z.~X., Schoelkopf, H., Tang, X., Radev, D., Aji, A.~F., Almubarak, K., Albanie, S., Alyafeai, Z., Webson, A., Raff, E., and Raffel, C.
\newblock Crosslingual generalization through multitask finetuning.
\newblock In \emph{Proceedings of the 61st Annual Meeting of the Association for Computational Linguistics (Volume 1: Long Papers)}, pp.\  15991--16111, Toronto, Canada, July 2023. Association for Computational Linguistics.
\newblock \doi{10.18653/v1/2023.acl-long.891}.
\newblock URL \url{https://aclanthology.org/2023.acl-long.891}.

\bibitem[Pfeiffer et~al.(2020)Pfeiffer, Vuli{\'c}, Gurevych, and Ruder]{pfeiffer-etal-2020-mad}
Pfeiffer, J., Vuli{\'c}, I., Gurevych, I., and Ruder, S.
\newblock {MAD-X}: {A}n {A}dapter-{B}ased {F}ramework for {M}ulti-{T}ask {C}ross-{L}ingual {T}ransfer.
\newblock In \emph{Proceedings of the 2020 Conference on Empirical Methods in Natural Language Processing (EMNLP)}, pp.\  7654--7673, Online, November 2020. Association for Computational Linguistics.
\newblock \doi{10.18653/v1/2020.emnlp-main.617}.
\newblock URL \url{https://aclanthology.org/2020.emnlp-main.617}.

\bibitem[Pfeiffer et~al.(2021)Pfeiffer, Vuli{\'c}, Gurevych, and Ruder]{pfeiffer-etal-2021-unks}
Pfeiffer, J., Vuli{\'c}, I., Gurevych, I., and Ruder, S.
\newblock {UNK}s everywhere: {A}dapting multilingual language models to new scripts.
\newblock In \emph{Proceedings of the 2021 Conference on Empirical Methods in Natural Language Processing}, pp.\  10186--10203, Online and Punta Cana, Dominican Republic, November 2021. Association for Computational Linguistics.
\newblock \doi{10.18653/v1/2021.emnlp-main.800}.
\newblock URL \url{https://aclanthology.org/2021.emnlp-main.800}.

\bibitem[Pfeiffer et~al.(2022)Pfeiffer, Goyal, Lin, Li, Cross, Riedel, and Artetxe]{pfeiffer-etal-2022-lifting}
Pfeiffer, J., Goyal, N., Lin, X., Li, X., Cross, J., Riedel, S., and Artetxe, M.
\newblock Lifting the curse of multilinguality by pre-training modular transformers.
\newblock In \emph{Proceedings of the 2022 Conference of the North American Chapter of the Association for Computational Linguistics: Human Language Technologies}, pp.\  3479--3495, Seattle, United States, July 2022. Association for Computational Linguistics.
\newblock \doi{10.18653/v1/2022.naacl-main.255}.
\newblock URL \url{https://aclanthology.org/2022.naacl-main.255}.

\bibitem[Riviere et~al.(2024)]{team2024gemma}
Riviere, M. et~al.
\newblock Gemma 2: Improving open language models at a practical size, 2024.

\bibitem[Robinson et~al.(2023)Robinson, Ogayo, Mortensen, and Neubig]{robinson-etal-2023-chatgpt}
Robinson, N., Ogayo, P., Mortensen, D.~R., and Neubig, G.
\newblock {C}hat{GPT} {MT}: Competitive for high- (but not low-) resource languages.
\newblock In Koehn, P., Haddow, B., Kocmi, T., and Monz, C. (eds.), \emph{Proceedings of the Eighth Conference on Machine Translation}, pp.\  392--418, Singapore, December 2023. Association for Computational Linguistics.
\newblock \doi{10.18653/v1/2023.wmt-1.40}.
\newblock URL \url{https://aclanthology.org/2023.wmt-1.40}.

\bibitem[Sakaguchi et~al.(2021)Sakaguchi, Bras, Bhagavatula, and Choi]{winogrande}
Sakaguchi, K., Bras, R.~L., Bhagavatula, C., and Choi, Y.
\newblock Winogrande: an adversarial winograd schema challenge at scale.
\newblock \emph{Commun. ACM}, 64\penalty0 (9):\penalty0 99–106, August 2021.
\newblock ISSN 0001-0782.
\newblock \doi{10.1145/3474381}.
\newblock URL \url{https://doi.org/10.1145/3474381}.

\bibitem[Sennrich et~al.(2016)Sennrich, Haddow, and Birch]{sennrich-etal-2016-neural}
Sennrich, R., Haddow, B., and Birch, A.
\newblock Neural machine translation of rare words with subword units.
\newblock In \emph{Proceedings of the 54th Annual Meeting of the Association for Computational Linguistics (Volume 1: Long Papers)}, pp.\  1715--1725, Berlin, Germany, August 2016. Association for Computational Linguistics.
\newblock \doi{10.18653/v1/P16-1162}.
\newblock URL \url{https://aclanthology.org/P16-1162}.

\bibitem[Shi et~al.(2023)Shi, Suzgun, Freitag, Wang, Srivats, Vosoughi, Chung, Tay, Ruder, Zhou, Das, and Wei]{shi2023language}
Shi, F., Suzgun, M., Freitag, M., Wang, X., Srivats, S., Vosoughi, S., Chung, H.~W., Tay, Y., Ruder, S., Zhou, D., Das, D., and Wei, J.
\newblock Language models are multilingual chain-of-thought reasoners.
\newblock In \emph{The Eleventh International Conference on Learning Representations}, 2023.
\newblock URL \url{https://openreview.net/forum?id=fR3wGCk-IXp}.

\bibitem[Shi et~al.(2024)Shi, Xu, Wang, Qin, Wang, Wang, Wang, Ebrahimi, and Wang]{shi2024continuallearninglargelanguage}
Shi, H., Xu, Z., Wang, H., Qin, W., Wang, W., Wang, Y., Wang, Z., Ebrahimi, S., and Wang, H.
\newblock Continual learning of large language models: A comprehensive survey, 2024.
\newblock URL \url{https://arxiv.org/abs/2404.16789}.

\bibitem[Singh et~al.(2024)Singh, Gupta, Bharadwaj, Tewari, and Talukdar]{singh-etal-2024-indicgenbench}
Singh, H., Gupta, N., Bharadwaj, S., Tewari, D., and Talukdar, P.
\newblock {I}ndic{G}en{B}ench: A multilingual benchmark to evaluate generation capabilities of {LLM}s on {I}ndic languages.
\newblock In Ku, L.-W., Martins, A., and Srikumar, V. (eds.), \emph{Proceedings of the 62nd Annual Meeting of the Association for Computational Linguistics (Volume 1: Long Papers)}, pp.\  11047--11073, Bangkok, Thailand, August 2024. Association for Computational Linguistics.
\newblock \doi{10.18653/v1/2024.acl-long.595}.
\newblock URL \url{https://aclanthology.org/2024.acl-long.595}.

\bibitem[Tang et~al.(2024)Tang, Luo, Huang, Zhang, Wang, Zhao, Wei, and Wen]{tang-etal-2024-language}
Tang, T., Luo, W., Huang, H., Zhang, D., Wang, X., Zhao, X., Wei, F., and Wen, J.-R.
\newblock Language-specific neurons: The key to multilingual capabilities in large language models.
\newblock In Ku, L.-W., Martins, A., and Srikumar, V. (eds.), \emph{Proceedings of the 62nd Annual Meeting of the Association for Computational Linguistics (Volume 1: Long Papers)}, pp.\  5701--5715, Bangkok, Thailand, August 2024. Association for Computational Linguistics.
\newblock \doi{10.18653/v1/2024.acl-long.309}.
\newblock URL \url{https://aclanthology.org/2024.acl-long.309}.

\bibitem[Tao et~al.(2024)Tao, Liu, Dou, Muennighoff, Wan, Luo, Lin, and Wong]{tao2024scaling}
Tao, C., Liu, Q., Dou, L., Muennighoff, N., Wan, Z., Luo, P., Lin, M., and Wong, N.
\newblock Scaling laws with vocabulary: Larger models deserve larger vocabularies.
\newblock In \emph{The Thirty-eighth Annual Conference on Neural Information Processing Systems}, 2024.
\newblock URL \url{https://openreview.net/forum?id=sKCKPr8cRL}.

\bibitem[Tay et~al.(2023)Tay, Dehghani, Tran, Garcia, Wei, Wang, Chung, Bahri, Schuster, Zheng, Zhou, Houlsby, and Metzler]{tay2023ul}
Tay, Y., Dehghani, M., Tran, V.~Q., Garcia, X., Wei, J., Wang, X., Chung, H.~W., Bahri, D., Schuster, T., Zheng, S., Zhou, D., Houlsby, N., and Metzler, D.
\newblock {UL}2: Unifying language learning paradigms.
\newblock In \emph{The Eleventh International Conference on Learning Representations}, 2023.
\newblock URL \url{https://openreview.net/forum?id=6ruVLB727MC}.

\bibitem[{\"U}st{\"u}n et~al.(2024){\"U}st{\"u}n, Aryabumi, Yong, Ko, D{'}souza, Onilude, Bhandari, Singh, Ooi, Kayid, Vargus, Blunsom, Longpre, Muennighoff, Fadaee, Kreutzer, and Hooker]{ustun-etal-2024-aya}
{\"U}st{\"u}n, A., Aryabumi, V., Yong, Z., Ko, W.-Y., D{'}souza, D., Onilude, G., Bhandari, N., Singh, S., Ooi, H.-L., Kayid, A., Vargus, F., Blunsom, P., Longpre, S., Muennighoff, N., Fadaee, M., Kreutzer, J., and Hooker, S.
\newblock Aya model: An instruction finetuned open-access multilingual language model.
\newblock In Ku, L.-W., Martins, A., and Srikumar, V. (eds.), \emph{Proceedings of the 62nd Annual Meeting of the Association for Computational Linguistics (Volume 1: Long Papers)}, pp.\  15894--15939, Bangkok, Thailand, August 2024. Association for Computational Linguistics.
\newblock \doi{10.18653/v1/2024.acl-long.845}.
\newblock URL \url{https://aclanthology.org/2024.acl-long.845}.

\bibitem[Wei et~al.(2022)Wei, Bosma, Zhao, Guu, Yu, Lester, Du, Dai, and Le]{wei2022finetuned}
Wei, J., Bosma, M., Zhao, V., Guu, K., Yu, A.~W., Lester, B., Du, N., Dai, A.~M., and Le, Q.~V.
\newblock Finetuned language models are zero-shot learners.
\newblock In \emph{International Conference on Learning Representations}, 2022.
\newblock URL \url{https://openreview.net/forum?id=gEZrGCozdqR}.

\bibitem[Wendler et~al.(2024)Wendler, Veselovsky, Monea, and West]{wendler-etal-2024-llamas}
Wendler, C., Veselovsky, V., Monea, G., and West, R.
\newblock Do llamas work in {E}nglish? on the latent language of multilingual transformers.
\newblock In Ku, L.-W., Martins, A., and Srikumar, V. (eds.), \emph{Proceedings of the 62nd Annual Meeting of the Association for Computational Linguistics (Volume 1: Long Papers)}, pp.\  15366--15394, Bangkok, Thailand, August 2024. Association for Computational Linguistics.
\newblock \doi{10.18653/v1/2024.acl-long.820}.
\newblock URL \url{https://aclanthology.org/2024.acl-long.820}.

\bibitem[Xue et~al.(2021)Xue, Constant, Roberts, Kale, Al-Rfou, Siddhant, Barua, and Raffel]{xue-etal-2021-mt5}
Xue, L., Constant, N., Roberts, A., Kale, M., Al-Rfou, R., Siddhant, A., Barua, A., and Raffel, C.
\newblock m{T}5: A massively multilingual pre-trained text-to-text transformer.
\newblock In \emph{Proceedings of the 2021 Conference of the North American Chapter of the Association for Computational Linguistics: Human Language Technologies}, pp.\  483--498, Online, June 2021. Association for Computational Linguistics.
\newblock \doi{10.18653/v1/2021.naacl-main.41}.
\newblock URL \url{https://aclanthology.org/2021.naacl-main.41}.

\bibitem[Yang et~al.(2024)]{yang2024qwen2technicalreport}
Yang, A. et~al.
\newblock Qwen2 technical report, 2024.
\newblock URL \url{https://arxiv.org/abs/2407.10671}.

\bibitem[Yong et~al.(2023)Yong, Schoelkopf, Muennighoff, Aji, Adelani, Almubarak, Bari, Sutawika, Kasai, Baruwa, Winata, Biderman, Raff, Radev, and Nikoulina]{yong-etal-2023-bloom}
Yong, Z.~X., Schoelkopf, H., Muennighoff, N., Aji, A.~F., Adelani, D.~I., Almubarak, K., Bari, M.~S., Sutawika, L., Kasai, J., Baruwa, A., Winata, G., Biderman, S., Raff, E., Radev, D., and Nikoulina, V.
\newblock {BLOOM}+1: Adding language support to {BLOOM} for zero-shot prompting.
\newblock In \emph{Proceedings of the 61st Annual Meeting of the Association for Computational Linguistics (Volume 1: Long Papers)}, pp.\  11682--11703, Toronto, Canada, July 2023. Association for Computational Linguistics.
\newblock \doi{10.18653/v1/2023.acl-long.653}.
\newblock URL \url{https://aclanthology.org/2023.acl-long.653}.

\bibitem[Yue et~al.(2024)Yue, Zheng, Zhang, and Chen]{yue2024mammoth}
Yue, X., Zheng, T., Zhang, G., and Chen, W.
\newblock {MA}mmo{TH}2: Scaling instructions from the web.
\newblock In \emph{The Thirty-eighth Annual Conference on Neural Information Processing Systems}, 2024.
\newblock URL \url{https://openreview.net/forum?id=yVu5dnPlqA}.

\bibitem[Zellers et~al.(2019)Zellers, Holtzman, Bisk, Farhadi, and Choi]{zellers-etal-2019-hellaswag}
Zellers, R., Holtzman, A., Bisk, Y., Farhadi, A., and Choi, Y.
\newblock {H}ella{S}wag: Can a machine really finish your sentence?
\newblock In \emph{Proceedings of the 57th Annual Meeting of the Association for Computational Linguistics}, pp.\  4791--4800, Florence, Italy, July 2019. Association for Computational Linguistics.
\newblock \doi{10.18653/v1/P19-1472}.
\newblock URL \url{https://aclanthology.org/P19-1472}.

\bibitem[Zhang et~al.(2022)Zhang, Chaudhary, Goyal, Cross, Wenzek, Bansal, and Guzman]{zhang-etal-2022-robust}
Zhang, S., Chaudhary, V., Goyal, N., Cross, J., Wenzek, G., Bansal, M., and Guzman, F.
\newblock How robust is neural machine translation to language imbalance in multilingual tokenizer training?
\newblock In \emph{Proceedings of the 15th biennial conference of the Association for Machine Translation in the Americas (Volume 1: Research Track)}, pp.\  97--116, Orlando, USA, September 2022. Association for Machine Translation in the Americas.
\newblock URL \url{https://aclanthology.org/2022.amta-research.8}.

\bibitem[Zhao et~al.(2024{\natexlab{a}})Zhao, Zhang, Gao, Zhang, Gui, and Huang]{zhao2024llamaenglishempiricalstudy}
Zhao, J., Zhang, Z., Gao, L., Zhang, Q., Gui, T., and Huang, X.
\newblock Llama beyond english: An empirical study on language capability transfer, 2024{\natexlab{a}}.
\newblock URL \url{https://arxiv.org/abs/2401.01055}.

\bibitem[Zhao et~al.(2024{\natexlab{b}})Zhao, Zhang, Chen, Kawaguchi, and Bing]{zhao2024how}
Zhao, Y., Zhang, W., Chen, G., Kawaguchi, K., and Bing, L.
\newblock How do large language models handle multilingualism?
\newblock In \emph{The Thirty-eighth Annual Conference on Neural Information Processing Systems}, 2024{\natexlab{b}}.
\newblock URL \url{https://openreview.net/forum?id=ctXYOoAgRy}.

\bibitem[Zheng et~al.(2024)Zheng, Pan, Xu, Qin, Yue, and Zhou]{zheng-etal-2024-breaking}
Zheng, W., Pan, W., Xu, X., Qin, L., Yue, L., and Zhou, M.
\newblock Breaking language barriers: Cross-lingual continual pre-training at scale.
\newblock In Al-Onaizan, Y., Bansal, M., and Chen, Y.-N. (eds.), \emph{Proceedings of the 2024 Conference on Empirical Methods in Natural Language Processing}, pp.\  7725--7738, Miami, Florida, USA, November 2024. Association for Computational Linguistics.
\newblock URL \url{https://aclanthology.org/2024.emnlp-main.441}.

\end{thebibliography}
\bibliographystyle{icml2025}

\clearpage
\appendix
\section*{Overview of Appendix}
Our supplementary includes the following sections:
\begin{itemize}
    \item Section~\ref{appendix:exp_settings}: Experimental Settings, including implementation details for training and evaluation.
    \item Section~\ref{appendix:english_tasks}: Detailed result comparison between instruction-tuned models and \ouradapter in English tasks.
    \item Section~\ref{appendix:extra_main_results}: Additional \la and \ouradapter results on \palmtwo, \aya and \gemmatwo-\texttt{IT} models.
    \item Section~\ref{appendix:cpt}: Results of the continued pre-training baseline for language adaptation.
    \item Section~\ref{appendix:supplementary_analysis}: Supplementary analysis including additional data ablations.
    \item Section~\ref{appendix:detailed_lang_results}: Results by language for both language adaptation and \ouradapter.
\end{itemize}

\section{Experimental Settings}\label{appendix:exp_settings}
\subsection{Training Details}\label{appendix:training_details}
For language adaptation, we use a constant learning rate of $1\times 10^{-4}$ for \palmtwo and $1\times 10^{-5}$ for \gemmatwo and \aya with the Adam optimizer~\citep{kingma2014adam}. The embeddings are trained on a total of 200B tokens,\footnote{English accounts for 30.8\% of the tokens.} with each batch consisting of examples packed to a maximum sequence length of 2K for \palmtwo and 8K for \gemmatwo and \aya. We pre-train the model using the UL2 objectives~\citep{tay2023ul} for \palmtwo and causal language modeling objectives for \gemmatwo and \aya. Language adaptation consumes up to 256 TPU-v5 chips for the largest \gemmatwo-\texttt{27B} and \aya-\texttt{35B} models. The batch size is selected based on the model size and computing resources we have, with batch sizes of 256, 128, and 64 assigned to the \gemmatwo-\texttt{2B}, \texttt{9B}, and \texttt{27B} models, respectively. A similar strategy is applied to the \aya models. For \palmtwo models, we use the same batch size of 2048 for all variants. We choose the best checkpoint based on the performance of \flores development sets corresponding to each target language group. The training time varies with model size, where smaller models complete training within 24 hours while larger models require up to 1 week to finish.

We instruction-tune the transformer body of LLMs on $\mathcal{D}_{it}$ using the same hyper-parameter setting to obtain \texttt{XX-FLAN}, where we sample up to 200M instances from the FLAN mixture to construct $\mathcal{D}_{it}$. We use early stopping to select the best model based on the performance on MMLU~\citep{hendrycks2021measuring} and assemble it with customized embedding to obtain \texttt{XX-FA}. We observe all LLMs converge fast, as indicated by the average performance on MMLU. In most cases, training is completed within 24 hours.

For LoRA-Adaptation, we add LoRA weights to the self-attention module of all transformer layers with a LoRA rank of 64 and exclusively optimize these weights.\footnote{This adds less than 1\% parameters.} We use a learning rate of $5\times 10^{-6}$ for all models with 10\% steps of warm-up. Analogous to embedding tuning, the \flores development set is used for model selection. The training process is computationally efficient, completing within 12 hours even for the largest model.

\begin{table}[t]
\setlength{\tabcolsep}{5pt}
\footnotesize
\centering
\vspace{-3mm}
\caption{Prompt templates used in each of the evaluation dataset. For few-shot evaluation, n-shot examples have the same format as the last test instance, which comes after the preamble but before the test instances.}
\begin{tabular}{l|p{5.5cm}}
    \toprule
    Dataset & Prompt \\
    \midrule
    \belebele & The following are multiple choice questions (with answers). \\ \\
    & Passage: \textcolor{blue!80}{[Target Language Passage]} \\
    & Question: \textcolor{blue!80}{[Target Language Question]} \\
    & (A) \textcolor{blue!80}{[Choice A]} (B) \textcolor{blue!80}{[Choice B]} (C) \textcolor{blue!80}{[Choice C]} (D) \textcolor{blue!80}{[Choice D]} \\
    & Answer: \\ 

    \midrule
    \sib & \textcolor{blue!80}{[News Article in Target Language]} \\
    & Question: What label best describes this news article? \\
    & (A) \textcolor{blue!80}{science/technology} (B) \textcolor{blue!80}{travel} (C) \textcolor{blue!80}{politics} (D) \textcolor{blue!80}{sports} (E) \textcolor{blue!80}{health} (F) \textcolor{blue!80}{entertainment} (G) \textcolor{blue!80}{geography} \\
    & Answer: \\
    
    \midrule
    \flores & Translate this from English to \textcolor{blue!80}{[Target Language Name]}: \\
    \\
    & English: \textcolor{blue!80}{[Sentence in English]} \\
    & \textcolor{blue!80}{[Target Language Name]}: \\
    
    \midrule
    \xsumin & I will first show a news article in English and then provide a one sentence summary of it in \textcolor{blue!80}{[Target Language Name]}. \\ 
    \\
    & Summarize the following article: \textcolor{blue!80}{[Article in English]} \\
    & Summary in \textcolor{blue!80}{[Target Language Name]}: \\
    
    \midrule
    \xorqain & Generate an answer in \textcolor{blue!80}{[Target Language Name/English]} for the question based on the given passage: \\ 
    \\
    & \textcolor{blue!80}{[Passage in English]} \\
    & Q: \textcolor{blue!80}{[Question in Target Language]} \\
    & A: \\
    
    \midrule
    GSM8K-NTL & Q: \textcolor{blue!80}{[Question in Target Language]} \\
    & A: \textcolor{blue!80}{[Let's think step by step.]} \\
    
    \bottomrule
    \multicolumn{2}{c}{} \vspace{-8mm}
    \end{tabular}
\label{tab:prompt_formats}
\end{table}
\begin{figure*}[t]
    \setlength{\belowcaptionskip}{-0.35cm}
    \centering
    \includegraphics[width=0.95\linewidth]{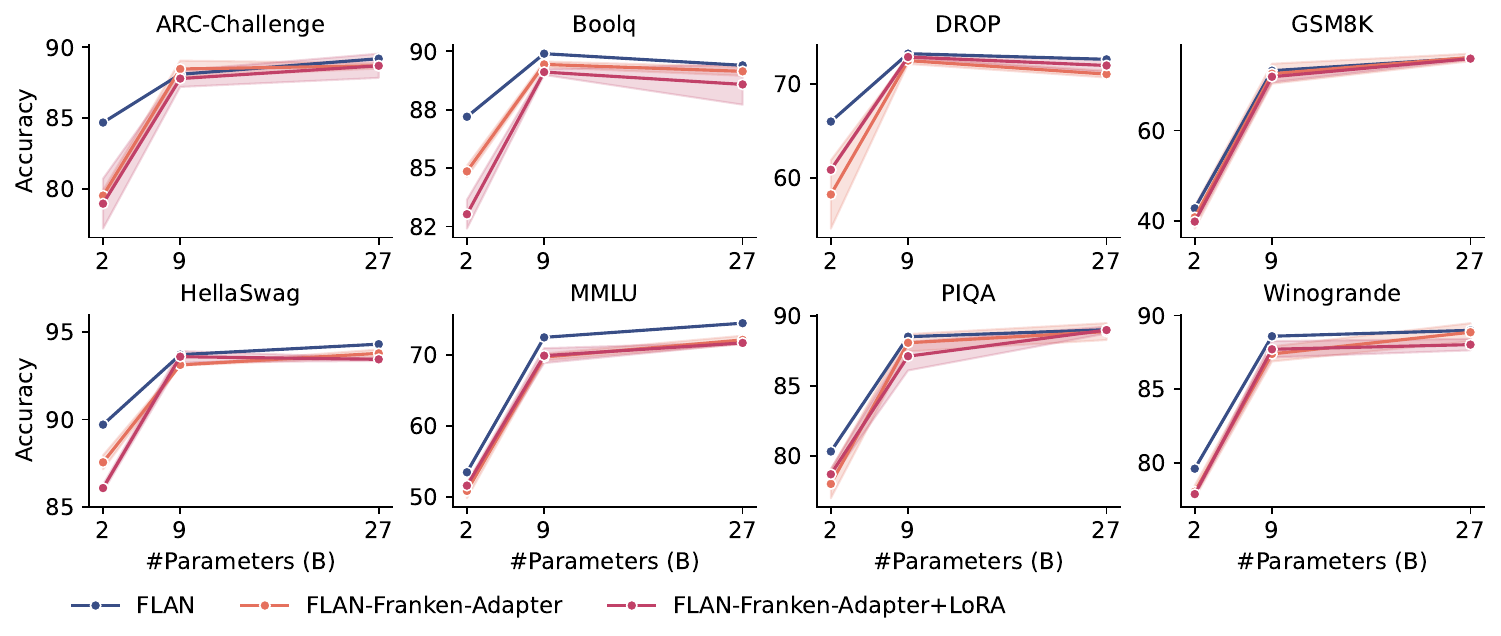}
    \vspace{-6mm}
    \caption{Performance on English tasks. For \ouradapter methods, we present the averaged performance with variances across three embeddings (\ie \sea, \afr, \ind). Shading indicates the standard deviations measured over three embeddings.}
    \vspace{-4mm}
    \label{fig:franken_adapter_english_benchmark}
\end{figure*}

\subsection{Evaluation Details}\label{appendix:eval_details}
We use the prompt formats listed in Table~\ref{tab:prompt_formats} for evaluation. For generation tasks, greedy decoding is employed with a maximum sequence length of 256 tokens. For classification tasks, we calculate the logits of each available option (\eg (A), (B)) and select the option with the highest score as the predicted answer.

For evaluating models in Figure~\ref{fig:result_summarization}, we use the same prompt formats shown in Table~\ref{tab:prompt_formats} for \texttt{Aya23-35B}~\citep{aryabumi2024aya}, \texttt{BLOOMZ-7B}~\citep{muennighoff-etal-2023-crosslingual}, and \texttt{Qwen2.5-32B-IT}~\citep{yang2024qwen2technicalreport}. We collect evaluation results of \texttt{ChatGPT-3.5-Turbo} from different papers: \flores from~\citet{robinson-etal-2023-chatgpt}, \sib from~\citet{adelani-etal-2024-sib}, \belebele from~\citet{bandarkar-etal-2024-belebele},\footnote{The results of shn\_Mymr and ory\_Orya are missing.} and \xorqain and \xsumin from~\citet{singh-etal-2024-indicgenbench}.\footnote{We copy the 1-shot results from~\citet{singh-etal-2024-indicgenbench}.} We use the officially-released translations\footnote{\url{https://github.com/facebookresearch/fairseq/tree/nllb}} for obtaining the results in \texttt{NLLB-54B-MOE}~\citep{nllbteam2022languageleftbehindscaling}.\footnote{The results of min\_Arab, ful\_Latn, orm\_Latn, and kon\_Latn are missing, which are excluded when computing the average score in NLLB-54B-MOE.}

\section{Details on English Benchmarking}\label{appendix:english_tasks}
For benchmarking performance in English, we choose ARC-Challenge~\citep{clark2018thinksolvedquestionanswering}, Boolq~\citep{clark-etal-2019-boolq}, DROP~\citep{dua-etal-2019-drop}, GSM8K~\citep{cobbe2021trainingverifierssolvemath}, HellaSwag~\citep{zellers-etal-2019-hellaswag}, MMLU~\citep{hendrycks2021measuring}, PIQA~\citep{bisk2019piqa}, and Winogrande~\citep{winogrande}. We use different number of shots for evaluation following~\citet{team2024gemma}.

Figure~\ref{fig:franken_adapter_english_benchmark} shows the comparison between the instruction-tuned FLAN model and our \ouradapter across various sizes of \gemmatwo. Both of our \ouradapter variants exhibit performance regressions across all tasks compared to the FLAN model, although the performance gap closes as model capacity scales up. We believe that these minor regressions are justifiable in light of the substantial gains achieved in zero-shot cross-lingual transfer.

\begin{table*}[t]
\setlength{\belowcaptionskip}{-0.3cm}
\setlength{\tabcolsep}{6pt}
\footnotesize
\centering
\vspace{-3mm}
\caption{Additional \ouradapter results on \palmtwo, \aya and \gemmatwo-\texttt{IT} models. The best and second-best results are marked in \textbf{bold} and \underline{underlined}. \textcolor{red!80}{Red} values indicate \ouradapter hurts the performance.} 
\resizebox{\linewidth}{!}{
\begin{tabular}{l|c|ccc|ccc|ccc|cc|cc|c}
    \toprule
    \bf Task Type & \multirow{4}{*}[-1ex]{\rotatebox[origin=c]{90}{\bf \textsc{English}}} & \multicolumn{6}{c|}{\bf \textsc{Classification}} & \multicolumn{7}{c|}{\bf \textsc{Generation}} \\
    \cmidrule(lr){3-8} \cmidrule(lr){9-15}
    \multirow{2}{*}{\bf Eval. Metric} & & \multicolumn{3}{c|}{\bf \belebele} & \multicolumn{3}{c|}{\bf \sib} & \multicolumn{3}{c|}{\bf \flores} & \multicolumn{2}{c|}{\bf \textsc{XorQA-In}} & \multicolumn{2}{c|}{\bf \textsc{XSum-In}} & \multirow{4}{*}[2ex]{\bf Avg.} \\
    & & \multicolumn{3}{c|}{\bf \texttt{Accuracy}} & \multicolumn{3}{c|}{\bf \texttt{Accuracy}} & \multicolumn{3}{c|}{\bf \texttt{ChrF++}} & \multicolumn{2}{c|}{\bf \texttt{Token-F1}} & \multicolumn{2}{c|}{\bf \texttt{ChrF}} \\
    \cmidrule(lr){3-5} \cmidrule(lr){6-8} \cmidrule(lr){9-11} \cmidrule(lr){12-13} \cmidrule(lr){14-15}
    \bf Model & & \sea & \afr & \ind & \sea & \afr & \ind & \sea & \afr & \ind & \ind & \textsc{En} & \ind & \textsc{En} \\
    \midrule
    \palmtwo-\texttt{XXS-FLAN} & \underline{59.7} & 54.5 & 40.0 & 49.4 & 67.2 & 51.3 & 70.9 & \underline{28.5} & \underline{13.9} & \underline{21.0} &  9.7 & 42.1 & \bf 4.8 & \underline{33.7} & 32.8 \\
    \palmtwo-\texttt{XXS-FA} & \bf 62.6 & \underline{58.0} & \underline{40.2} & \underline{50.9} & \bf 76.8 & \underline{62.2} & \bf 77.2 & \textcolor{red!80}{27.7} & \textcolor{red!80}{12.2} & \textcolor{red!80}{17.9} & \underline{10.4} & \underline{59.4} & \underline{\textcolor{red!80}{2.1}} & \bf 34.2 & \underline{35.9} \\
    {\pfix}  \texttt{LoRA-Adapt} & 58.9 & \bf 59.7 & \bf 43.4 & \bf 53.8 & \underline{73.7} & \bf 63.5 & \underline{76.9} & \bf 32.3 & \bf 16.0 & \bf 26.7 & \bf 11.7 & \bf 64.2 & \textcolor{red!80}{1.0} & \textcolor{red!80}{30.7} & \bf 37.7 \\
    \midrule
    \palmtwo-\texttt{S-FLAN} & \bf 86.3 & 80.2 & \underline{67.4} & \bf 82.1 & 70.6 & 60.9 & 74.5 & \underline{37.6} & 19.3 & \underline{36.7} & \bf 21.3 & 46.9 & \underline{15.8} & \underline{41.1} & 45.4 \\
    \palmtwo-\texttt{S-FA} & 85.1 & \underline{83.2} & \textcolor{red!80}{67.1} & \textcolor{red!80}{79.9} & \underline{77.0} & \underline{66.9} & \underline{74.6} & \textcolor{red!80}{36.1} & \underline{23.1} & \textcolor{red!80}{34.6} & \underline{\textcolor{red!80}{21.1}} & \underline{51.8} & \bf 17.3 & \textcolor{red!80}{40.1} & \underline{47.1} \\
    {\pfix}  \texttt{LoRA-Adapt} & \underline{85.7} & \bf 83.6 & \bf 68.4 & \underline{\textcolor{red!80}{81.8}} & \bf 77.7 & \bf 68.7 & \bf 77.8 & \bf 39.3 & \bf 24.1 & \bf 38.7 & \textcolor{red!80}{18.9} & \bf 57.3 & \textcolor{red!80}{15.2} & \bf 41.2 & \bf 48.2 \\
    \midrule
    \aya-\texttt{8B-FLAN} & \bf 74.3 & 47.8 & 34.1 & 42.7 & 64.3 & 48.5 & 60.6 & 22.9 & 6.0 & \underline{16.1} & \underline{10.5} & \underline{58.5} & \bf 11.7 & \bf 34.5 & 32.3 \\
    \aya-\texttt{8B-FA} & 71.1 & \underline{56.6} & \underline{38.4} & \underline{48.0} & \bf 72.4 & \bf 58.7 & \bf 71.3 & \underline{28.8} & \underline{9.1} & \textcolor{red!80}{11.5} & \textcolor{red!80}{7.8} & \bf 61.6 & \textcolor{red!80}{3.9} & \textcolor{red!80}{25.7} & \underline{34.2} \\
    {\pfix}  \texttt{LoRA-Adapt} & \underline{71.3} & \bf 59.1 & \bf 39.0 & \bf 50.8 & \underline{65.6} & \underline{55.3} & \underline{65.2} & \bf 37.1 & \bf 16.9 & \bf 33.4 & \underline{\textcolor{red!80}{9.6}} & \textcolor{red!80}{56.5} & \underline{\textcolor{red!80}{9.4}} & \underline{\textcolor{red!80}{30.7}} & \bf 36.8 \\
    \midrule
    \aya-\texttt{35B-FLAN} & \bf 83.6 & 56.9 & \underline{40.3} & \underline{54.6} & 67.0 & 52.9 & 67.2 & 27.1 & 9.1 & \underline{23.6} & \underline{11.4} & \underline{64.7} & \underline{10.4} & \bf 37.9 & 36.5 \\
    \aya-\texttt{35B-FA} & 79.2 & \underline{57.6} & \textcolor{red!80}{40.2} & \textcolor{red!80}{53.0} & \underline{70.8} & \underline{53.8} & \underline{68.5} & \underline{27.7} & \underline{9.5} & \textcolor{red!80}{13.9} & \textcolor{red!80}{6.8} & \textcolor{red!80}{62.3} & \textcolor{red!80}{7.3} & \textcolor{red!80}{22.1} & \textcolor{red!80}{35.2} \\
    {\pfix}  \texttt{LoRA-Adapt} & \underline{82.0} & \bf 72.6 & \bf 50.1 & \bf 65.2 & \bf 75.2 & \bf 64.7 & \bf 74.0 & \bf 41.6 & \bf 23.4 & \bf 40.1 & \bf 12.6 & \bf 67.8 & \bf 15.7 & \underline{\textcolor{red!80}{37.1}} & \bf 45.1 \\
    \midrule
    \gemmatwo-\texttt{2B-IT} & \underline{55.5} & 47.5 & 35.4 & 45.8 & 57.4 & 42.8 & 61.2 & 24.2 & 7.3 & 17.7 & \underline{14.9} & 53.0 & \underline{10.1} & \underline{31.9} & 31.6 \\
    \ \ + \texttt{Lang-Adapt} & 54.1 & 51.2 & 36.7 & 42.9 & \underline{67.0} & 48.7 & \underline{67.1} & \underline{35.4} & \underline{14.2} & \underline{29.2} & 12.3 & 40.2 & 7.1 & 27.5 & 33.0 \\
    \gemmatwo-\texttt{2B-FA} & 55.1 & \underline{53.7} & \underline{37.4} & \underline{46.0} & 66.7 & \underline{49.5} & 64.0 & 27.4 & 11.7 & 27.0 & \bf 16.7 & \underline{58.7} & \bf 13.4 & \bf 32.0 & \underline{35.9} \\
    {\pfix}  \texttt{LoRA Adapt} & \bf 58.4 & \bf 56.4 & \bf 39.9 & \bf 50.7 & \bf 67.5 & \bf 53.0 & \bf 69.9 & \bf 37.6 & \bf 18.6 & \bf 32.4 & \textcolor{red!80}{11.7} & \bf 60.5 & \textcolor{red!80}{9.3} & \textcolor{red!80}{27.7} & \bf 37.6 \\
    \midrule
    \gemmatwo-\texttt{9B-IT} & \bf 79.9 & 71.5 & 51.4 & 72.5 & 72.5 & 57.8 & 79.3 & 32.2 & 14.0 & 31.1 & \bf 24.0 & 64.4 & \underline{16.6} & \bf 34.3 & 44.5 \\
    \ \ + \texttt{Lang-Adapt} & 73.9 & 75.4 & \underline{57.9} & \bf 72.7 & 77.2 & \underline{69.2} & \underline{79.7} & \underline{39.3} & \underline{22.2} & \underline{38.7} & \underline{23.1} & \bf 69.6 & \bf 17.0 & \underline{33.9} & \underline{48.3} \\
    \gemmatwo-\texttt{9B-FA} & 78.5 & \bf 77.4 & 54.5 & \underline{72.6} & \underline{78.2} & 66.1 & \textcolor{red!80}{78.5} & 33.8 & 16.6 & \textcolor{red!80}{18.8} & \textcolor{red!80}{21.4} & 65.2 & \textcolor{red!80}{9.6} & \textcolor{red!80}{34.2} & \textcolor{red!80}{43.8} \\
    {\pfix}  \texttt{LoRA-Adapt} & \underline{79.5} & \underline{76.5} & \bf 58.1 & \textcolor{red!80}{70.8} & \bf 81.3 & \bf 71.2 & \bf 82.8 & \bf 41.5 & \bf 25.1 & \bf 39.8 & \textcolor{red!80}{18.2} & \underline{68.2} & \textcolor{red!80}{15.7} & \textcolor{red!80}{33.0} & \bf 48.4 \\
    \midrule
    \gemmatwo-\texttt{27B-IT} & \underline{82.1} & 74.9 & 54.6 & \underline{75.9} & 76.2 & 63.1 & \bf 83.0 & 35.2 & 20.1 & 34.5 & \bf 27.7 & 68.2 & \bf 19.0 & \underline{34.7} & 48.0 \\
    \ \ + \texttt{Lang-Adapt} & 81.9 & \bf 81.7 & \bf 62.7 & \bf 76.0 & \bf 82.1 & \underline{70.0} & 81.9 & \bf 42.5 & \underline{24.8} & \underline{39.3} & \underline{26.4} & \bf 73.0 & \underline{18.3} & \bf 35.8 & \bf 51.1 \\
    \gemmatwo-\texttt{27B-FA} & 79.9 & \underline{80.8} & 59.5 & \textcolor{red!80}{75.9} & 81.4 & 68.9 & \bf 83.0 & \textcolor{red!80}{35.1} & \textcolor{red!80}{6.2} & \textcolor{red!80}{29.9} & \textcolor{red!80}{20.0} & \underline{68.8} & \textcolor{red!80}{15.0} & \textcolor{red!80}{34.6} & \textcolor{red!80}{46.7} \\
    {\pfix}  \texttt{LoRA-Adapt} & \bf 82.2 & \underline{78.8} & \underline{60.1} & \textcolor{red!80}{74.3} & \underline{81.7} & \bf 71.7 & \underline{\textcolor{red!80}{82.6}} & \underline{42.3} & \bf 25.6 & \bf 40.6 & \textcolor{red!80}{24.2} & \textcolor{red!80}{67.9} & \textcolor{red!80}{15.6} & \textcolor{red!80}{34.0} & \underline{49.6} \\
    \bottomrule
    \multicolumn{13}{c}{} \vspace{-8mm}
    \end{tabular}
    }
\label{tab:franken_adapter_results_other_models}
\end{table*}
\section{Additional Main Results}\label{appendix:extra_main_results}

\begin{figure}
    \setlength{\abovecaptionskip}{-0.0001cm}
    \setlength{\belowcaptionskip}{-0.35cm}
    \centering
    \includegraphics[width=\linewidth]{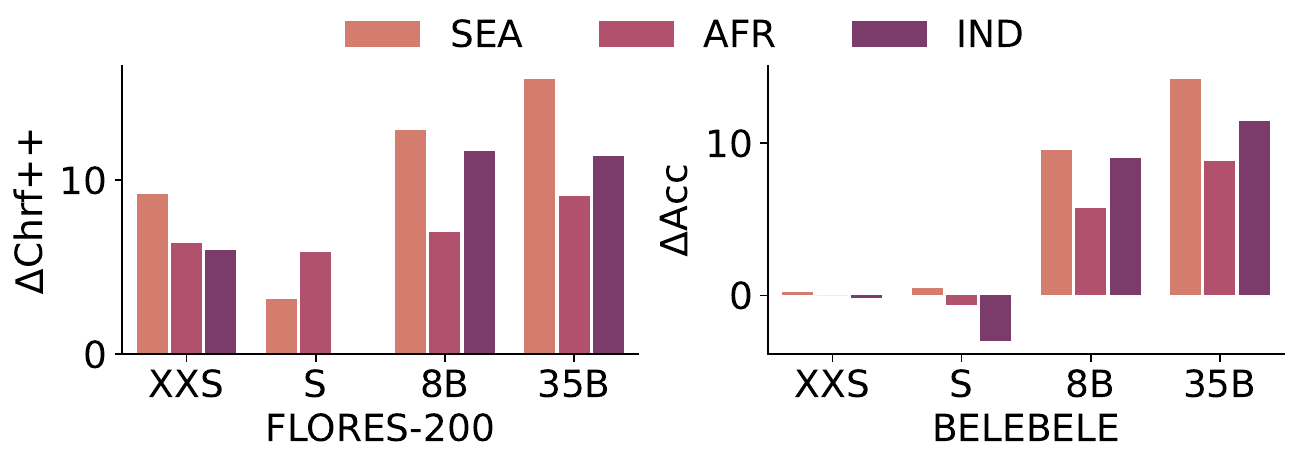}
    \vspace{-8mm}
    \caption{\la on \palmtwo (XXS, S) and \aya (8B, 35B). Absolute gains over the pre-trained models are reported.}
    \vspace{-6mm}
    \label{fig:palm_aya_results}
\end{figure}
\subsection{\la on \palmtwo and \aya}\label{appendix:extra_la_results}
We evaluate the generalization ability of language adaptation on two LLMs with varying levels of multilingualism. Among them, \palmtwo-\texttt{S} exhibits the strongest multilingual abilities while \aya models demonstrate limited multilingual performance. Figure~\ref{fig:palm_aya_results} shows that the performance gains from language adaptation decrease as the original multilingual scope of the LLMs expand (\aya $\rightarrow$ \palmtwo-\texttt{XXS} $\rightarrow$ \palmtwo-\texttt{S}). Moreover, larger performance improvements are observed on \aya models when scaling up their size, suggesting that language adaptation may be particularly effective for models with stronger English proficiency.

\subsection{\ouradapter Results on \palmtwo, \aya and \gemmatwo-\texttt{IT}}\label{appendix:extra_fa_results}
Employing the same \ouradapter pipeline, we observe similar patterns of performance improvement across all sizes of the \palmtwo and \aya models, as shown in Table~\ref{tab:franken_adapter_results_other_models}. This demonstrates the broad applicability of our method to various types of LLMs. In addition, we also show the detailed results of \gemmatwo-\texttt{IT} models, as supplementary to Figure~\ref{fig:franken_adapter_vs_emb_surgery_on_it}. The \ouradapter method also performs effectively with off-the-shelf \textsc{It} models that have undergone complex supervised fine-tuning and reinforcement learning. This further underscores the versatility of our proposed approach in integrating pre-trained multilingual embeddings into LLMs that have been instruction-tuned using diverse methodologies for efficient zero-shot cross-lingual transfer.

\begin{figure*}
    \setlength{\abovecaptionskip}{-0.0001cm}
    \setlength{\belowcaptionskip}{-0.35cm}
    \centering
    \includegraphics[width=0.85\linewidth]{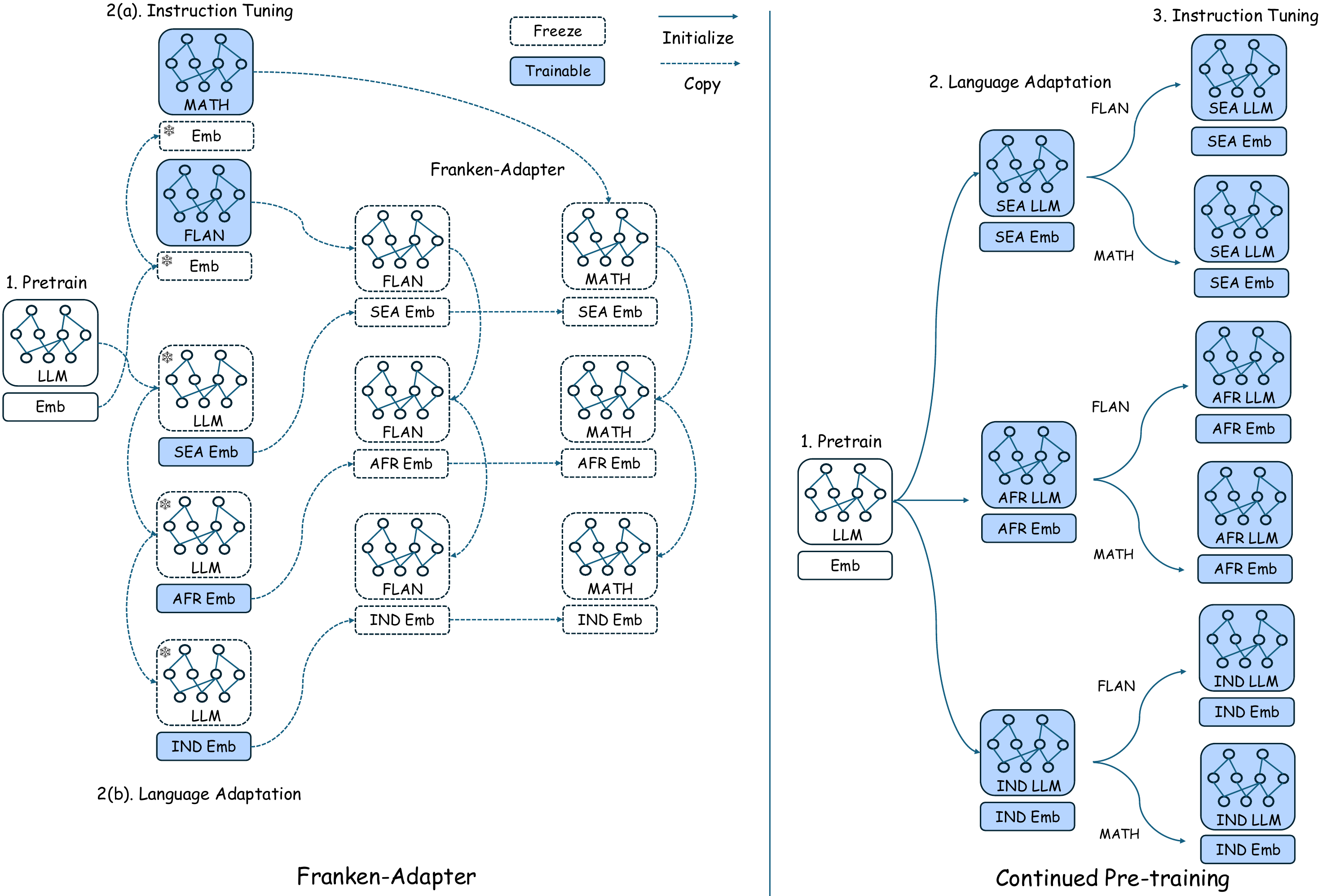}
    \vspace{-4mm}
    \caption{The difference between our \ouradapter and the continued pre-training (CPT) baseline for zero-shot cross-lingual transfer. The same customized tokenizers are used by CPT. For $M$ language groups and $N$ target skills to be adapted, our \ouradapter avoids redundant training through model composition, requiring only $M$ instances of embedding tuning and $N$ instances of transformer body tuning. By contrast, the CPT baseline requires separate adaptation for each target skill, resulting in a total number of $M+M*N$ instances of full-parameter tuning.}
    \vspace{-6mm}
    \label{fig:fa_vs_sct}
\end{figure*}

\begin{figure}
    \setlength{\abovecaptionskip}{-0.0001cm}
    \setlength{\belowcaptionskip}{-0.35cm}
    \centering
    \includegraphics[width=\linewidth]{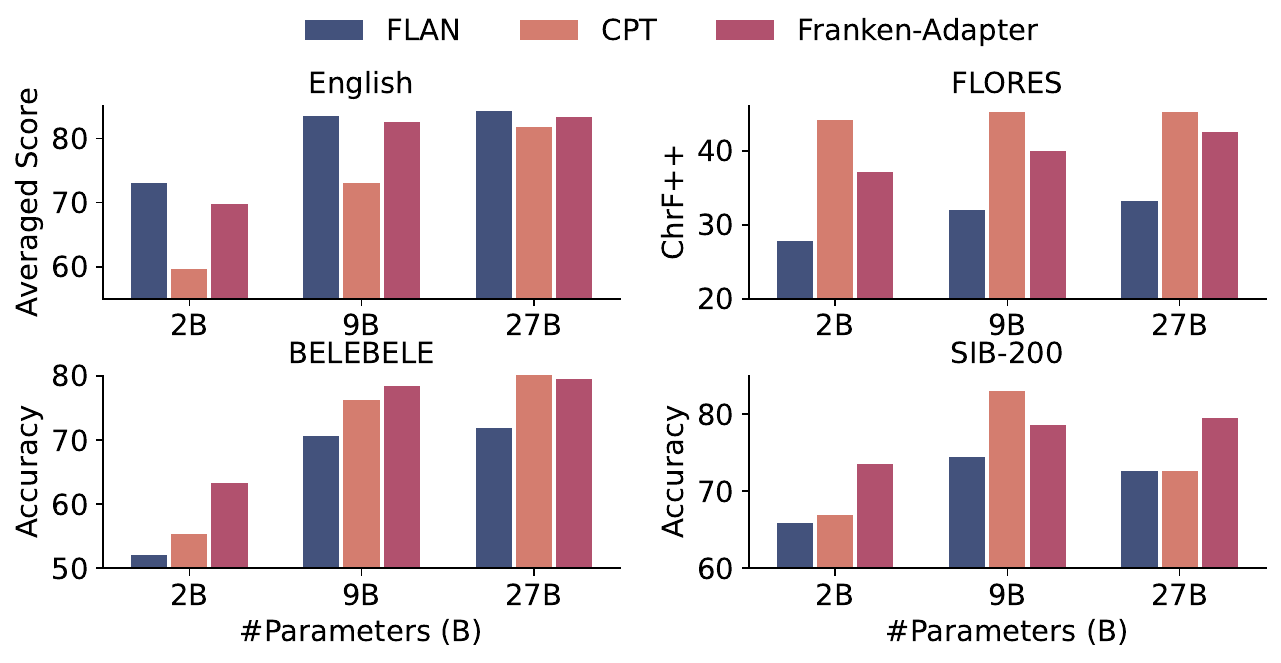}
    \vspace{-8mm}
    \caption{The performance comparison between \ouradapter (w/ LoRA-Adaptation) and the continued pre-training (CPT) baseline. For English, we report the average across 8 datasets. For the rest of multilingual datasets, we report scores of the \sea subsets.}
    \vspace{-6mm}
    \label{fig:fa_vs_cpt_sea}
\end{figure}

\section{Comparison to Continued Pre-training}\label{appendix:cpt}
We compare our \ouradapter approach with a standard continued pre-training (CPT) baseline. Figure~\ref{fig:fa_vs_sct} illustrates the training process differences between these two methods. Specifically, we first pre-train the entire model using our curated multilingual dataset, $\mathcal{D}_{la}$, followed by instruction tuning the model with English alignment data, $\mathcal{D}_{mix}$. Both stages are performed using full-parameter tuning. To ensure a fair comparison, we adopt the same customized tokenizer and embedding initialization techniques to exclude the influence of tokenization.

Figure~\ref{fig:fa_vs_cpt_sea} reports the results of CPT and \ouradapter based on \gemmatwo. CPT demonstrates significant declines in English proficiency across all model scales, indicating  catastrophic forgetting of knowledge. By contrast, \ouradapter exhibits only minor regressions, underscoring the advantages of model composition that reuses the knowledge of existing models. For zero-shot cross-lingual transfer, \ouradapter generally outperforms CPT on \belebele and \sib, both of which rely on knowledge transfer from English. However, on \flores, CPT achieves substantially better performance than \ouradapter. This outcome is unsurprising, as CPT employs full fine-tuning of the model, thereby using its greater capacity to learn from parallel data within the multilingual pre-training dataset $\mathcal{D}_{la}$. Besides, CPT experiences significant performance drop on GSM8K-NTL compared to \ouradapter (29.8\% \versus 40.4\% on 9B and 41.0\% \versus 44.9\% on 27B). Overall, \ouradapter effectively mitigates the knowledge-forgetting challenges inherent in conventional language adaptation approaches and offers a more robust solution for transferring capabilities across languages.

\section{Supplementary Analysis}\label{appendix:supplementary_analysis}

\begin{figure}[t]
    \setlength{\abovecaptionskip}{-0.0001cm}
    \setlength{\belowcaptionskip}{-0.35cm}
    \centering
    \includegraphics[width=\linewidth]{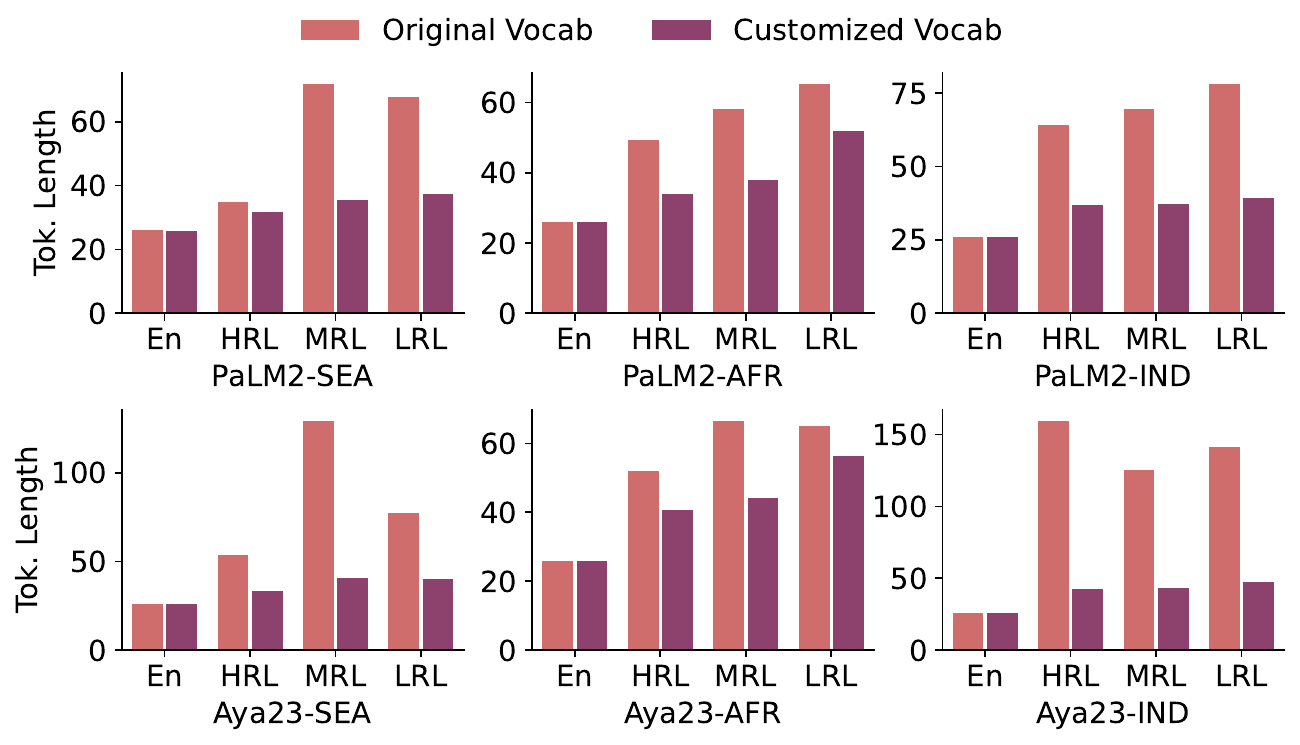}
    \vspace{-6mm}
    \caption{The tokenization comparison between using the vanilla and customized multilingual tokenizers on \gemmatwo. Tok. Length refers to the average number of tokens required to represent the same amount of texts.}
    \vspace{-4mm}
    \label{fig:fertility_comparison_palm_aya}
\end{figure}
\inlinetitle{Additional tokenization results on the customized vocabularies.}
We present additional results on tokenization fertility using customized vocabularies developed for \palmtwo and \aya. Figure~\ref{fig:fertility_comparison_palm_aya} shows similar patterns as those reported for \gemmatwo, where the fertility tokenization for both MRLs and LRLs shows a substantial decrease, while the English tokenization remains roughly unchanged. This effect is particularly pronounced in \aya, where we observe over $\times$3 reduction in fertility for \sea and \ind languages. In Table~\ref{tab:tokenization_qualitative_examples}, we also show a few tokenized examples for low-resource languages. We find that the customized tokenizer produces more meaningful tokens and avoids overtokenization.

\begin{figure}[t]
    \setlength{\abovecaptionskip}{-0.0001cm}
    \setlength{\belowcaptionskip}{-0.35cm}
    \centering
    \includegraphics[width=\linewidth]{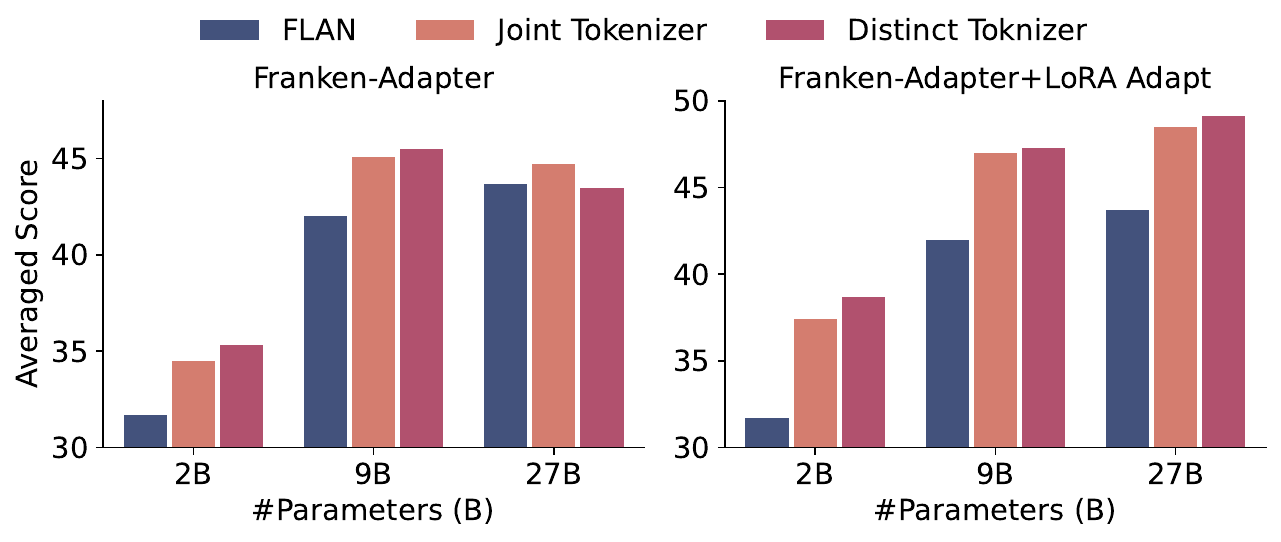}
    \vspace{-8mm}
    \caption{Performance comparison between employing distinct tokenizers for each language group and using a single tokenizer for all language groups (\ie Joint Tokenizer). The averaged performance of five tasks included in Table~\ref{tab:franken_adapter_results} is reported.}
    \vspace{-4mm}
    \label{fig:joint_vs_distinct_tokenizer}
\end{figure}
\inlinetitle{A joint customized tokenizer for all language groups is inferior to language-group specific tokenizers.}
In this ablation, we examine the effects of using distinct tokenizers for each language group in our experiments. For comparison, we create a joint tokenizer for the \sea, \afr, and \ind language groups by merging their tokenizer training corpora. The joint tokenizer maintains the same vocabulary size as the distinct tokenizers. We first conduct language adaptation with embedding tuning using the joint tokenizer on the combined pre-training dataset for the three language groups. This is followed by applying the standard \ouradapter method with LoRA-Adaptation. Importantly, this approach creates a single model, which is subsequently evaluated across all \sea, \afr, and \ind languages.

As shown in Figure~\ref{fig:joint_vs_distinct_tokenizer}, the use of a joint tokenizer significantly improves the cross-lingual transfer capabilities of \gemmatwo models across all model sizes. However, this approach underperforms compared to our primary setup, which employs distinct tokenizers for each language group. The performance difference is particularly pronounced in 2B models. This discrepancy arises because the use of a joint tokenizer reduces the capacity allocated to each language, as the total vocabulary size remains fixed. Consequently, this results in an approximate 10\% increase in fertility across languages at all resource levels, potentially hindering language learning. This finding aligns with the trends observed in Figure~\ref{fig:effects_customized_vocabulary}, where smaller models are more impacted by changes in fertility.

\begin{figure}[t]
    \setlength{\abovecaptionskip}{-0.0001cm}
    \setlength{\belowcaptionskip}{-0.35cm}
    \centering
    \includegraphics[width=\linewidth]{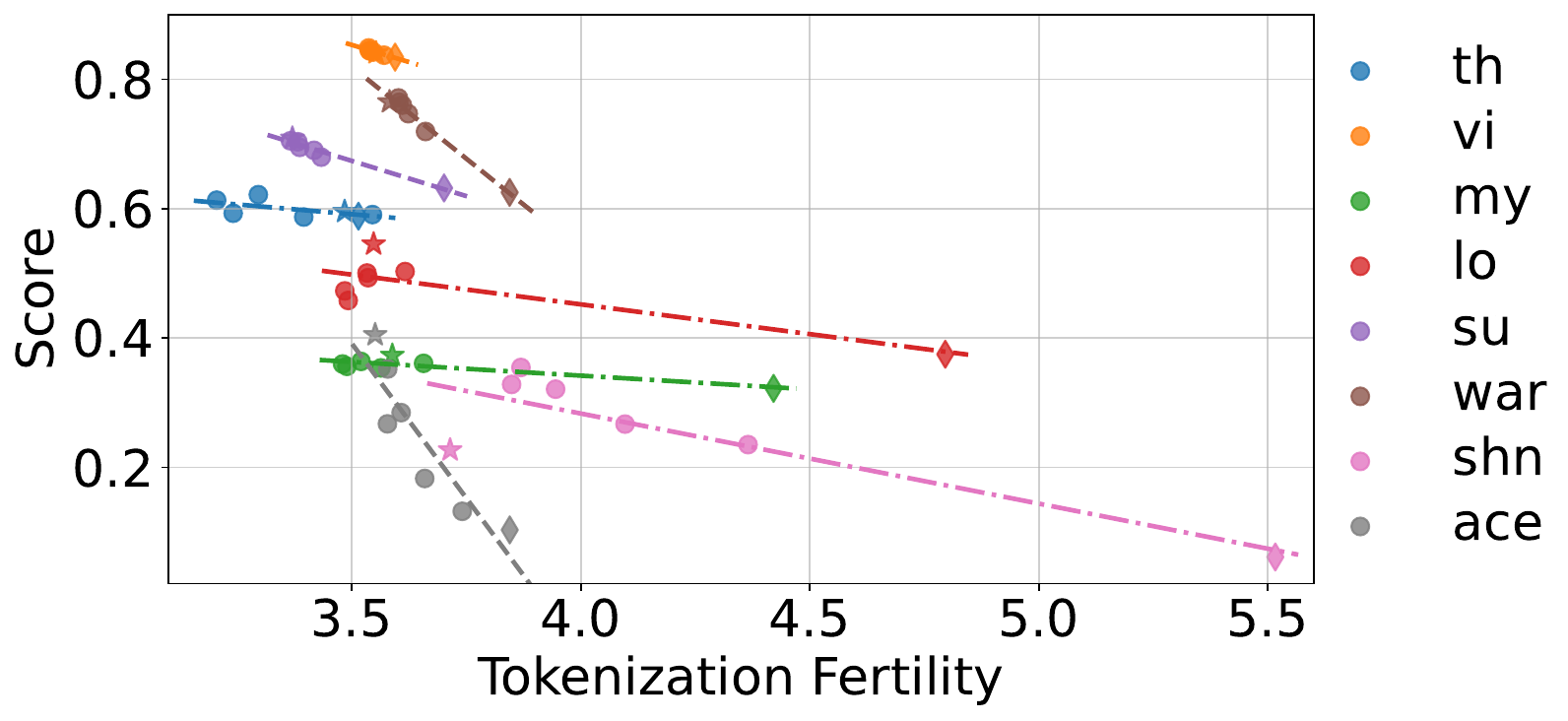}
    \vspace{-8mm}
    \caption{Correlation between the performance of language adaptation on \palmtwo-\texttt{XXS} with tokenizer fertility. Normalized \texttt{ChrF++} on \textsc{Flores}-\sea are reported. {$\bLozenge$} and {\pfix}  indicate the original and customized tokenizers in \palmtwo.}
    \vspace{-6mm}
    \label{fig:fertility_vs_performance_palm2}
\end{figure}
\inlinetitle{Tokenizer fertility is also inversely correlated to downstream performance in \palmtwo.}
We replicate the analysis shown in Figure~\ref{fig:fertility_vs_performance} using \palmtwo-\texttt{XXS}, with results shown in Figure~\ref{fig:fertility_vs_performance_palm2}. We observe patterns analogous to those reported for \gemmatwo-\texttt{2B}, wherein reduced tokenizer fertility is generally associated with improved downstream performance and this relationship is particularly pronounced in LRLs and languages written in non-Latin script.

\begin{figure}[t]
    \setlength{\abovecaptionskip}{-0.0001cm}
    \setlength{\belowcaptionskip}{-0.35cm}
    \centering
    \includegraphics[width=\linewidth]{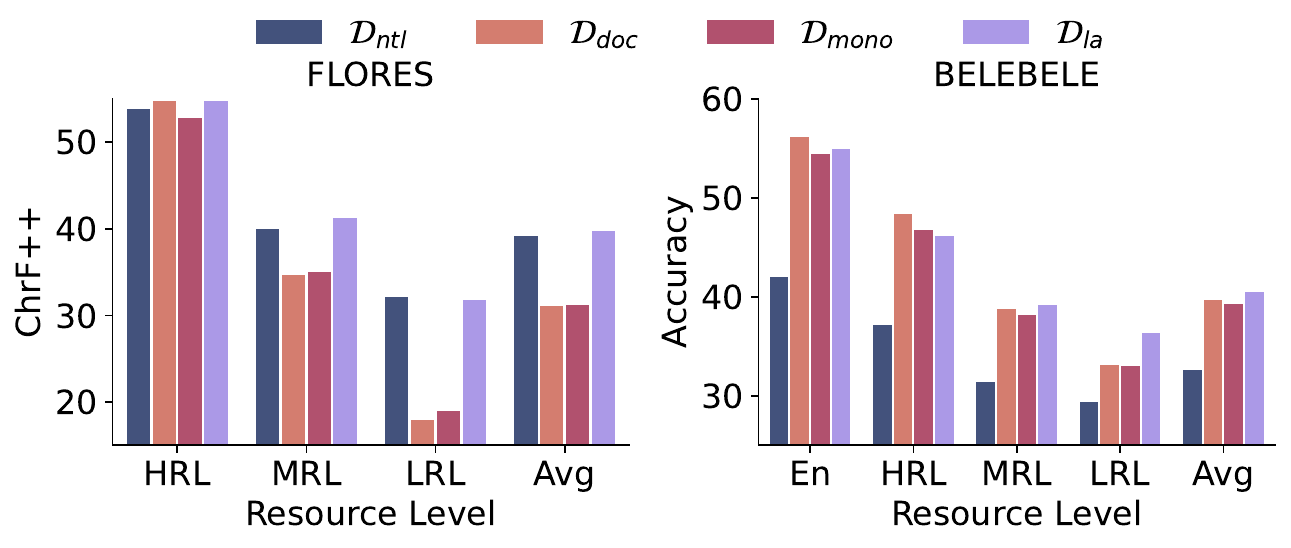}
    \vspace{-8mm}
    \caption{Pre-training data ablation for language adaptation. $\mathcal{D}_{ntl}$: multilingual data sampled from the NTL corpus; $\mathcal{D}_{doc}$: multilingual data sampled Wikipedia and mC4; $\mathcal{D}_{la}$: our final data mixture for language adaptation, \ie $\mathcal{D}_{la}=\mathcal{D}_{ntl}\cup\mathcal{D}_{doc}$; $\mathcal{D}_{mono}$: monolingual data by excluding parallel sentences from $\mathcal{D}_{la}$. \sea languages results on \gemmatwo-\texttt{2B} are reported.}
    \vspace{-4mm}
    \label{fig:data_ablation_on_pretraining}
\end{figure}
\inlinetitle{Long-tail NTL and document-level data are both important for language adaptation.}
We perform language adaptation on \gemmatwo-\texttt{2B} in \sea languages with different data mixture. Figure~\ref{fig:data_ablation_on_pretraining} reveals that improved performance in long-tail languages primarily stems from the inclusion of NTL data, while document-level data  plays a crucial role in preserving knowledge for high-resource languages. The significance of incorporating document-level data is further underscored by the results on \belebele, where the removal of $\mathcal{D}_{doc}$ leads to a substantial performance decline across all resource levels. This finding highlights the importance of $\mathcal{D}_{doc}$ in maintaining the ability of LLMs in processing various types of texts. In addition, excluding parallel data from the training mixture (\ie, $\mathcal{D}_{mono}$) leads to a significant decline in performance on translation tasks. Moreover, the performance of LRLs on \belebele also declines substantially, indicating the critical role of parallel data in enhancing tasks beyond translation through facilitated cross-lingual transfer~\citep{anil2023palm}.

\begin{figure}[t]
    \setlength{\abovecaptionskip}{-0.0001cm}
    \setlength{\belowcaptionskip}{-0.25cm}
    \centering
    \includegraphics[width=\linewidth]{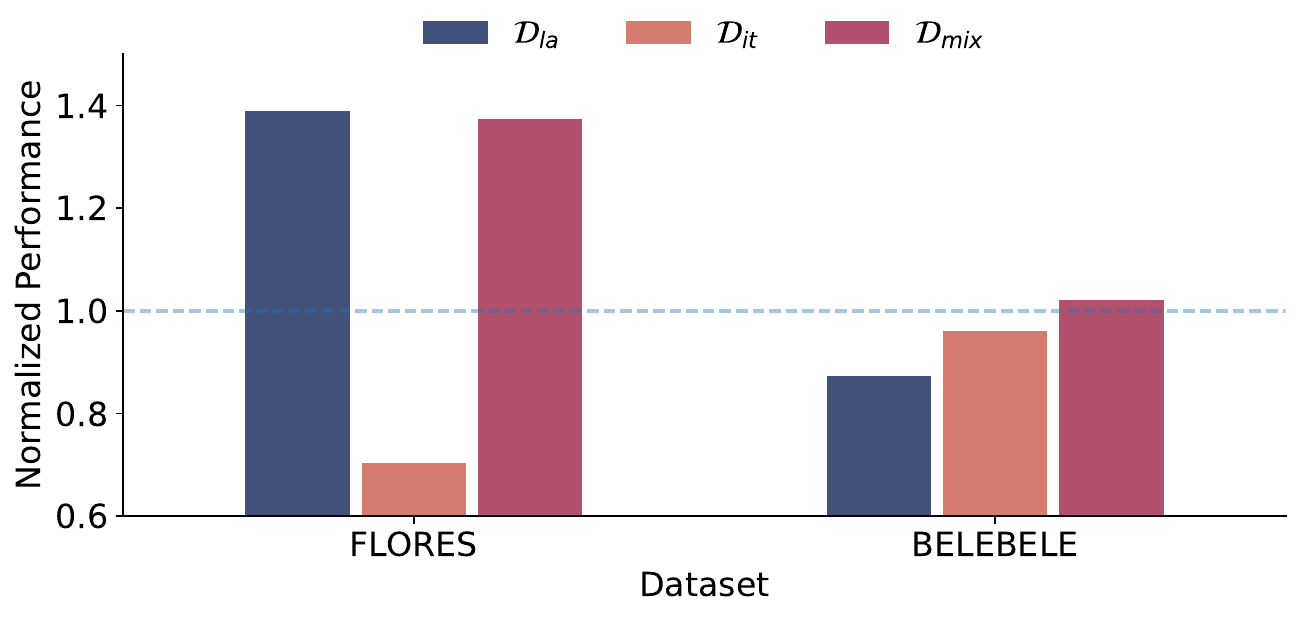}
    \vspace{-8mm}
    \caption{Training data ablation for Lora-Adaptation. $\mathcal{D}_{la}$: multilingual data for embedding tuning; $\mathcal{D}_{it}$: the FLAN mixture for instruction-tuning; $\mathcal{D}_{mix}$: a combination of equally sub-sampled $\mathcal{D}_{la}$ and $\mathcal{D}_{it}$. Averaged \sea language results (normalized score \versus \ouradapter) on \gemmatwo-\texttt{2B} are reported.}
    \vspace{-6mm}
    \label{fig:data_ablation_on_lora_adaptation}
\end{figure}
\inlinetitle{Both multilingual and instruction data are important for LoRA-Adaptation}
We investigate the impact of employing multilingual $\mathcal{D}_{la}$ and instruction-tuning data $\mathcal{D}_{it}$ in LoRA-Adaptation. As shown in Figure~\ref{fig:data_ablation_on_lora_adaptation}, the removal of either $\mathcal{D}_{la}$ or $\mathcal{D}_{it}$ harms the performance compared to the vanilla \ouradapter model on \flores or \belebele, while combining both datasets enables the adapted LLM to achieve the best overall results.

\begin{figure}[t]
    \setlength{\abovecaptionskip}{-0.0001cm}
    \setlength{\belowcaptionskip}{-0.25cm}
    \centering
    \includegraphics[width=\linewidth]{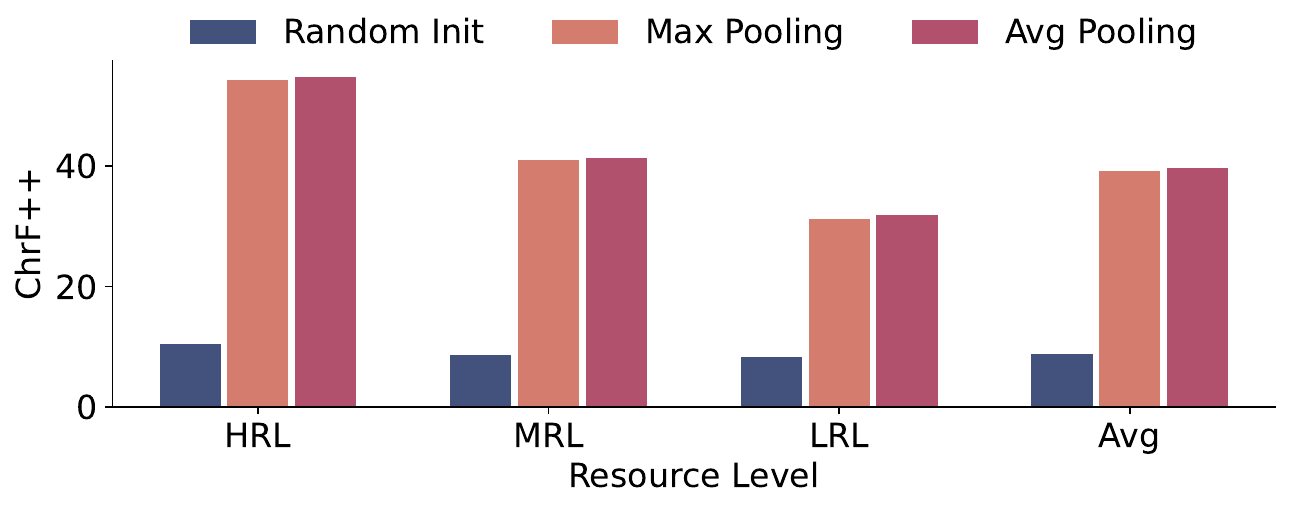}
    \vspace{-8mm}
    \caption{Ablations on embedding initialization methods. \textsc{Flores}-\sea language performance of language adaptation on \gemmatwo-\texttt{2B} is reported. \texttt{Max Pooling}: for each new token in the customized vocabulary, we use the original tokenizer to tokenize it and apply max pooling over the embeddings of the corresponding subtokens as the initialization.}
    \vspace{-5mm}
    \label{fig:ablation_on_emb_init}
\end{figure}
\inlinetitle{Employing the original embeddings for initialization is essential to language adaptation.}
In Figure~\ref{fig:ablation_on_emb_init}, we show that without initializing the customized embeddings using the original embeddings from the LLM, the language adaptation does not perform well at all, yielding \texttt{ChrF++} scores below 10 even for HRLs. In contrast, initializing with the original embeddings significantly improves the effectiveness of language adaptation, where the average pooling method slightly outperforms the max pooling variant.

\begin{figure}[t]
    \setlength{\abovecaptionskip}{-0.0001cm}
    \setlength{\belowcaptionskip}{-0.25cm}
    \centering
    \includegraphics[width=\linewidth]{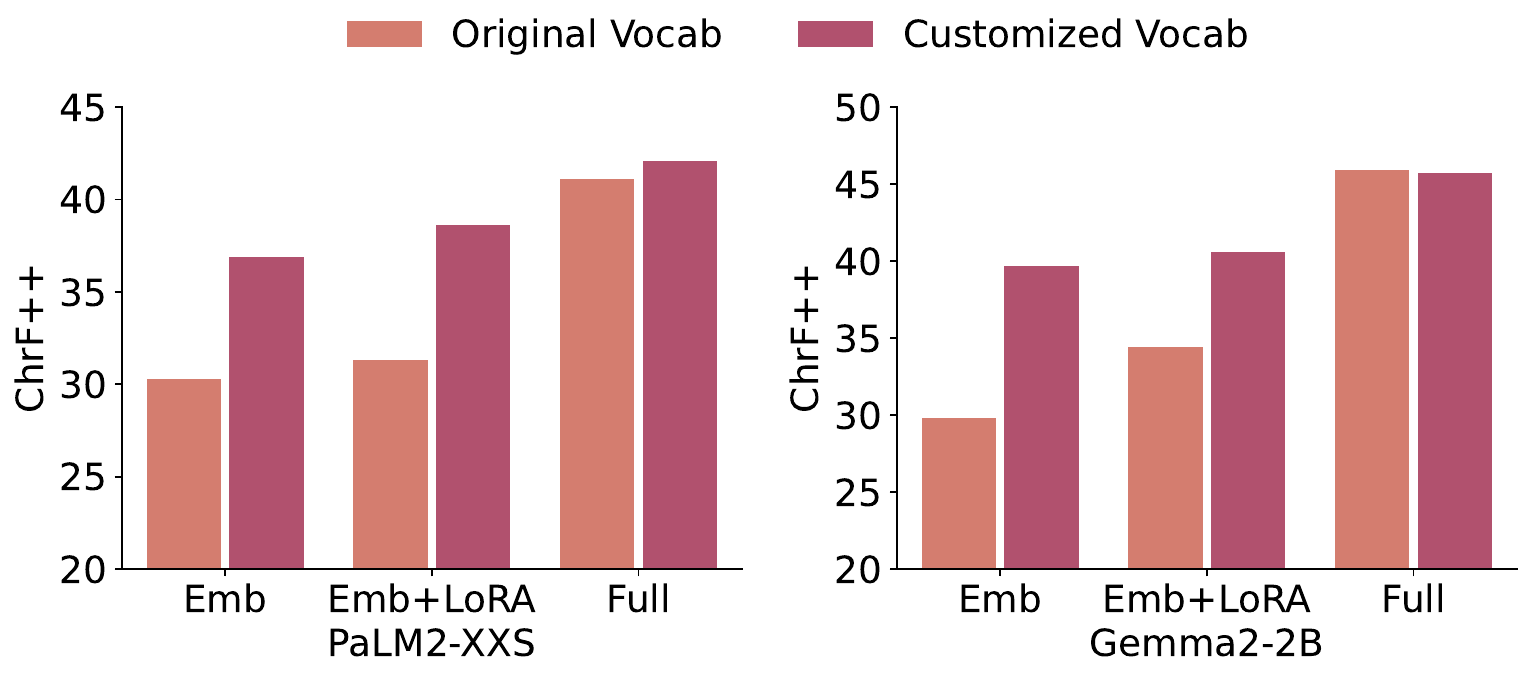}
    \vspace{-8mm}
    \caption{The effects of using customized vocabularies with different proportions of tuned parameters. The averaged score on \sea languages of \flores is reported.}
    \vspace{-6mm}
    \label{fig:emb_surgery_lora_fpt}
\end{figure}
\inlinetitle{The benefits of employing customized vocabulary decrease with more tuned parameters.}
We study the effects of employing customized embeddings with varying ratios of tuned parameters. As shown in Figure~\ref{fig:emb_surgery_lora_fpt}, the benefits of using customized embeddings diminish as the number of tuned parameters increases (Emb $\rightarrow$ Emb+LoRA $\rightarrow$ Full). Specifically, on \gemmatwo-\texttt{2B} model, customized embeddings show no advantage when full-parameter tuning is employed. This phenomenon arises because increasing the number of tuned parameters allocates greater model capacity for language adaptation, which simplifies the adaptation process compared to relying solely on embedding tuning. Nonetheless, full-parameter tuning could exacerbate the problem of catastrophic forgetting, while embedding tuning provides a safer alternative. Furthermore, the use of customized embeddings amplifies the advantages of embedding tuning, making it a promising technique.

\begin{figure}[t]
    \setlength{\abovecaptionskip}{-0.0001cm}
    \setlength{\belowcaptionskip}{-0.35cm}
    \centering
    \includegraphics[width=0.85\linewidth]{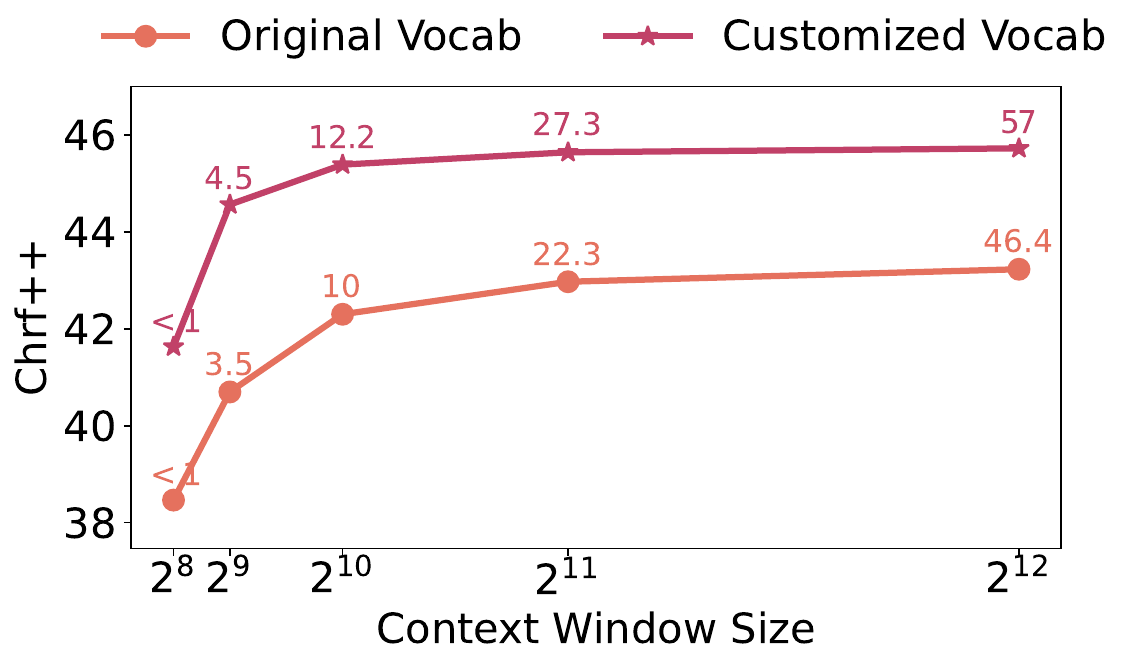}
    \vspace{-5mm}
    \caption{N-shot prompting results on \textsc{Flores}-\sea by varying context windows. The maximum average number of shots that fit in each window is indicated.}
    \vspace{-4mm}
    \label{fig:fixed_context_window_comparison}
\end{figure}
\inlinetitle{Customized vocabulary is more inference efficient.}
Figure~\ref{fig:fixed_context_window_comparison} indicates that employing the customized vocabulary can consistently leads to better results under varying context window sizes. Additionally, the customized vocabulary makes the model more inference and prompt efficient, achieving the same performance while needing roughly 10$\times$ fewer in-context learning examples examples and tokens.

\begin{figure}[t]
    \setlength{\abovecaptionskip}{-0.0001cm}
    \setlength{\belowcaptionskip}{-0.35cm}
    \centering
    \includegraphics[width=\linewidth]{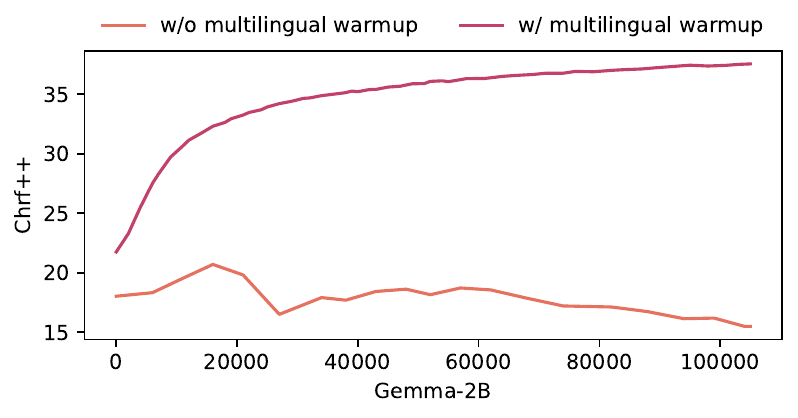}
    \vspace{-8mm}
    \caption{Learning curves of language adaptation on LLMs with limited multilingual abilities. Averaged results on \sea languages of \flores are reported.}
    \vspace{-6mm}
    \label{fig:limited_multilinguality_llms}
\end{figure}

\inlinetitle{Multilingual warmup is necessary for language adaptation on LLMs with limited multilingual abilities.}
We investigate whether embedding tuning with customized embeddings is a universal technique for the language adaptation of any types of LLMs. As shown in Figure~\ref{fig:limited_multilinguality_llms}, simply performing language adaptation on \texttt{Gemma-2B}, a model with very limited multilingual capabilities, does not successfully adapt it to various languages. In contrast, when a multilingual continued pre-training process is conducted prior to language adaptation, where the document-level data $\mathcal{D}_{doc}$ is used to warm up the LLM, we observe consistent improvements throughout the training process. This suggests that, for LLMs, a good initial multilingual ability is essential for the success of language adaptation.

\section{Results by Language}\label{appendix:detailed_lang_results}
Table~\ref{tab:dataset_by_language} presents an overview of languages available across our evaluation benchmarks. We show the per-language results on each task for both language adaptation
(Tables~\ref{tab:flores_results} --~\ref{tab:belebele_results})
and \ouradapter (Tables~\ref{tab:franken_adapter_belebele_results} --~\ref{tab:franken_adapter_xorqa_results}).

\onecolumn

\begin{table*}[t]
\setlength{\belowcaptionskip}{-0.2cm}
\setlength{\tabcolsep}{5pt}
\footnotesize
\centering
\vspace{-4mm}
\caption{Qualitative examples for comparing tokenization using the customized vocabulary against Gemma's original one. We find that the customized tokenizer reduces overtokenization without affecting English tokenization. Sentences have been word-segmented to simplify presentation.}
\begin{tabular}{l}
    \includegraphics[width=0.95\linewidth]{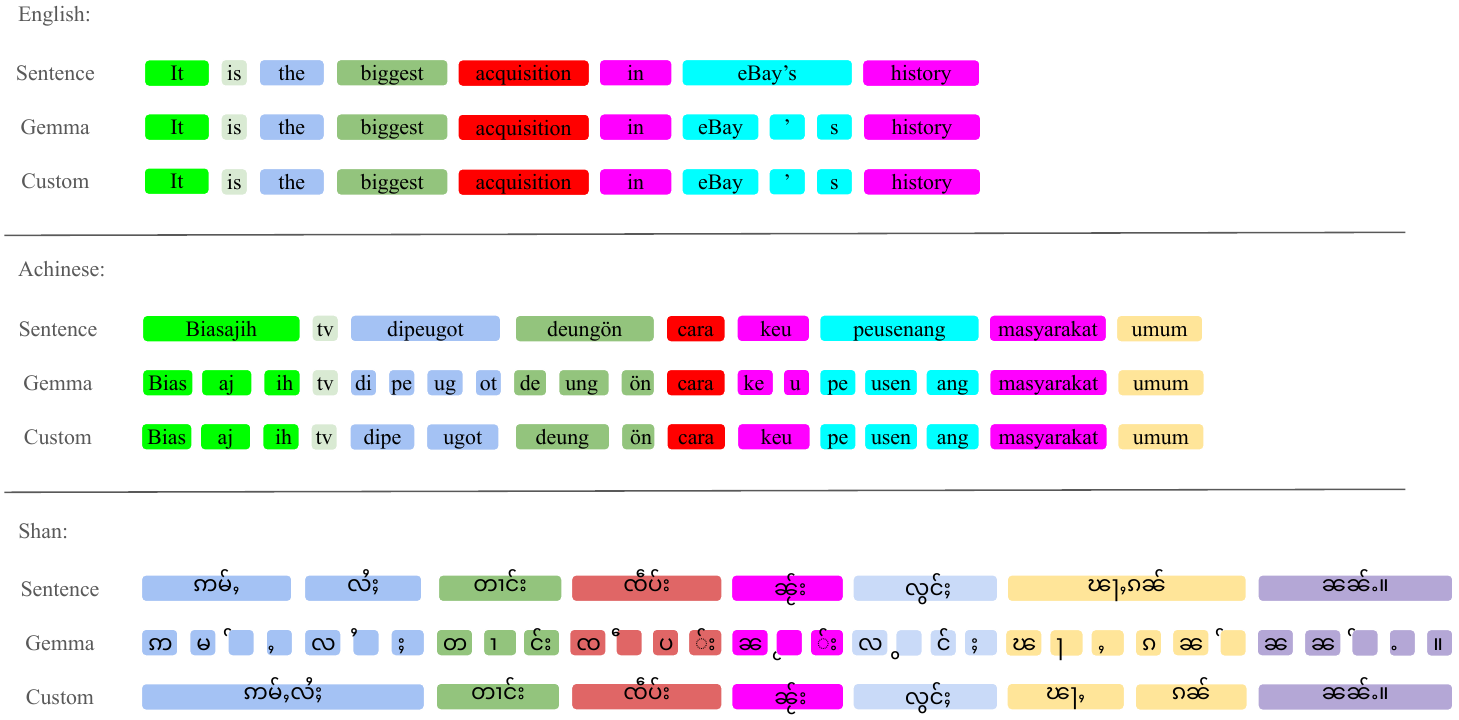}
    \end{tabular}
    \vspace{-4mm}
\label{tab:tokenization_qualitative_examples}
\end{table*}

\begin{small}
    \footnotesize
    \setlength{\tabcolsep}{4pt}
    \setlength{\LTcapwidth}{0.98\linewidth}
    \begin{longtable}{l|c|cccccc}
    \caption{Overview of languages included in each of our evaluation benchmark. $\text{H}, \text{M}, \text{L}$ indicate high, medium, and low-resource languages, respectively.} \\
    \toprule
        \bf Language name & ISO code & \belebele & \sib & \flores & \xorqain & \xsumin & GSM8K-NTL \\
        \midrule
        \endhead
        
        English & eng\_Latn & \cmark & \cmark &   & \cmark & \cmark &   \\
        \midrule

        \multicolumn{8}{c}{\bf\sea} \\
        \midrule
        Achinese (Arabic) & ace\_Arab$^\text{L}$ &   & \cmark & \cmark &   &   &   \\
        Achinese & ace\_Latn$^\text{L}$ &   & \cmark & \cmark &   &   &   \\
        Balinese & ban\_Latn$^\text{L}$ &   & \cmark & \cmark &   &   &   \\
        Betawi & bew\_Latn$^\text{M}$ &   &   &   &   &   & \cmark \\
        Banjar (Arabic) & bjn\_Arab$^\text{L}$ &   & \cmark & \cmark &   &   &   \\
        Banjar & bjn\_Latn$^\text{L}$ &   & \cmark & \cmark &   &   &   \\
\end{longtable}
\end{small}

\begin{small}
    \setlength{\LTcapwidth}{0.98\linewidth}
    \begin{longtable}{c| cc|cc| cc|cc|cc| cc|cc}
    \caption{Language Adaptation results on \flores \textsc{En-Xx} with 5-shot in-context learning. \pt: Pre-training; \es: Language Adaptation.} \\
    \toprule
        \multicolumn{15}{c}{\bf Language Adaptation Results on \flores \textsc{En-Xx}} \\
        \midrule
        \bf Model & \multicolumn{4}{c|}{\bf \palmtwo} & \multicolumn{6}{c|}{\bf \gemmatwo} & \multicolumn{4}{c}{\bf \aya} \\
        \cmidrule(lr){2-5}  \cmidrule(lr){6-11}  \cmidrule(lr){12-15}
        \bf Size & \multicolumn{2}{c|}{\bf XXS} & \multicolumn{2}{c|}{\bf S} & \multicolumn{2}{c|}{\bf 2B} & \multicolumn{2}{c|}{\bf 9B} & \multicolumn{2}{c|}{\bf 27B} & \multicolumn{2}{c|}{\bf 8B} & \multicolumn{2}{c}{\bf 35B} \\
        \cmidrule(lr){2-3} \cmidrule(lr){4-5} \cmidrule(lr){6-7} \cmidrule(lr){8-9} \cmidrule(lr){10-11} \cmidrule(lr){12-13} \cmidrule(lr){14-15}
        \bf Variant & \pt & \es & \pt & \es  & \pt & \es  & \pt & \es  & \pt & \es  & \pt & \es  & \pt & \es \\
        \midrule
        \endhead
        
        \multicolumn{15}{c}{\bf \sea} \\
        \midrule
        ind\_Latn & 65.9 & 65.0 & 70.7 & 69.1 & 64.4 & 64.6 & 68.9 & 68.3 & 69.9 & 68.5 & 73.2 & 69.5 & 71.9 & 68.6 \\
        tha\_Thai & 42.9 & 40.9 & 50.3 & 49.8 & 35.6 & 35.3 & 44.7 & 43.8 & 47.4 & 45.6 & 20.3 & 33.9 & 28.4 & 37.0 \\
        vie\_Latn & 56.4 & 56.4 & 61.7 & 61.6 & 52.1 & 54.2 & 58.7 & 58.3 & 60.0 & 59.8 & 65.7 & 63.5 & 64.9 & 62.3 \\
        zsm\_Latn & 63.0 & 64.6 & 68.0 & 67.4 & 58.4 & 63.1 & 64.6 & 65.6 & 65.6 & 65.9 & 54.2 & 64.6 & 57.4 & 64.9 \\
        tgl\_Latn & 54.2 & 55.1 & 62.1 & 61.4 & 48.4 & 55.2 & 57.8 & 58.3 & 60.4 & 59.7 & 36.7 & 55.3 & 47.4 & 57.9 \\
        mya\_Mymr & 22.4 & 26.9 & 40.9 & 37.4 & 12.9 & 27.5 & 25.4 & 33.5 & 30.1 & 33.3 & 6.5 & 23.4 & 7.3 & 29.5 \\
        lao\_Laoo & 30.6 & 37.8 & 48.8 & 47.9 & 9.5 & 39.0 & 26.7 & 42.9 & 33.7 & 42.1 & 4.0 & 27.0 & 10.2 & 41.1 \\
\end{longtable}
\end{small}

\begin{small}
    \setlength{\LTcapwidth}{0.98\linewidth}
    \begin{longtable}{c| cccc| cc|cc| cc|cc}
    \caption{Language Adaptation results on GSM8K-NTL with 5-shot in-context learning. \pt: Pre-training; \es: Language Adaptation.} \\
    \toprule
        \multicolumn{13}{c}{\bf Language Adaptation Results on GSM8K-NTL} \\
        \midrule
        \bf Model & \multicolumn{4}{c|}{\bf \palmtwo} & \multicolumn{4}{c|}{\bf \gemmatwo} & \multicolumn{4}{c}{\bf \aya} \\
        \cmidrule(lr){2-5}  \cmidrule(lr){6-9}  \cmidrule(lr){10-13}
        \bf Size & \multicolumn{4}{c|}{\bf S} & \multicolumn{2}{c|}{\bf 9B} & \multicolumn{2}{c|}{\bf 27B} & \multicolumn{2}{c|}{\bf 8B} & \multicolumn{2}{c}{\bf 35B} \\
        \cmidrule(lr){2-5} \cmidrule(lr){6-7} \cmidrule(lr){8-9} \cmidrule(lr){10-11} \cmidrule(lr){12-13}
        \bf Variant & \pt & \textsc{Ntl} & CALM & \es & \pt & \es & \pt & \es & \pt & \es & \pt & \es \\
        \midrule
        \endhead
        asm\_Beng & 5.2 & 4.0 & 9.2 & 17.2 & 14.8 & 29.2 & 15.6 & 29.6 & 0.8 & 3.6 & 2.0 & 10.4 \\
        bew\_Latn & 33.6 & 33.6 & 34.8 & 33.6 & 33.6 & 42.8 & 48.4 & 50.4 & 33.6 & 31.6 & 45.6 & 42.4 \\
        bho\_Deva & 23.6 & 22.8 & 29.2 & 26.0 & 19.6 & 35.6 & 29.2 & 40.0 & 14.8 & 12.0 & 21.6 & 24.0 \\
        doi\_Deva & 17.2 & 21.6 & 22.4 & 26.8 & 12.0 & 33.2 & 17.2 & 30.0 & 9.6 & 9.6 & 18.8 & 24.0 \\
        div\_Thaa & 11.2 & 13.2 & 14.8 & 19.2 & 6.8 & 26.8 & 8.0 & 21.2 & 1.6 & 3.2 & 1.2 & 14.0 \\
        dzo\_Tibt & 0.8 & 0.0 & 0.4 & 7.6 & 1.2 & 15.2 & 2.8 & 8.8 & 0.4 & 0.0 & 0.0 & 1.2 \\
        efi\_Latn & 14.8 & 14.0 & 18.0 & 22.0 & 10.8 & 26.0 & 16.4 & 31.2 & 4.0 & 5.2 & 4.0 & 18.4 \\
        gom\_Deva & 22.4 & 22.8 & 25.2 & 30.0 & 18.8 & 36.4 & 26.8 & 33.6 & 8.4 & 8.0 & 9.2 & 20.4 \\
\end{longtable}
\end{small}

\begin{small}
    \setlength\LTcapwidth\textwidth
    \begin{longtable}{c |cc|cc| cc|cc|cc| cc|cc}
    \caption{Language Adaptation results on \belebele with 5-shot in-context learning. \pt: Pre-training; \es: Language Adaptation.} \\
    \toprule
        \multicolumn{15}{c}{\bf Language Adaptation Results on \belebele } \\
        \midrule
        \bf Model & \multicolumn{4}{c|}{\bf \palmtwo} & \multicolumn{6}{c|}{\bf \gemmatwo} & \multicolumn{4}{c}{\bf \aya} \\
        \cmidrule(lr){2-5}  \cmidrule(lr){6-11}  \cmidrule(lr){12-15}
        \bf Size & \multicolumn{2}{c|}{\bf XXS} & \multicolumn{2}{c|}{\bf S} & \multicolumn{2}{c|}{\bf 2B} & \multicolumn{2}{c|}{\bf 9B} & \multicolumn{2}{c|}{\bf 27B} & \multicolumn{2}{c|}{\bf 8B} & \multicolumn{2}{c}{\bf 35B} \\
        \cmidrule(lr){2-3} \cmidrule(lr){4-5} \cmidrule(lr){6-7} \cmidrule(lr){8-9} \cmidrule(lr){10-11} \cmidrule(lr){12-13} \cmidrule(lr){14-15}
        \bf Variant & \pt & \es & \pt & \es  & \pt & \es  & \pt & \es  & \pt & \es  & \pt & \es  & \pt & \es \\
        \midrule
        \endhead
        
        \multicolumn{15}{c}{\bf \sea} \\
        \midrule
        eng\_Latn & 23.8 & 28.0 & 85.9 & 86.8 & 61.2 & 54.9 & 90.6 & 91.3 & 92.3 & 91.6 & 82.4 & 80.0 & 91.4 & 89.6 \\
        ind\_Latn & 25.9 & 26.4 & 80.7 & 80.0 & 49.3 & 45.6 & 84.2 & 84.1 & 87.8 & 87.0 & 73.7 & 71.7 & 84.2 & 80.8 \\
        tha\_Thai & 27.2 & 26.6 & 77.9 & 78.2 & 44.1 & 40.0 & 78.4 & 76.7 & 82.8 & 81.8 & 44.6 & 56.9 & 65.0 & 68.2 \\
        vie\_Latn & 25.9 & 25.8 & 82.1 & 80.6 & 46.3 & 43.9 & 84.3 & 84.7 & 88.1 & 86.6 & 73.3 & 71.3 & 84.2 & 81.1 \\
        mya\_Mymr & 25.8 & 25.9 & 75.7 & 71.2 & 32.6 & 38.1 & 68.0 & 72.6 & 65.7 & 75.2 & 27.9 & 48.8 & 36.2 & 61.7 \\
        lao\_Laoo & 29.1 & 27.0 & 70.3 & 67.7 & 28.7 & 36.7 & 60.6 & 69.6 & 59.1 & 73.4 & 27.4 & 47.1 & 29.4 & 61.6 \\
        khm\_Khmr & 27.1 & 26.9 & 77.4 & 70.1 & 30.6 & 38.6 & 64.7 & 72.1 & 65.2 & 75.6 & 25.9 & 47.0 & 33.8 & 64.4 \\
\end{longtable}
\end{small}

\begin{small}
    \fasize
    \setlength{\tabcolsep}{2.9pt}
    \setlength{\LTcapwidth}{0.98\linewidth}
    \begin{longtable}{c| ccc|ccc| ccc|ccc|ccc| ccc|ccc}
    \caption{\ouradapter results on \belebele with zero-shot prompting. \flan: LLMs instruction-tuned on \textsc{Flan} mixture; \franken: \ouradapter; \fala: \ouradapter with LoRA Adaptation.} \\
    \toprule
        \multicolumn{22}{c}{\bf \ouradapter Results on \belebele} \\
        \midrule
        \bf Model & \multicolumn{6}{c|}{\bf \palmtwo} & \multicolumn{9}{c|}{\bf \gemmatwo} & \multicolumn{6}{c}{\bf \aya} \\
        \cmidrule(lr){2-7}  \cmidrule(lr){8-16}  \cmidrule(lr){17-22}
        \bf Size & \multicolumn{3}{c|}{\bf XXS} & \multicolumn{3}{c|}{\bf S} & \multicolumn{3}{c|}{\bf 2B} & \multicolumn{3}{c|}{\bf 9B} & \multicolumn{3}{c|}{\bf 27B} & \multicolumn{3}{c|}{\bf 8B} & \multicolumn{3}{c}{\bf 35B} \\
        \cmidrule(lr){2-4} \cmidrule(lr){5-7} \cmidrule(lr){8-10} \cmidrule(lr){11-13} \cmidrule(lr){14-16} \cmidrule(lr){17-19} \cmidrule(lr){20-22}
        \bf Variant & \flan & \franken & \fala & \flan & \franken & \fala & \flan & \franken & \fala & \flan & \franken & \fala & \flan & \franken & \fala & \flan & \franken & \fala & \flan & \franken & \fala \\
        \midrule
        \endhead
        
        \multicolumn{22}{c}{\bf \sea} \\
        \midrule
        eng\_Latn & 80.9 & 76.0 & 78.9 & 95.7 & 95.9 & 95.8 & 87.7 & 85.9 & 87.4 & 93.9 & 93.9 & 93.6 & 95.4 & 95.0 & 94.0 & 88.9 & 83.4 & 86.4 & 94.2 & 87.0 & 93.2 \\
        ind\_Latn & 74.0 & 72.3 & 73.4 & 91.8 & 91.9 & 92.0 & 78.7 & 76.1 & 75.8 & 88.8 & 87.6 & 88.2 & 90.4 & 89.7 & 88.3 & 82.0 & 74.3 & 74.3 & 89.6 & 72.7 & 86.3 \\
        tha\_Thai & 66.8 & 65.2 & 65.8 & 88.3 & 88.3 & 88.1 & 70.0 & 67.7 & 68.3 & 83.6 & 83.3 & 82.7 & 85.7 & 85.1 & 85.3 & 57.2 & 59.3 & 63.7 & 73.9 & 60.7 & 79.4 \\
        vie\_Latn & 74.1 & 72.0 & 72.0 & 92.6 & 92.7 & 92.6 & 75.6 & 74.0 & 74.9 & 89.7 & 89.2 & 88.4 & 90.0 & 89.7 & 88.8 & 81.9 & 75.9 & 74.6 & 90.2 & 75.6 & 87.6 \\
        mya\_Mymr & 52.0 & 56.1 & 56.7 & 86.8 & 85.1 & 85.7 & 36.3 & 62.1 & 60.8 & 69.7 & 76.1 & 77.9 & 65.2 & 78.2 & 78.2 & 34.0 & 51.1 & 56.2 & 38.1 & 49.2 & 69.0 \\
        lao\_Laoo & 44.9 & 55.9 & 57.3 & 81.2 & 82.3 & 83.1 & 43.3 & 60.9 & 61.4 & 61.4 & 77.3 & 75.9 & 59.7 & 77.1 & 77.3 & 31.8 & 50.6 & 51.9 & 38.8 & 49.2 & 66.4 \\
        khm\_Khmr & 51.9 & 52.6 & 55.8 & 86.3 & 84.6 & 86.7 & 45.1 & 65.0 & 62.6 & 69.2 & 78.6 & 78.4 & 68.6 & 81.1 & 79.4 & 26.8 & 46.9 & 51.4 & 43.6 & 51.7 & 71.2 \\
\end{longtable}
\end{small}

\begin{small}
    \fasize
    \setlength{\tabcolsep}{2.9pt}
    \setlength{\LTcapwidth}{0.98\linewidth}
    \begin{longtable}{c| ccc|ccc| ccc|ccc|ccc| ccc|ccc}
    \caption{\ouradapter results on \sib with zero-shot prompting. \flan: LLMs instruction-tuned on \textsc{Flan} mixture; \franken: \ouradapter; \fala: \ouradapter with LoRA Adaptation.} \\
    \toprule
        \multicolumn{22}{c}{\bf \ouradapter Results on \sib} \\
        \midrule
        \bf Model & \multicolumn{6}{c|}{\bf \palmtwo} & \multicolumn{9}{c|}{\bf \gemmatwo} & \multicolumn{6}{c}{\bf \aya} \\
        \cmidrule(lr){2-7}  \cmidrule(lr){8-16}  \cmidrule(lr){17-22}
        \bf Size & \multicolumn{3}{c|}{\bf XXS} & \multicolumn{3}{c|}{\bf S} & \multicolumn{3}{c|}{\bf 2B} & \multicolumn{3}{c|}{\bf 9B} & \multicolumn{3}{c|}{\bf 27B} & \multicolumn{3}{c|}{\bf 8B} & \multicolumn{3}{c}{\bf 35B} \\
        \cmidrule(lr){2-4} \cmidrule(lr){5-7} \cmidrule(lr){8-10} \cmidrule(lr){11-13} \cmidrule(lr){14-16} \cmidrule(lr){17-19} \cmidrule(lr){20-22}
        \bf Variant & \flan & \franken & \fala & \flan & \franken & \fala & \flan & \franken & \fala & \flan & \franken & \fala & \flan & \franken & \fala & \flan & \franken & \fala & \flan & \franken & \fala \\
        \midrule
        \endhead
        
        \multicolumn{22}{c}{\bf \sea} \\
        \midrule
        eng\_Latn & 85.3 & 83.8 & 82.4 & 76.0 & 76.5 & 77.9 & 82.4 & 82.4 & 79.9 & 83.3 & 81.4 & 80.4 & 83.3 & 81.9 & 82.4 & 82.8 & 80.8 & 79.8 & 82.8 & 80.8 & 79.8 \\
        ace\_Arab & 27.9 & 51.5 & 47.1 & 53.4 & 70.1 & 67.6 & 37.7 & 45.6 & 50.5 & 53.4 & 64.2 & 65.2 & 46.6 & 62.7 & 67.6 & 37.4 & 46.5 & 38.4 & 39.4 & 51.5 & 65.7 \\
        ace\_Latn & 63.2 & 78.4 & 71.6 & 71.1 & 80.9 & 77.0 & 62.3 & 68.6 & 74.0 & 73.5 & 76.5 & 79.9 & 71.1 & 73.0 & 81.9 & 63.6 & 74.7 & 61.6 & 68.7 & 73.7 & 76.8 \\
        ban\_Latn & 75.0 & 80.9 & 76.5 & 73.0 & 77.0 & 77.9 & 68.6 & 74.0 & 76.0 & 81.4 & 78.4 & 78.9 & 81.9 & 77.5 & 81.4 & 70.7 & 75.8 & 66.7 & 71.7 & 70.7 & 75.8 \\
        bjn\_Arab & 17.2 & 43.6 & 47.1 & 55.4 & 63.7 & 62.3 & 31.4 & 42.2 & 48.0 & 47.1 & 59.8 & 62.3 & 44.1 & 59.3 & 64.7 & 36.4 & 42.4 & 39.4 & 40.4 & 48.5 & 58.6 \\
        bjn\_Latn & 73.0 & 78.9 & 75.0 & 76.5 & 79.4 & 81.4 & 67.6 & 71.6 & 71.1 & 80.4 & 78.9 & 79.9 & 77.9 & 78.9 & 79.9 & 72.7 & 73.7 & 60.6 & 73.7 & 70.7 & 71.7 \\
        bug\_Latn & 57.8 & 67.2 & 71.1 & 63.7 & 74.0 & 72.5 & 57.4 & 62.3 & 64.2 & 62.7 & 69.6 & 69.1 & 63.2 & 68.6 & 70.1 & 55.6 & 66.7 & 54.5 & 65.7 & 69.7 & 75.8 \\
\end{longtable}
\end{small}

\begin{small}
    \fasize
    \setlength{\tabcolsep}{2.9pt}
    \setlength{\LTcapwidth}{0.98\linewidth}
    \begin{longtable}{c| ccc|ccc| ccc|ccc|ccc| ccc|ccc}
    \caption{\ouradapter results on \flores with zero-shot prompting. \flan: LLMs instruction-tuned on \textsc{Flan} mixture; \franken: \ouradapter; \fala: \ouradapter with LoRA Adaptation.} \\
    \toprule
        \multicolumn{22}{c}{\bf \ouradapter Results on \flores} \\
        \midrule
        \bf Model & \multicolumn{6}{c|}{\bf \palmtwo} & \multicolumn{9}{c|}{\bf \gemmatwo} & \multicolumn{6}{c}{\bf \aya} \\
        \cmidrule(lr){2-7}  \cmidrule(lr){8-16}  \cmidrule(lr){17-22}
        \bf Size & \multicolumn{3}{c|}{\bf XXS} & \multicolumn{3}{c|}{\bf S} & \multicolumn{3}{c|}{\bf 2B} & \multicolumn{3}{c|}{\bf 9B} & \multicolumn{3}{c|}{\bf 27B} & \multicolumn{3}{c|}{\bf 8B} & \multicolumn{3}{c}{\bf 35B} \\
        \cmidrule(lr){2-4} \cmidrule(lr){5-7} \cmidrule(lr){8-10} \cmidrule(lr){11-13} \cmidrule(lr){14-16} \cmidrule(lr){17-19} \cmidrule(lr){20-22}
        \bf Variant & \flan & \franken & \fala & \flan & \franken & \fala & \flan & \franken & \fala & \flan & \franken & \fala & \flan & \franken & \fala & \flan & \franken & \fala & \flan & \franken & \fala \\
        \midrule
        \endhead
        
        \multicolumn{22}{c}{\bf \sea} \\
        \midrule
        ind\_Latn & 65.1 & 63.9 & 65.7 & 68.4 & 66.9 & 67.0 & 61.5 & 54.5 & 61.6 & 65.9 & 64.9 & 65.4 & 65.6 & 64.5 & 65.6 & 68.7 & 67.1 & 66.6 & 67.0 & 54.2 & 67.3 \\
        tha\_Thai & 44.0 & 38.4 & 42.3 & 48.2 & 40.6 & 44.6 & 37.4 & 37.6 & 35.1 & 42.4 & 42.4 & 41.9 & 43.6 & 41.8 & 42.4 & 23.4 & 29.0 & 35.1 & 29.3 & 18.9 & 39.2 \\
        vie\_Latn & 57.0 & 56.4 & 57.2 & 60.5 & 59.7 & 59.6 & 49.4 & 49.9 & 50.3 & 57.1 & 57.3 & 56.5 & 57.2 & 56.8 & 58.0 & 61.8 & 62.4 & 60.0 & 61.8 & 57.3 & 60.9 \\
        zsm\_Latn & 62.6 & 63.8 & 64.8 & 67.9 & 66.8 & 67.0 & 56.6 & 59.6 & 60.7 & 63.1 & 65.1 & 64.5 & 64.6 & 63.6 & 65.1 & 50.5 & 62.6 & 63.9 & 54.2 & 46.0 & 64.9 \\
        tgl\_Latn & 55.2 & 55.4 & 56.4 & 59.5 & 59.1 & 59.4 & 51.4 & 52.1 & 54.2 & 54.9 & 56.6 & 57.0 & 55.2 & 54.1 & 57.7 & 37.1 & 52.0 & 55.7 & 47.2 & 50.2 & 56.7 \\
        mya\_Mymr & 9.3 & 2.6 & 21.8 & 37.1 & 29.6 & 32.8 & 12.9 & 13.3 & 23.8 & 14.7 & 33.2 & 27.8 & 16.0 & 16.9 & 31.1 & 3.6 & 9.3 & 24.3 & 7.0 & 20.4 & 28.1 \\
        lao\_Laoo & 29.9 & 30.8 & 39.9 & 40.5 & 27.6 & 38.0 & 27.3 & 26.7 & 34.3 & 24.3 & 40.9 & 40.1 & 25.0 & 1.2 & 41.7 & 9.6 & 26.0 & 31.1 & 17.0 & 25.4 & 37.5 \\
\end{longtable}
\end{small}

\begin{small}
    \fasize
    \setlength{\tabcolsep}{5pt}
    \setlength{\LTcapwidth}{0.98\linewidth}
    \begin{longtable}{c|cc|cc|cc|cc|cc|cc|cc|cc|cc}
    \caption{\ouradapter results on GSM8K-NTL with zero-shot prompting. \flan: LLMs instruction-tuned on \textsc{Flan} mixture; \textsc{It}: LLMs aligned with supervised fine-tuning and reinforcement learning with human feedback; \textsc{Mh}: LLMs instruction-tuned on the \texttt{WebInstruct} math dataset; \franken: \ouradapter;} \\
    \toprule
        \multicolumn{19}{c}{\bf \ouradapter Results on GSM8K-NTL} \\
        \midrule
        \bf Model & \multicolumn{6}{c}{\bf \gemmatwo} & \multicolumn{6}{|c}{\bf \gemmatwo} & \multicolumn{6}{|c}{\bf \gemmatwo} \\
        \cmidrule(lr){2-7}  \cmidrule(lr){8-13}  \cmidrule(lr){14-19}
        \bf Size & \multicolumn{2}{c|}{\bf 2B} & \multicolumn{2}{c|}{\bf 9B} & \multicolumn{2}{c|}{\bf 27B} & \multicolumn{2}{c|}{\bf 2B} & \multicolumn{2}{c|}{\bf 9B} & \multicolumn{2}{c|}{\bf 27B} & \multicolumn{2}{c|}{\bf 2B} & \multicolumn{2}{c|}{\bf 9B} & \multicolumn{2}{c}{\bf 27B} \\
        \cmidrule(lr){2-3} \cmidrule(lr){4-5} \cmidrule(lr){6-7} \cmidrule(lr){8-9} \cmidrule(lr){10-11} \cmidrule(lr){12-13} \cmidrule(lr){14-15} \cmidrule(lr){16-17} \cmidrule(lr){18-19}
        \bf Variant & \flan & \franken & \flan & \franken & \flan & \franken & \textsc{It} & \franken & \textsc{It} & \franken & \textsc{It} & \franken & \textsc{Mh} & \franken & \textsc{Mh} & \franken & \textsc{Mh} & \franken \\
        \midrule
        \endhead
        
        asm\_Beng & 11.2 & 14.4 & 30.8 & 26.4 & 34.8 & 35.2 & 15.6 & 12.4 & 48.0 & 42.4 & 53.2 & 48.0 & 20.4 & 29.2 & 46.0 & 28.0 & 43.2 & 36.0 \\
        bew\_Latn & 28.4 & 24.8 & 48.4 & 46.4 & 48.8 & 50.4 & 41.6 & 30.0 & 58.0 & 57.6 & 62.4 & 53.2 & 51.6 & 42.0 & 70.0 & 68.8 & 71.6 & 68.8 \\
        bho\_Deva & 16.4 & 17.2 & 29.2 & 35.6 & 38.0 & 35.6 & 23.2 & 19.2 & 48.4 & 48.0 & 50.8 & 51.6 & 29.2 & 36.4 & 51.2 & 48.8 & 49.6 & 49.6 \\
        doi\_Deva & 11.2 & 14.0 & 19.6 & 34.0 & 26.4 & 33.6 & 13.6 & 19.6 & 32.4 & 44.0 & 37.6 & 46.0 & 17.6 & 25.2 & 36.4 & 36.4 & 29.2 & 37.6 \\
        div\_Thaa & 6.8 & 10.4 & 9.6 & 23.2 & 18.8 & 26.8 & 2.8 & 12.4 & 15.6 & 36.8 & 22.0 & 32.8 & 2.8 & 15.6 & 17.2 & 26.4 & 13.2 & 12.0 \\
        dzo\_Tibt & 2.0 & 2.8 & 6.8 & 12.8 & 10.4 & 8.4 & 0.4 & 3.6 & 10.0 & 14.0 & 14.0 & 20.4 & 0.8 & 3.6 & 11.6 & 6.4 & 8.8 & 4.0 \\
        efi\_Latn & 8.0 & 10.4 & 13.2 & 27.6 & 16.0 & 30.8 & 5.2 & 5.6 & 17.2 & 31.6 & 24.8 & 20.8 & 6.8 & 20.0 & 21.6 & 48.0 & 19.6 & 38.0 \\
        gom\_Deva & 15.2 & 15.6 & 25.2 & 35.6 & 28.4 & 37.2 & 13.6 & 14.4 & 37.2 & 48.4 & 44.0 & 49.2 & 11.6 & 30.4 & 38.4 & 47.6 & 36.4 & 41.6 \\
\end{longtable}
\end{small}
\begin{small}
    \fasize
    \setlength{\tabcolsep}{2.9pt}
    \setlength{\LTcapwidth}{0.98\linewidth}
    \begin{longtable}{c| ccc|ccc| ccc|ccc|ccc| ccc|ccc}
    \caption{\ouradapter results on \xsumin with zero-shot prompting. \flan: LLMs instruction-tuned on \textsc{Flan} mixture; \franken: \ouradapter; \fala: \ouradapter with LoRA Adaptation.} \\
    \toprule
        \multicolumn{22}{c}{\bf \ouradapter Results on \xsumin} \\
        \midrule
        \bf Model & \multicolumn{6}{c|}{\bf \palmtwo} & \multicolumn{9}{c|}{\bf \gemmatwo} & \multicolumn{6}{c}{\bf \aya} \\
        \cmidrule(lr){2-7}  \cmidrule(lr){8-16}  \cmidrule(lr){17-22}
        \bf Size & \multicolumn{3}{c|}{\bf XXS} & \multicolumn{3}{c|}{\bf S} & \multicolumn{3}{c|}{\bf 2B} & \multicolumn{3}{c|}{\bf 9B} & \multicolumn{3}{c|}{\bf 27B} & \multicolumn{3}{c|}{\bf 8B} & \multicolumn{3}{c}{\bf 35B} \\
        \cmidrule(lr){2-4} \cmidrule(lr){5-7} \cmidrule(lr){8-10} \cmidrule(lr){11-13} \cmidrule(lr){14-16} \cmidrule(lr){17-19} \cmidrule(lr){20-22}
        \bf Variant & \flan & \franken & \fala & \flan & \franken & \fala & \flan & \franken & \fala & \flan & \franken & \fala & \flan & \franken & \fala & \flan & \franken & \fala & \flan & \franken & \fala \\
        \midrule
        \endhead
        eng\_Latn & 33.7 & 34.2 & 30.7 & 41.1 & 40.7 & 41.2 & 36.6 & 34.1 & 31.9 & 37.5 & 37.1 & 37.0 & 37.6 & 36.8 & 36.9 & 34.5 & 25.7 & 30.7 & 37.9 & 22.1 & 37.1 \\
        asm\_Beng & 0.3 & 0.3 & 0.3 & 25.5 & 16.1 & 6.3 & 2.7 & 1.3 & 7.6 & 13.7 & 16.5 & 20.1 & 17.8 & 11.3 & 21.3 & 5.0 & 1.3 & 4.6 & 6.7 & 6.7 & 19.4 \\
        awa\_Deva & 0.1 & 0.1 & 0.1 & 6.6 & 20.5 & 11.1 & 5.2 & 5.9 & 14.2 & 19.0 & 17.3 & 17.2 & 13.7 & 17.3 & 17.6 & 0.2 & 0.0 & 4.5 & 9.4 & 8.8 & 17.1 \\
        bgc\_Deva & 0.2 & 0.2 & 0.2 & 5.6 & 3.2 & 4.3 & 0.2 & 0.7 & 11.8 & 16.5 & 0.9 & 16.1 & 12.1 & 15.6 & 16.2 & 16.1 & 0.2 & 0.4 & 3.1 & 8.4 & 16.4 \\
        bho\_Deva & 0.1 & 0.1 & 0.1 & 19.3 & 21.6 & 18.5 & 2.0 & 6.9 & 9.9 & 19.1 & 18.0 & 16.7 & 14.2 & 15.8 & 18.1 & 16.3 & 0.0 & 7.5 & 3.7 & 12.8 & 18.1 \\
        ben\_Beng & 16.7 & 2.2 & 0.1 & 29.8 & 28.3 & 25.9 & 6.7 & 14.4 & 16.9 & 22.4 & 23.9 & 21.6 & 25.5 & 20.0 & 22.0 & 16.9 & 11.3 & 17.2 & 17.0 & 14.5 & 22.2 \\
        bod\_Tibt & 0.2 & 0.2 & 0.1 & 10.8 & 9.2 & 9.8 & 0.2 & 0.9 & 5.7 & 0.5 & 3.4 & 8.8 & 0.8 & 0.8 & 7.4 & 4.6 & 2.5 & 6.0 & 3.7 & 4.0 & 8.6 \\
        brx\_Deva & 0.2 & 0.2 & 0.2 & 0.3 & 1.0 & 3.5 & 0.2 & 0.7 & 1.5 & 0.2 & 1.3 & 5.2 & 0.3 & 11.4 & 14.8 & 4.5 & 0.0 & 0.7 & 0.2 & 0.4 & 8.4 \\
\end{longtable}
\end{small}

\begin{small}
    \fasize
    \setlength{\tabcolsep}{2.9pt}
    \setlength{\LTcapwidth}{0.98\linewidth}
    \begin{longtable}{c| ccc|ccc| ccc|ccc|ccc| ccc|ccc}
    \caption{\ouradapter results on \xorqain with zero-shot prompting. \flan: LLMs instruction-tuned on \textsc{Flan} mixture; \franken: \ouradapter; \fala: \ouradapter with LoRA Adaptation.} \\
    \toprule
        \multicolumn{22}{c}{\bf \ouradapter Results on \xorqain} \\
        \midrule
        \bf Model & \multicolumn{6}{c|}{\bf \palmtwo} & \multicolumn{9}{c|}{\bf \gemmatwo} & \multicolumn{6}{c}{\bf \aya} \\
        \cmidrule(lr){2-7}  \cmidrule(lr){8-16}  \cmidrule(lr){17-22}
        \bf Size & \multicolumn{3}{c|}{\bf XXS} & \multicolumn{3}{c|}{\bf S} & \multicolumn{3}{c|}{\bf 2B} & \multicolumn{3}{c|}{\bf 9B} & \multicolumn{3}{c|}{\bf 27B} & \multicolumn{3}{c|}{\bf 8B} & \multicolumn{3}{c}{\bf 35B} \\
        \cmidrule(lr){2-4} \cmidrule(lr){5-7} \cmidrule(lr){8-10} \cmidrule(lr){11-13} \cmidrule(lr){14-16} \cmidrule(lr){17-19} \cmidrule(lr){20-22}
        \bf Variant & \flan & \franken & \fala & \flan & \franken & \fala & \flan & \franken & \fala & \flan & \franken & \fala & \flan & \franken & \fala & \flan & \franken & \fala & \flan & \franken & \fala \\
        \midrule
        \endhead
        
        \multicolumn{22}{c}{\bf \textsc{Answer in En}} \\
        \midrule
        eng\_Latn & 70.9 & 75.5 & 74.8 & 75.1 & 82.1 & 82.6 & 71.2 & 66.6 & 73.3 & 75.7 & 74.6 & 74.4 & 75.5 & 78.5 & 75.9 & 75.6 & 72.6 & 71.0 & 78.4 & 65.2 & 74.9 \\
        asm\_Beng & 47.9 & 63.0 & 68.4 & 52.0 & 51.1 & 58.3 & 54.3 & 61.3 & 67.7 & 71.6 & 72.4 & 71.2 & 65.7 & 64.3 & 74.1 & 56.7 & 67.2 & 57.9 & 64.8 & 69.2 & 73.9 \\
        awa\_Deva & 48.4 & 64.0 & 64.6 & 49.7 & 52.9 & 57.7 & 49.6 & 54.2 & 57.9 & 63.2 & 66.3 & 68.1 & 63.3 & 58.8 & 70.7 & 68.3 & 64.9 & 63.2 & 77.1 & 67.6 & 72.9 \\
        bgc\_Deva & 43.4 & 61.1 & 67.0 & 52.9 & 59.5 & 58.1 & 45.1 & 53.0 & 60.4 & 57.9 & 68.6 & 70.9 & 60.8 & 57.1 & 73.7 & 70.1 & 69.0 & 66.4 & 74.6 & 67.9 & 71.1 \\
        bho\_Deva & 50.8 & 64.8 & 69.7 & 47.4 & 58.5 & 58.5 & 50.4 & 57.5 & 59.4 & 63.8 & 67.2 & 67.0 & 62.0 & 60.8 & 72.8 & 68.8 & 67.2 & 65.8 & 73.2 & 65.1 & 69.4 \\
        ben\_Beng & 57.3 & 62.5 & 71.0 & 52.0 & 58.5 & 63.1 & 59.8 & 64.1 & 67.7 & 74.5 & 74.2 & 72.6 & 65.9 & 64.9 & 73.7 & 68.4 & 69.0 & 63.1 & 76.6 & 68.6 & 70.2 \\
        bod\_Tibt & 3.2 & 2.6 & 17.5 & 23.5 & 12.4 & 38.5 & 11.4 & 36.0 & 40.9 & 27.5 & 41.7 & 33.8 & 43.5 & 44.7 & 60.8 & 32.0 & 30.1 & 23.5 & 30.3 & 36.4 & 30.6 \\
\end{longtable}
\end{small}


\end{document}